\pdfoutput=1

\documentclass[11pt]{article}

\usepackage{EMNLP2023}

\usepackage{times}
\usepackage{latexsym}

\usepackage[T1]{fontenc}

\usepackage[utf8]{inputenc}

\usepackage{microtype}

\usepackage{inconsolata}
\usepackage{graphicx}
\usepackage{enumitem}
\usepackage{xcolor}
\usepackage{soul}
\usepackage{CJKutf8}
\usepackage{xpinyin}
\usepackage{alltt}
\usepackage{amsmath}
\usepackage{multirow}
\usepackage{amssymb}
\usepackage{booktabs}
\usepackage{makecell}
\usepackage{pifont}
\usepackage{listings}
\usepackage{color}

\usepackage[labelformat=simple]{subcaption}

\newcommand{\hlc}[2][yellow]{{%
    \colorlet{foo}{#1}%
    \sethlcolor{foo}\hl{#2}}%
}

\colorlet{punct}{red!60!black}
\definecolor{background}{HTML}{EEEEEE}
\definecolor{delim}{RGB}{20,105,176}
\colorlet{numb}{magenta!60!black}

\lstdefinelanguage{json}{
    basicstyle=\normalfont\ttfamily,
    numbers=left,
    numberstyle=\scriptsize,
    stepnumber=1,
    numbersep=8pt,
    showstringspaces=false,
    breaklines=true,
    frame=lines,
    backgroundcolor=\color{background},
    literate=
     *{0}{{{\color{numb}0}}}{1}
      {1}{{{\color{numb}1}}}{1}
      {2}{{{\color{numb}2}}}{1}
      {3}{{{\color{numb}3}}}{1}
      {4}{{{\color{numb}4}}}{1}
      {5}{{{\color{numb}5}}}{1}
      {6}{{{\color{numb}6}}}{1}
      {7}{{{\color{numb}7}}}{1}
      {8}{{{\color{numb}8}}}{1}
      {9}{{{\color{numb}9}}}{1}
      {:}{{{\color{punct}{:}}}}{1}
      {,}{{{\color{punct}{,}}}}{1}
      {\{}{{{\color{delim}{\{}}}}{1}
      {\}}{{{\color{delim}{\}}}}}{1}
      {[}{{{\color{delim}{[}}}}{1}
      {]}{{{\color{delim}{]}}}}{1},
}

\newcommand{\platform}{\textsc{CLEVA}}

\definecolor{ugreen}{cmyk}{1,0,1,0.498}

\title{CLEVA: Chinese Language Models EVAluation Platform}

\author{Yanyang Li$^1$\thanks{\ \ Equal contributions.}\ , Jianqiao Zhao$^1$\footnotemark[1]\ , Duo Zheng$^1$, Zi-Yuan Hu$^1$, Zhi Chen$^3$, Xiaohui Su$^3$,\\
\textbf{Yongfeng Huang$^1$, Shijia Huang$^1$, Dahua Lin$^{2,3}$, Michael R. Lyu$^1$, Liwei Wang$^1$\thanks{\ \ Project leader and corresponding author}}\\
$^1$Department of Computer Science and Engineering, The Chinese University of Hong Kong \\
$^2$Department of Information Engineering, The Chinese University of Hong Kong\\
$^3$Shanghai AI Laboratory\\
}

\begin{document}
\maketitle
\begin{abstract}
With the continuous emergence of Chinese Large Language Models (LLMs), how to evaluate a model's capabilities has become an increasingly significant issue.
The absence of a comprehensive Chinese benchmark that thoroughly assesses a model's performance, the unstandardized and incomparable prompting procedure, and the prevalent risk of contamination pose major challenges in the current evaluation of Chinese LLMs.
We present \platform{}, a user-friendly platform crafted to holistically evaluate Chinese LLMs.  
Our platform employs a standardized workflow to assess LLMs' performance across various dimensions, regularly updating a competitive leaderboard. 
To alleviate contamination, \platform{} curates a significant proportion of new data and develops a sampling strategy that guarantees a unique subset for each leaderboard round.
Empowered by an easy-to-use interface that requires just a few mouse clicks and a model API, users can conduct a thorough evaluation with minimal coding.
Large-scale experiments featuring 23 Chinese LLMs have validated \platform{}'s efficacy.
Our GitHub repo is \url{https://github.com/LaVi-Lab/CLEVA}.
\end{abstract}

\section{Introduction}
Large language models (LLMs) have fundamentally revolutionized natural language processing.
Transformer models with more than 100B parameters have exhibited remarkable generalization ability across diverse tasks without the need for fine-tuning.
The success of GPT-4~\cite{DBLP:journals/corr/abs-2303-08774} and ChatGPT sparked a trend of training Chinese LLMs, with new models launching almost every week~\cite{DBLP:conf/iclr/ZengLDWL0YXZXTM23,2023internlm,leng2023chinese-vicuna,DBLP:journals/corr/abs-2303-14742,DBLP:journals/corr/abs-2304-08177}.
These rapid developments aggravate the need for Chinese LLM evaluation.

\noindent\textbf{Assessing the capacity of LLMs is non-trivial.}
Traditional practices that evaluate models on a single task at a time are gradually becoming obsolete, since a single task can hardly characterize a full view of an LLM's capacity.
Instead, to effectively grasp a holistic view of an LLM's capacity, we need to decompose its capacity into various abilities, evaluate these abilities with numerous corresponding tasks, and measure the competence of each task with multiple metrics.
In this sense, HELM~\cite{DBLP:journals/corr/abs-2211-09110}, leads the way in English LLM evaluation, as it conducts an in-depth evaluation of English LLMs on various NLP tasks using seven metrics.
In Chinese, previous attempts have shown limitations, either in task selection or the metrics adopted. 
C-Eval~\cite{DBLP:journals/corr/abs-2305-08322}, M3KE~\cite{DBLP:journals/corr/abs-2305-10263}, CMMLU~\cite{DBLP:journals/corr/abs-2306-09212}, GAOKAO-Bench~\cite{DBLP:journals/corr/abs-2305-12474}, and MMCU~\cite{DBLP:journals/corr/abs-2304-12986} narrow down to knowledge and reasoning abilities, whose datasets are mostly constructed using Chinese exams.
By the time of our submission, OpenCompass~\cite{2023opencompass}, with around 74K Chinese queries out of 300K total, leans on accuracy as its sole metric, overlooking other important aspects in LLM evaluation.
FlagEval~\cite{2023flageval} inherits four out of seven metrics from HELM and 22 existing Chinese datasets, having limited coverage on some significant tasks.
A comprehensive Chinese benchmark incorporating diverse metrics to holistically evaluate Chinese LLMs is urgently demanded.

\noindent\textbf{Prompt-based evaluation in Chinese is largely unstandardized.}
Previous evaluations, such as HELM~\cite{DBLP:journals/corr/abs-2211-09110}, do not explicitly optimize prompts, though LLMs' significant sensitivity to the format of prompt has been observed~\cite{DBLP:conf/naacl/WebsonP22,DBLP:conf/acl/AbdouRKS22, DBLP:conf/iclr/T0pp}.
Moreover, unlike many English benchmarks that have well-developed prompts (\S~\ref{sec:prelim}), many Chinese benchmarks are in their early stage and do not enjoy such privileges.
Chinese LLMs are evaluated using different prompts, making the results incomparable and hence untrustworthy.

Consuming up to trillions of tokens during pretraining, LLMs are prone to train-test contamination~\cite{DBLP:conf/nips/BrownMRSKDNSSAA20}, which significantly threatens the validity of an evaluation.
Previous work~\cite{DBLP:journals/corr/abs-2303-08774, DBLP:journals/corr/abs-2211-09110} approaches this issue more from a consequentialist perspective: They examine the contamination risk, by methods like long n-gram overlap, only after the evaluation has been done.
These post-evaluation analyses, though responsibly examining if train-test contamination happens, cannot alleviate the risk of contamination in the first place.
A proactive method to mitigate the contamination risk is of great importance.

We present \platform{}, \textbf{C}hinese \textbf{L}anguage models \textbf{EVA}luation platform that tackles the aforementioned problems with the following features:
\begin{itemize}[noitemsep, nolistsep]
    \item \textbf{A comprehensive Chinese benchmark.} 
    Inspired by HELM~\cite{DBLP:journals/corr/abs-2211-09110}, \platform{} organizes the evaluation tasks into two parts: \emph{ability evaluation}, which gauges specific LLM skills and \emph{application assessment}, which tests how well LLMs apply their skills to real-world applications (\S~\ref{sec:taxonomy}).
    Most of the well-accepted Chinese datasets relevant to our ability evaluation or application assessment are organized, standardized, and then adopted by our platform.
    More importantly, we design new Chinese-specific tasks, e.g., Pinyin transliteration and intent understanding, and collect a substantial amount of new data, accounting for 33.98\% of our total data.
    As for the metrics (\S~\ref{sec:taxonomy}), we incorporate metrics for diversity and privacy into our system in addition to the seven in HELM.
    With 370K (over 9 million queries after augmentation) test instances from 84 datasets and 9 metrics, \platform{}, so far, stands as the most extensive Chinese evaluation dataset and possesses the most dimensions, facilitating a holistic evaluation of Chinese LLMs. 
    \item \textbf{Standardized prompt-based evaluation methodology.} 
    \platform{} takes full control of key aspects of LLM evaluation, with data and prompts being the most important among them.
    All data are jointly prepared with unified preprocessing steps, ensuring a level playing field for all LLMs.
    Meanwhile, \platform{} provides a set of prompts, instead of just one prompt as in prior work, for each task for prompting-based inference~\cite{DBLP:conf/nips/BrownMRSKDNSSAA20}.
    This prompt design ensures comparable evaluation results by standardizing the prompts used for testing, while also encouraging further analysis of LLMs' sensitivity to different prompts~\cite{DBLP:journals/corr/abs-2306-04528}.
    \item \textbf{An up-to-date and trustworthy leaderboard.} 
    \platform{} advocates a proactive method for securing trustworthy evaluation results.
    By collecting extensive new data, \platform{} suppresses the leakage of testing data prior to the evaluation.
    Moreover, we frequently organize new evaluation rounds, sampling a unique test set from 9 million augmented instances.
    This strategy further mitigates the risk of train-test contamination, improving the trustworthiness and timeliness of the leaderboard.
\end{itemize}

\platform{} is thoroughly validated by benchmarking 23 Chinese LLMs on our large-scale test sets (\S~\ref{sec:evaluation}).
The corresponding leaderboard and all other user-friendly features will be continuously maintained and improved to accommodate new models and evaluation methods.

\section{Related Work}

LLM evaluation is a rapidly developing field in recent years to delineate the boundary of LLM's capability.
In English, various systematic evaluation benchmarks have been proposed.
For example, BIG-Bench~\cite{srivastava2023beyond} is the largest collection that covers more than 200 tasks.
HELM~\cite{DBLP:journals/corr/abs-2211-09110} organizes tasks into core scenarios, which focus on use cases, and targeted evaluation, which aims to better understand models.
HELM also presents a multi-metric measurement that enables analysis of tradeoffs for each scenario.
Recently, AGIEval~\cite{DBLP:journals/corr/abs-2304-06364} is proposed to evaluate LLMs using challenging human exams.
PromptBench~\cite{DBLP:journals/corr/abs-2306-04528}, on the other hand, measures the robustness of LLMs to prompts via adversarial attacks.
MT-Bench~\cite{DBLP:journals/corr/abs-2306-05685} collects multi-turn questions and presents the Chatbot Arena platform that treats GPT-4~\cite{DBLP:journals/corr/abs-2303-08774} as the judge.

While \platform{} shares the same fundamental motivation with HELM~\cite{DBLP:journals/corr/abs-2211-09110}, to holistically evaluate language learning models in their original languages, \platform{} is far from a mere Chinese replica of HELM.
Building on the foundation of HELM's taxonomy, \platform{} introduces a range of tasks, with particular emphasis on those unique to Chinese, to better assess the capabilities of Chinese LLMs.
It offers a new perspective on prompts, providing abundant prompt templates to standardize evaluation and encourage in-depth exploration of models' sensitivity. 
In terms of metrics, \platform{} expands into new areas of diversity and privacy for a more comprehensive evaluation. 
Finally, \platform{} proactively mitigates train-test contamination by collecting a significant amount of new data, creating unique test sets by sampling, and regularly updating the leaderboard.
All of these evaluation designs are neatly packaged in a user-friendly platform to facilitate community usage.


There is also a lot of progress in evaluating Chinese LLMs~\cite{DBLP:journals/corr/abs-2305-08322,DBLP:journals/corr/abs-2305-10263,DBLP:journals/corr/abs-2306-09212,DBLP:journals/corr/abs-2305-12474,DBLP:journals/corr/abs-2304-12986}.
OpenCompass~\cite{2023opencompass} and FlagEval~\cite{2023flageval} are two important attempts to evaluate Chinese LLMs.
OpenCompass pools 53 public datasets and uses standard accuracy-like metrics as the only measurement for each dataset.
FlagEval, with a smaller number of datasets and metrics, still needs further expansion to achieve sufficient coverage.
Compared to previous efforts, \platform{} offers Chinese data from 84 datasets, including 33.98\% original queries, while employing the broadest range of metrics to promote holistic evaluation.
\platform{} standardizes prompts (\S~\ref{sec:system}) and mitigates data contamination issues, pioneering new paths for LLM evaluation in general.

\section{Preliminaries}
\label{sec:prelim}


To measure the model performance on a task, a relevant \emph{test set} is constituted from a collection of \emph{instances}.
A test instance will contain multiple \emph{input fields} (string typically) and a list of \emph{references}.

We then adopt a \emph{prompt template}, which essentially describes how to assemble the model input, a.k.a, \emph{prompt}, from input fields~\cite{bach2022promptsource}.
For example, a Chinese paraphrase identification prompt template (and its translation) is:
\begin{quote}
    \small
    \textbf{\textsl{Chinese Example}}:\\
    \begin{CJK*}{UTF8}{gbsn}
    ``\hlc[cyan!50]{\{sentence1\}}''和``\hlc[orange!50]{\{sentence2\}}''这两个问题是在问同一件事情吗？
    \end{CJK*}\\
    \\
    \textbf{\textsl{English Translation}}:\\
    Are the questions ``\hlc[cyan!50]{\{sentence1\}}'' and ``\hlc[orange!50]{\{sentence2\}}'' asking the same thing?
\end{quote}
where \hlc[cyan!50]{\{sentence1\}} and \hlc[orange!50]{\{sentence1\}} are two input fields that will be replaced by the two candidate questions in the test instance.
The prompt will be fed into a black-box LLM to predict an output string together with its probability.

Finally, all model predictions and the corresponding test instances will be passed into a \emph{metric} to obtain a numerical value that indicates how well the model performs.
Following HELM~\cite{DBLP:journals/corr/abs-2211-09110}, a \emph{metric} in this paper is an umbrella for a dimension of measures that share similar purposes.
For example, the \emph{accuracy} metric corresponds to BLEU for translation and pass@k for code synthesis.
We employ nine metrics, foregrounding metrics beyond accuracy and ensuring a holistic evaluation.





\section{System Design}
\label{sec:system}

\platform{} aims to deliver the following two key assets to users who try to evaluate their own LLMs:
\begin{itemize}[noitemsep, nolistsep]
    \item A comprehensive and thorough \textbf{assessment report} that informs users of the strength and limitations of their models.
    \item A trustworthy \textbf{leaderboard} reflecting the latest advancement of LLMs.
\end{itemize}
We will discuss our taxonomy that ensures comprehensive evaluations, and challenges like train-test contamination in leaderboard maintenance.



\subsection{Evaluation Taxonomy}
\label{sec:taxonomy}

\begin{figure*}[t!]
\begin{center}
\includegraphics[width=0.8\linewidth]{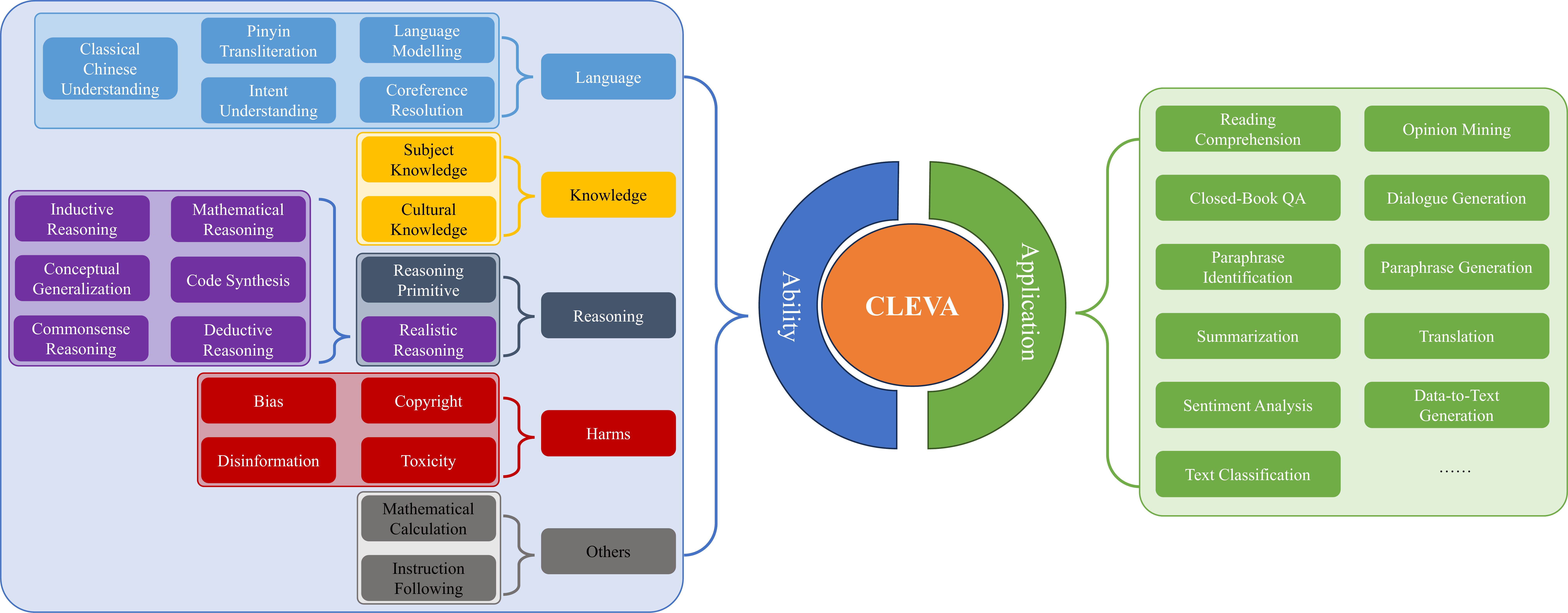}
\end{center}
\vspace{-10pt}
\caption{\platform{} benchmark.} 
\label{fig:taxonomy}
\vspace{-10pt}
\end{figure*}

Inspired by HELM~\cite{DBLP:journals/corr/abs-2211-09110}, we present a \textbf{Tasks$\times$Prompts$\times$Metrics} evaluation taxonomy for users to evaluate their models.
Our evaluation taxonomy carefully designs a Chinese benchmark targeting various LLM abilities, employs a set of diverse prompt templates for each task to characterize the model performance variance, and adopts multiple metrics to comprehensively assess LLMs.

\noindent\textbf{Tasks.}
As shown in Figure~\ref{fig:taxonomy}, our Chinese LLM evaluation benchmark consists of two parts: \emph{ability evaluation} and \emph{application assessment}.
Each task in ability evaluation focuses on one special skill of LLMs, while application assessment involves real-world NLP tasks that require LLMs to solve practical use cases with their skill sets.
Ability evaluation assesses LLM ability from five aspects:
\begin{itemize}[noitemsep, nolistsep]
    \item \textbf{Language} measures how well LLMs understand Chinese.
    In addition to three conventional tasks, we incorporate two tasks specific to Chinese: \emph{Pinyin transliteration} and \emph{classical Chinese understanding}.
    \item \textbf{Knowledge} focuses on assessing the capacity of knowledge acquired by LLMs.
    We further segment our evaluation into \emph{subject knowledge} and \emph{cultural knowledge} (mainly Chinese culture) based on the source of knowledge.
    This fine-grained design allows users to closely analyze the model performance across different knowledge categories.
    \item \textbf{Reasoning} evaluates LLMs' reasoning ability in two settings: \emph{reasoning primitives}, which is independent of language and knowledge background, and \emph{realistic reasoning} that requires reasoning with domain knowledge on practical scenarios.
    On top of HELM, we additionally include \emph{commonsense reasoning}, \emph{inductive reasoning}, \emph{conceptual generalization}, and \emph{deductive reasoning}.
    \item \textbf{Harms} evaluates the potential risk of LLMs in \emph{copyright}, \emph{disinformation}, \emph{bias}, and \emph{toxicity}.
    \item \textbf{Others} is newly introduced to include crucial yet uncategorized tasks like \emph{mathematical calculation} and \emph{instruction following}.
\end{itemize}

For application assessment, \platform{} features 11 real-world NLP tasks. 
In addition to the core scenarios of HELM, we newly include \emph{opinion mining}, \emph{dialogue generation}, \emph{paraphrase generation}, \emph{translation}, \emph{paraphrase identification}, and \emph{data-to-text generation}.
A detailed description of each task is documented in Appendix~\ref{app:benchmark}.

We instantiate the aforementioned tasks in two ways: by directly adopting related public Chinese datasets and by collecting new data.
For well-studied tasks, widely-recognized datasets are the best options for forming our benchmark.
However, many important tasks, such as \emph{reasoning primitive}, \emph{Pinyin transliteration}, and \emph{disinformation}, lack corresponding Chinese datasets, making the evaluation even more challenging.
On these occasions, we either synthesize using sophisticated rule-based scripts (e.g., reasoning primitive) or enlist professional human annotators to construct new test sets (See Appendix~\ref{app:annotate} for annotation details).
In total, the 31 tasks include 370K test instances from 84 datasets (9M queries in total after applying multiple prompt templates and data augmentation), 33.98\% of which are newly collected.

\noindent\textbf{Prompts.}
Ideally, an LLM should be a general interface, capable of understanding prompts with the same semantics, regardless of variations in surface forms.
However, LLMs' notorious sensitivity to prompt templates hinders accurate evaluation~\cite{DBLP:conf/naacl/WebsonP22,DBLP:conf/acl/AbdouRKS22}, leading to results that are sometimes incomparable.
To better understand an LLM's sensitivity to plausible human instructions, multiple prompt templates are needed, rather than a single template as in previous work~\cite{2023flageval,2023opencompass,DBLP:journals/corr/abs-2211-09110}.

In this work, we manually annotate an average of 3.95 prompt templates for each test set and support all major prompting formats.
\platform{} calculates the performance statistics across the entire set of prompts.
These statistics do more than just examine the robustness to prompt templates, as reflected by the standard deviation; they also help estimate the upper and lower bounds of an LLM's performance on a specific test set, as indicated by the minimum and maximum values.
Users can benefit from these statistics to select models and to make informed trade-offs between performance and investment in prompt engineering.
More discussions on prompt templates we provided are in Appendix~\ref{app:prompt}.

\noindent\textbf{Metrics.}
We adopt the 7 metrics from HELM for a holistic evaluation, and, to address recent interest in chatbots and safety concerns, we add two new dimensions: \emph{diversity} and \emph{privacy}.
\begin{itemize}[noitemsep, nolistsep]
    \item \textbf{Accuracy.} Accuracy refers to the standard metrics to measure model performance on different tasks, e.g., F1 score for question answering and ROUGE score for summarization.
    \item \textbf{Calibration and uncertainty.} Calibration represents the gap between the model confidence and its actual error rate and is measured mainly by expected calibration error (ECE, ~\cite{DBLP:conf/aaai/NaeiniCH15}).
    \item \textbf{Robustness.} Robustness is the worst-case performance of a model across transformations of test instances.
    We focus on semantics-preserving perturbations as there are many well-studied data augmentation tools. 
    \item \textbf{Fairness.} Similar to robustness, fairness employs perturbations related to social groups to test the disparate treatment and disparate impact of LLMs.
    \item \textbf{Bias and stereotypes.} We quantify bias as the disproportionate representation of different social groups. This is gauged through the rates at which these groups are mentioned during model generation. Additionally, we interpret stereotypes as uneven associations between these social groups and certain stereotyped terms, such as occupational roles.
    \item \textbf{Toxicity.} Following HELM~\cite{DBLP:journals/corr/abs-2211-09110}, toxicity is a general term that covers hate speech, abusive language, etc.
    \item \textbf{Efficiency.} Efficiency is a rather broad concept that has many subtleties.
    It could refer to training or inference efficiency and is measured by energy, carbon, and wall-clock time.
    As most information could be confidential, we focus only on the inference wall-clock time.
    \item \textbf{Diversity.} 
    Given the popularity of LLM-based chatbots, we incorporate the conventional diversity metric in dialogue systems that evaluates the response surface form diversity~\cite{DBLP:conf/naacl/LiGBGD16}.
    Here, we employ the diversity metrics from \citet{DBLP:conf/emnlp/MillerFBBFLPW17}.
    \item \textbf{Privacy.} 
    In the real-world deployment of LLMs, detecting private information in the generated text, such as Personally Identifiable Information (PII), is a challenging yet important question.
    We report the portion of PII in the whole test set to make the privacy evaluation generalizable.
    \platform{} adopts some established tools to smoothly detect PII, and we are working on accommodating more aspects of private content in the near future.
\end{itemize}
Detailed metric lists are provided in Appendix~\ref{app:metrics}.

\begin{figure*}[t!]
\begin{center}
\includegraphics[width=\linewidth]{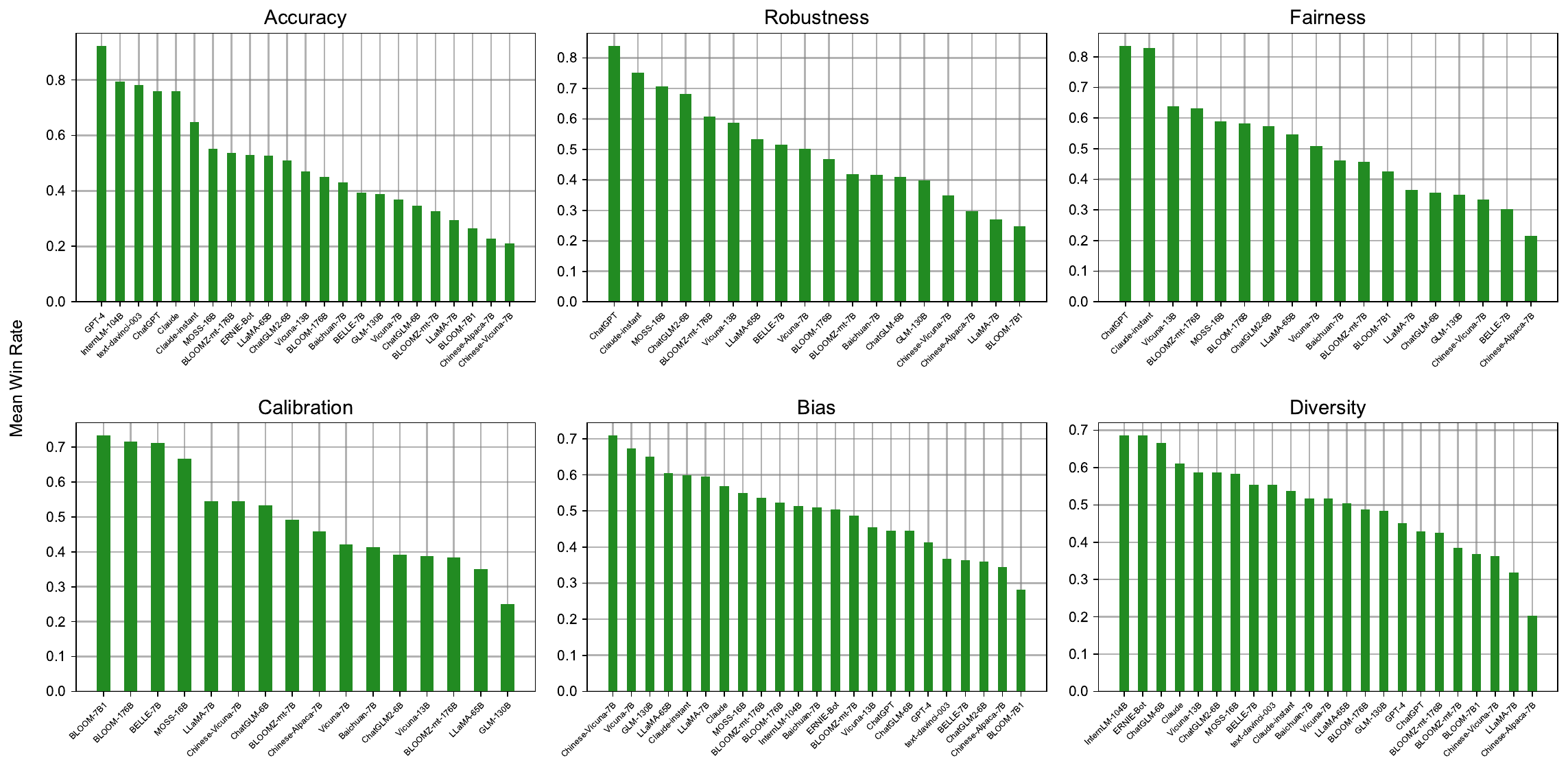}
\end{center}
\vspace{-10pt}
\caption{
The mean win rate of 23 models in 31 tasks. 
The mean win rate is the probability of a model outperforming a random different model on a random task.
We exclude toxicity, privacy, and efficiency metrics as all models excel in the former two, and the latter is often paired with other metrics to deliver meaningful comparisons.
Since robustness and fairness involve expensive data augmentation, we only evaluate ChatGPT and Claude-instant.}
\label{fig:head}
\vspace{-10pt}
\end{figure*}

\subsection{Leaderboard \& Data Contamination}
\label{sec:leaderboard}

Ensuring fairness, objectivity, and authority is central to maintaining a trustworthy leaderboard.
Previous work~\cite{DBLP:conf/nips/BrownMRSKDNSSAA20} has reported \textbf{train-test contamination}, a situation where the test set is included in the training data, leading to unreliable evaluations.
Many existing benchmarks, e.g., ~\citet{DBLP:journals/corr/abs-2305-08322}, conceal the test set labels to avoid data contamination.
Given the small scale of their test sets and the large-scale training corpora used by modern LLMs, the risk of unintentional train-test contamination remains high.
\citet{DBLP:journals/corr/abs-2304-10436} address this problem by making the official test set private and requiring users to submit models' weights for evaluation.
However, this arrangement is unpopular because numerous cutting-edge models consider their weights highly confidential.

We advocate ``mutual confidentiality'' in LLM evaluation:
Users need not expose their model details, and the platform should minimize the risk of disclosing its test set.
Instead of model weights, \platform{} only requires API access.
We proactively achieve the other half of mutual confidentiality by continuously collecting new data and frequently organizing leaderboard rounds with unique test sets sampling from our full-scale 9 million augmented instances.
These strategies not only improve evaluation efficiency but also alleviate train-test contamination from data and temporal perspectives.

To make sure that the sampled subset delivers accurate results, our sampling strategy is not just random sampling: It estimates an acceptable approximation error threshold (i.e., within this threshold, the evaluation results on the sampled set have at least a 70\% chance to correctly rank any model pairs), then adjusts the sampling rate for each task according to this threshold, reducing the risk of over-/under-estimating the model performance.

\section{Usage Example}

Upon authentication, users are immediately presented with an interactive summary of our evaluation results of 23 LLMs. 
Users can select from these models, freely exploring the evaluation results from all 9 metrics and 31 tasks.

\platform{} simplifies the evaluation process of new models with minimal coding required. 
If a user has a model to evaluate, the user only needs a few minutes to finish these three steps: entering the model's API, selecting relevant tasks from 31 choices, and picking desired metrics from 9 options.
\platform{} will autonomously call the user's model, extract the corresponding responses, and compute the final metrics. 
Detailed descriptions and screenshots of \platform{} are listed in Appendix~\ref{app:platform}.


\section{Evaluation}
\label{sec:evaluation}

\noindent\textbf{Setup.}
We sample 6.43\% of our data to test 23 models that support Chinese (See Appendix~\ref{app:model}).
As for the cost, for example, it takes roughly 1600 GPU hours (NVIDIA A100 80G) to evaluate BLOOMZ-176B-mt~\cite{DBLP:conf/acl/MuennighoffWSRB23}.

\noindent\textbf{Results \& Analysis.}
Figure~\ref{fig:head} ranks all models by their mean win rates under different metrics.
\begin{itemize}[noitemsep, nolistsep]
\item \textbf{Accuracy.} It can be seen that GPT-4~\cite{DBLP:journals/corr/abs-2303-08774} has the highest winning rate, followed by other limited-accessed models.
This result shows \ul{a considerable margin between the performance of open-source models and limited-accessed models}.
Recent small instruction-following models are better than large LLMs without instruction-tuning, and are even better than some early large instruction-following models, indicating the necessity of effective instruction tuning.
\item \textbf{Robustness.} The trend on robustness is roughly the same as that of accuracy, with the exception of LLaMA~\cite{DBLP:journals/corr/abs-2302-13971}.
\item \textbf{Fairness.} 
Most of the model rankings have changed.
One possible reason is that fairness involves simplified-to-traditional conversion (See Appendix~\ref{app:metrics}), and many models have rarely seen traditional Chinese in pretraining.
\item \textbf{Calibration.} 
We report \texttt{ECE-10}~\cite{DBLP:conf/nips/KumarLM19} following HELM.
We find that \ul{models with more parameters tend to have higher ECE.}
For example, GLM-130B~\cite{DBLP:conf/iclr/ZengLDWL0YXZXTM23} and LLaMA-65B rank at the bottom.
For BLOOMZ-mt-7B vs BLOOMZ-mt-176B and BLOOM-7B1 vs BLOOM-176B~\cite{DBLP:journals/corr/abs-2211-05100}, the smaller one wins.
\item \textbf{Bias.} We focus on gender bias for comparison.
GPT-4 and other models, which rank top by other metrics, are at the bottom, while \ul{most of the open-source models have low bias}.
This is because open-source models usually output shorter, resulting in a lower risk of bias.
\item \textbf{Diversity.} 
We choose inter-distinct to compare different models.
\ul{Open-source models generate more diverse and innovative expression} than limited-accessed ones, probably due to their fewer safety concerns.
\end{itemize}
More detailed results and analysis are in Appendix~\ref{app:results}.

\section{Conclusion}

We present \platform{}, a Chinese LLM evaluation platform.
With the largest scale of Chinese instances and broadest metrics, \platform{} provides a comprehensive benchmark to holistically evaluate Chinese LLMs.
\platform{} standardizes key components, such as prompt templates, to make evaluation comparable.
It also proactively mitigates the contamination issue by collecting large-scale new data, sampling for unique test sets, and regularly updating the leaderboard.

\section*{Limitations}

Without further information needed from users, we can only use the inference walk-clock time as the metric, which may have a larger variance when the network is unstable.
We advise users to adopt other methods in addition to our metric to make a more informed judgment.

In addition, how to evaluate privacy is still a challenging problem. 
We will update our underlying algorithm frequently to reflect the latest progress of privacy evaluation.


\section*{Ethics Statement}
We consider the ethics issue in two folds, responsible data collection and usage.
We widely adopt manual data collection to enhance the variety of the tasks supported by \platform{}.
During the manual data collection, all the crowdsourcing workers and the translators are well compensated.
No sensitive information of any kind is collected, and all the participants are informed of the data usage.

\platform{} involves tasks that evaluate LLMs' performance on harm.
Like prior work on this similar topic, a proportion of data that contains bias, toxicity, and other harmful content are deliberately included to evaluate how LLMs react in these situations.
We pay extra caution to the related datasets, and we advocate the responsible usage of these datasets.
These datasets should only be used for LLM evaluation.
Our sampling mechanism also reduces the unwanted leakage of the data.

\section*{Acknowledgements}
This work was supported by National Key R\&D Program of China (Project No. 2022ZD0161200, 2022ZD0161201).
It was also partially funded by the Centre for Perceptual and Interactive Intelligence (CPIl) Ltd under the Innovation and Technology Commission (ITC)'s InnoHK. 
Liwei Wang is a Principal Investigator of CPII under the InnoHK. 
This work was partially supported by the Research Grants Council of the Hong Kong Special Administrative Region, China (No. CUHK 14206921 of the General Research Fund).

\bibliography{anthology,custom}

\begin{thebibliography}{93}
\expandafter\ifx\csname natexlab\endcsname\relax\def\natexlab#1{#1}\fi

\bibitem[{Abdou et~al.(2022)Abdou, Ravishankar, Kulmizev, and
  S{\o}gaard}]{DBLP:conf/acl/AbdouRKS22}
Mostafa Abdou, Vinit Ravishankar, Artur Kulmizev, and Anders S{\o}gaard. 2022.
\newblock \href {https://doi.org/10.18653/v1/2022.acl-long.476} {Word order
  does matter and shuffled language models know it}.
\newblock In \emph{Proceedings of the 60th Annual Meeting of the Association
  for Computational Linguistics (Volume 1: Long Papers), {ACL} 2022, Dublin,
  Ireland, May 22-27, 2022}, pages 6907--6919. Association for Computational
  Linguistics.

\bibitem[{Askell et~al.(2021)Askell, Bai, Chen, Drain, Ganguli, Henighan,
  Jones, Joseph, Mann, DasSarma, Elhage, Hatfield{-}Dodds, Hernandez, Kernion,
  Ndousse, Olsson, Amodei, Brown, Clark, McCandlish, Olah, and
  Kaplan}]{DBLP:journals/corr/abs-2112-00861}
Amanda Askell, Yuntao Bai, Anna Chen, Dawn Drain, Deep Ganguli, Tom Henighan,
  Andy Jones, Nicholas Joseph, Benjamin Mann, Nova DasSarma, Nelson Elhage, Zac
  Hatfield{-}Dodds, Danny Hernandez, Jackson Kernion, Kamal Ndousse, Catherine
  Olsson, Dario Amodei, Tom~B. Brown, Jack Clark, Sam McCandlish, Chris Olah,
  and Jared Kaplan. 2021.
\newblock \href {http://arxiv.org/abs/2112.00861} {A general language assistant
  as a laboratory for alignment}.
\newblock \emph{CoRR}, abs/2112.00861.

\bibitem[{Bach et~al.(2022)Bach, Sanh, Yong, Webson, Raffel, Nayak, Sharma,
  Kim, Bari, Fevry, Alyafeai, Dey, Santilli, Sun, Ben-David, Xu, Chhablani,
  Wang, Fries, Al-shaibani, Sharma, Thakker, Almubarak, Tang, Tang, Jiang, and
  Rush}]{bach2022promptsource}
Stephen~H. Bach, Victor Sanh, Zheng-Xin Yong, Albert Webson, Colin Raffel,
  Nihal~V. Nayak, Abheesht Sharma, Taewoon Kim, M~Saiful Bari, Thibault Fevry,
  Zaid Alyafeai, Manan Dey, Andrea Santilli, Zhiqing Sun, Srulik Ben-David,
  Canwen Xu, Gunjan Chhablani, Han Wang, Jason~Alan Fries, Maged~S.
  Al-shaibani, Shanya Sharma, Urmish Thakker, Khalid Almubarak, Xiangru Tang,
  Xiangru Tang, Mike Tian-Jian Jiang, and Alexander~M. Rush. 2022.
\newblock \href {http://arxiv.org/abs/2202.01279} {Promptsource: An integrated
  development environment and repository for natural language prompts}.

\bibitem[{Bai et~al.(2022{\natexlab{a}})Bai, Jones, Ndousse, Askell, Chen,
  DasSarma, Drain, Fort, Ganguli, Henighan, Joseph, Kadavath, Kernion, Conerly,
  Showk, Elhage, Hatfield{-}Dodds, Hernandez, Hume, Johnston, Kravec, Lovitt,
  Nanda, Olsson, Amodei, Brown, Clark, McCandlish, Olah, Mann, and
  Kaplan}]{DBLP:journals/corr/abs-2204-05862}
Yuntao Bai, Andy Jones, Kamal Ndousse, Amanda Askell, Anna Chen, Nova DasSarma,
  Dawn Drain, Stanislav Fort, Deep Ganguli, Tom Henighan, Nicholas Joseph,
  Saurav Kadavath, Jackson Kernion, Tom Conerly, Sheer~El Showk, Nelson Elhage,
  Zac Hatfield{-}Dodds, Danny Hernandez, Tristan Hume, Scott Johnston, Shauna
  Kravec, Liane Lovitt, Neel Nanda, Catherine Olsson, Dario Amodei, Tom~B.
  Brown, Jack Clark, Sam McCandlish, Chris Olah, Benjamin Mann, and Jared
  Kaplan. 2022{\natexlab{a}}.
\newblock \href {https://doi.org/10.48550/arXiv.2204.05862} {Training a helpful
  and harmless assistant with reinforcement learning from human feedback}.
\newblock \emph{CoRR}, abs/2204.05862.

\bibitem[{Bai et~al.(2022{\natexlab{b}})Bai, Kadavath, Kundu, Askell, Kernion,
  Jones, Chen, Goldie, Mirhoseini, McKinnon, Chen, Olsson, Olah, Hernandez,
  Drain, Ganguli, Li, Tran{-}Johnson, Perez, Kerr, Mueller, Ladish, Landau,
  Ndousse, Lukosiute, Lovitt, Sellitto, Elhage, Schiefer, Mercado, DasSarma,
  Lasenby, Larson, Ringer, Johnston, Kravec, Showk, Fort, Lanham,
  Telleen{-}Lawton, Conerly, Henighan, Hume, Bowman, Hatfield{-}Dodds, Mann,
  Amodei, Joseph, McCandlish, Brown, and
  Kaplan}]{DBLP:journals/corr/abs-2212-08073}
Yuntao Bai, Saurav Kadavath, Sandipan Kundu, Amanda Askell, Jackson Kernion,
  Andy Jones, Anna Chen, Anna Goldie, Azalia Mirhoseini, Cameron McKinnon,
  Carol Chen, Catherine Olsson, Christopher Olah, Danny Hernandez, Dawn Drain,
  Deep Ganguli, Dustin Li, Eli Tran{-}Johnson, Ethan Perez, Jamie Kerr, Jared
  Mueller, Jeffrey Ladish, Joshua Landau, Kamal Ndousse, Kamile Lukosiute,
  Liane Lovitt, Michael Sellitto, Nelson Elhage, Nicholas Schiefer,
  Noem{\'{\i}} Mercado, Nova DasSarma, Robert Lasenby, Robin Larson, Sam
  Ringer, Scott Johnston, Shauna Kravec, Sheer~El Showk, Stanislav Fort, Tamera
  Lanham, Timothy Telleen{-}Lawton, Tom Conerly, Tom Henighan, Tristan Hume,
  Samuel~R. Bowman, Zac Hatfield{-}Dodds, Ben Mann, Dario Amodei, Nicholas
  Joseph, Sam McCandlish, Tom Brown, and Jared Kaplan. 2022{\natexlab{b}}.
\newblock \href {https://doi.org/10.48550/arXiv.2212.08073} {Constitutional
  {AI:} harmlessness from {AI} feedback}.
\newblock \emph{CoRR}, abs/2212.08073.

\bibitem[{bench authors(2023)}]{srivastava2023beyond}
BIG bench authors. 2023.
\newblock \href {https://openreview.net/forum?id=uyTL5Bvosj} {Beyond the
  imitation game: Quantifying and extrapolating the capabilities of language
  models}.
\newblock \emph{Transactions on Machine Learning Research}.

\bibitem[{Borkan et~al.(2019)Borkan, Dixon, Sorensen, Thain, and
  Vasserman}]{DBLP:conf/www/BorkanDSTV19}
Daniel Borkan, Lucas Dixon, Jeffrey Sorensen, Nithum Thain, and Lucy Vasserman.
  2019.
\newblock \href {https://doi.org/10.1145/3308560.3317593} {Nuanced metrics for
  measuring unintended bias with real data for text classification}.
\newblock In \emph{Companion of The 2019 World Wide Web Conference, {WWW} 2019,
  San Francisco, CA, USA, May 13-17, 2019}, pages 491--500. {ACM}.

\bibitem[{Brown et~al.(2020)Brown, Mann, Ryder, Subbiah, Kaplan, Dhariwal,
  Neelakantan, Shyam, Sastry, Askell, Agarwal, Herbert{-}Voss, Krueger,
  Henighan, Child, Ramesh, Ziegler, Wu, Winter, Hesse, Chen, Sigler, Litwin,
  Gray, Chess, Clark, Berner, McCandlish, Radford, Sutskever, and
  Amodei}]{DBLP:conf/nips/BrownMRSKDNSSAA20}
Tom~B. Brown, Benjamin Mann, Nick Ryder, Melanie Subbiah, Jared Kaplan,
  Prafulla Dhariwal, Arvind Neelakantan, Pranav Shyam, Girish Sastry, Amanda
  Askell, Sandhini Agarwal, Ariel Herbert{-}Voss, Gretchen Krueger, Tom
  Henighan, Rewon Child, Aditya Ramesh, Daniel~M. Ziegler, Jeffrey Wu, Clemens
  Winter, Christopher Hesse, Mark Chen, Eric Sigler, Mateusz Litwin, Scott
  Gray, Benjamin Chess, Jack Clark, Christopher Berner, Sam McCandlish, Alec
  Radford, Ilya Sutskever, and Dario Amodei. 2020.
\newblock \href
  {https://proceedings.neurips.cc/paper/2020/hash/1457c0d6bfcb4967418bfb8ac142f64a-Abstract.html}
  {Language models are few-shot learners}.
\newblock In \emph{Advances in Neural Information Processing Systems 33: Annual
  Conference on Neural Information Processing Systems 2020, NeurIPS 2020,
  December 6-12, 2020, virtual}.

\bibitem[{Buchanan et~al.(2021)Buchanan, Lohn, Musser, and
  Sedova}]{Buchanan2021TruthLA}
Ben Buchanan, Andrew Lohn, Micah Musser, and Katerina Sedova. 2021.
\newblock \href {https://doi.org/10.51593/2021CA003} {Truth, lies, and
  automation: How language models could change disinformation}.

\bibitem[{Carlini et~al.(2021)Carlini, Tram{\`{e}}r, Wallace, Jagielski,
  Herbert{-}Voss, Lee, Roberts, Brown, Song, Erlingsson, Oprea, and
  Raffel}]{DBLP:conf/uss/CarliniTWJHLRBS21}
Nicholas Carlini, Florian Tram{\`{e}}r, Eric Wallace, Matthew Jagielski, Ariel
  Herbert{-}Voss, Katherine Lee, Adam Roberts, Tom~B. Brown, Dawn Song,
  {\'{U}}lfar Erlingsson, Alina Oprea, and Colin Raffel. 2021.
\newblock \href
  {https://www.usenix.org/conference/usenixsecurity21/presentation/carlini-extracting}
  {Extracting training data from large language models}.
\newblock In \emph{30th {USENIX} Security Symposium, {USENIX} Security 2021,
  August 11-13, 2021}, pages 2633--2650. {USENIX} Association.

\bibitem[{Che et~al.(2021)Che, Feng, Qin, and Liu}]{che-etal-2021-n}
Wanxiang Che, Yunlong Feng, Libo Qin, and Ting Liu. 2021.
\newblock \href {https://doi.org/10.18653/v1/2021.emnlp-demo.6} {N-{LTP}: An
  open-source neural language technology platform for {C}hinese}.
\newblock In \emph{Proceedings of the 2021 Conference on Empirical Methods in
  Natural Language Processing: System Demonstrations}, pages 42--49, Online and
  Punta Cana, Dominican Republic. Association for Computational Linguistics.

\bibitem[{Chen et~al.(2021)Chen, Tworek, Jun, Yuan, de~Oliveira~Pinto, Kaplan,
  Edwards, Burda, Joseph, Brockman, Ray, Puri, Krueger, Petrov, Khlaaf, Sastry,
  Mishkin, Chan, Gray, Ryder, Pavlov, Power, Kaiser, Bavarian, Winter, Tillet,
  Such, Cummings, Plappert, Chantzis, Barnes, Herbert{-}Voss, Guss, Nichol,
  Paino, Tezak, Tang, Babuschkin, Balaji, Jain, Saunders, Hesse, Carr, Leike,
  Achiam, Misra, Morikawa, Radford, Knight, Brundage, Murati, Mayer, Welinder,
  McGrew, Amodei, McCandlish, Sutskever, and
  Zaremba}]{DBLP:journals/corr/abs-2107-03374}
Mark Chen, Jerry Tworek, Heewoo Jun, Qiming Yuan, Henrique~Pond{\'{e}}
  de~Oliveira~Pinto, Jared Kaplan, Harrison Edwards, Yuri Burda, Nicholas
  Joseph, Greg Brockman, Alex Ray, Raul Puri, Gretchen Krueger, Michael Petrov,
  Heidy Khlaaf, Girish Sastry, Pamela Mishkin, Brooke Chan, Scott Gray, Nick
  Ryder, Mikhail Pavlov, Alethea Power, Lukasz Kaiser, Mohammad Bavarian,
  Clemens Winter, Philippe Tillet, Felipe~Petroski Such, Dave Cummings,
  Matthias Plappert, Fotios Chantzis, Elizabeth Barnes, Ariel Herbert{-}Voss,
  William~Hebgen Guss, Alex Nichol, Alex Paino, Nikolas Tezak, Jie Tang, Igor
  Babuschkin, Suchir Balaji, Shantanu Jain, William Saunders, Christopher
  Hesse, Andrew~N. Carr, Jan Leike, Joshua Achiam, Vedant Misra, Evan Morikawa,
  Alec Radford, Matthew Knight, Miles Brundage, Mira Murati, Katie Mayer, Peter
  Welinder, Bob McGrew, Dario Amodei, Sam McCandlish, Ilya Sutskever, and
  Wojciech Zaremba. 2021.
\newblock \href {http://arxiv.org/abs/2107.03374} {Evaluating large language
  models trained on code}.
\newblock \emph{CoRR}, abs/2107.03374.

\bibitem[{Chenghao~Fan and Tian(2023)}]{leng2023chinese-vicuna}
Zhenyi~Lu Chenghao~Fan and Jie Tian. 2023.
\newblock \href {https://github.com/Facico/Chinese-Vicuna} {Chinese-vicuna: A
  chinese instruction-following llama-based model}.

\bibitem[{Chiang et~al.(2023)Chiang, Li, Lin, Sheng, Wu, Zhang, Zheng, Zhuang,
  Zhuang, Gonzalez, Stoica, and Xing}]{vicuna2023}
Wei-Lin Chiang, Zhuohan Li, Zi~Lin, Ying Sheng, Zhanghao Wu, Hao Zhang, Lianmin
  Zheng, Siyuan Zhuang, Yonghao Zhuang, Joseph~E. Gonzalez, Ion Stoica, and
  Eric~P. Xing. 2023.
\newblock \href {https://lmsys.org/blog/2023-03-30-vicuna/} {Vicuna: An
  open-source chatbot impressing gpt-4 with 90\%* chatgpt quality}.

\bibitem[{Contributors(2023{\natexlab{a}})}]{2023flageval}
FlagEval Contributors. 2023{\natexlab{a}}.
\newblock Flageval.
\newblock \url{https://github.com/FlagOpen/FlagEval}.

\bibitem[{Contributors(2023{\natexlab{b}})}]{2023opencompass}
OpenCompass Contributors. 2023{\natexlab{b}}.
\newblock Opencompass: A universal evaluation platform for foundation models.
\newblock \url{https://github.com/InternLM/OpenCompass}.

\bibitem[{Cui et~al.(2023)Cui, Yang, and
  Yao}]{DBLP:journals/corr/abs-2304-08177}
Yiming Cui, Ziqing Yang, and Xin Yao. 2023.
\newblock \href {https://doi.org/10.48550/arXiv.2304.08177} {Efficient and
  effective text encoding for chinese llama and alpaca}.
\newblock \emph{CoRR}, abs/2304.08177.

\bibitem[{Deng et~al.(2022)Deng, Zhou, Sun, Zheng, Mi, Meng, and
  Huang}]{DBLP:conf/emnlp/DengZ0ZMMH22}
Jiawen Deng, Jingyan Zhou, Hao Sun, Chujie Zheng, Fei Mi, Helen Meng, and
  Minlie Huang. 2022.
\newblock \href {https://aclanthology.org/2022.emnlp-main.796} {{COLD:} {A}
  benchmark for chinese offensive language detection}.
\newblock In \emph{Proceedings of the 2022 Conference on Empirical Methods in
  Natural Language Processing, {EMNLP} 2022, Abu Dhabi, United Arab Emirates,
  December 7-11, 2022}, pages 11580--11599. Association for Computational
  Linguistics.

\bibitem[{Dhole et~al.(2021)Dhole, Gangal, Gehrmann, Gupta, Li, Mahamood,
  Mahendiran, Mille, Srivastava, Tan, Wu, Sohl-Dickstein, Choi, Hovy, Dusek,
  Ruder, Anand, Aneja, Banjade, Barthe, Behnke, Berlot-Attwell, Boyle, Brun,
  Cabezudo, Cahyawijaya, Chapuis, Che, Choudhary, Clauss, Colombo, Cornell,
  Dagan, Das, Dixit, Dopierre, Dray, Dubey, Ekeinhor, Giovanni, Gupta, Gupta,
  Hamla, Han, Harel-Canada, Honore, Jindal, Joniak, Kleyko, Kovatchev, Krishna,
  Kumar, Langer, Lee, Levinson, Liang, Liang, Liu, Lukyanenko, Marivate,
  de~Melo, Meoni, Meyer, Mir, Moosavi, Muennighoff, Mun, Murray, Namysl,
  Obedkova, Oli, Pasricha, Pfister, Plant, Prabhu, Pais, Qin, Raji, Rajpoot,
  Raunak, Rinberg, Roberts, Rodriguez, Roux, S., Sai, Schmidt, Scialom, Sefara,
  Shamsi, Shen, Shi, Shi, Shvets, Siegel, Sileo, Simon, Singh, Sitelew, Soni,
  Sorensen, Soto, Srivastava, Srivatsa, Sun, T, Tabassum, Tan, Teehan, Tiwari,
  Tolkiehn, Wang, Wang, Wang, Wang, Wei, Wilie, Winata, Wu, Wydmański, Xie,
  Yaseen, Yee, Zhang, and Zhang}]{dhole2021nlaugmenter}
Kaustubh~D. Dhole, Varun Gangal, Sebastian Gehrmann, Aadesh Gupta, Zhenhao Li,
  Saad Mahamood, Abinaya Mahendiran, Simon Mille, Ashish Srivastava, Samson
  Tan, Tongshuang Wu, Jascha Sohl-Dickstein, Jinho~D. Choi, Eduard Hovy, Ondrej
  Dusek, Sebastian Ruder, Sajant Anand, Nagender Aneja, Rabin Banjade, Lisa
  Barthe, Hanna Behnke, Ian Berlot-Attwell, Connor Boyle, Caroline Brun, Marco
  Antonio~Sobrevilla Cabezudo, Samuel Cahyawijaya, Emile Chapuis, Wanxiang Che,
  Mukund Choudhary, Christian Clauss, Pierre Colombo, Filip Cornell, Gautier
  Dagan, Mayukh Das, Tanay Dixit, Thomas Dopierre, Paul-Alexis Dray, Suchitra
  Dubey, Tatiana Ekeinhor, Marco~Di Giovanni, Rishabh Gupta, Rishabh Gupta,
  Louanes Hamla, Sang Han, Fabrice Harel-Canada, Antoine Honore, Ishan Jindal,
  Przemyslaw~K. Joniak, Denis Kleyko, Venelin Kovatchev, Kalpesh Krishna,
  Ashutosh Kumar, Stefan Langer, Seungjae~Ryan Lee, Corey~James Levinson,
  Hualou Liang, Kaizhao Liang, Zhexiong Liu, Andrey Lukyanenko, Vukosi
  Marivate, Gerard de~Melo, Simon Meoni, Maxime Meyer, Afnan Mir, Nafise~Sadat
  Moosavi, Niklas Muennighoff, Timothy Sum~Hon Mun, Kenton Murray, Marcin
  Namysl, Maria Obedkova, Priti Oli, Nivranshu Pasricha, Jan Pfister, Richard
  Plant, Vinay Prabhu, Vasile Pais, Libo Qin, Shahab Raji, Pawan~Kumar Rajpoot,
  Vikas Raunak, Roy Rinberg, Nicolas Roberts, Juan~Diego Rodriguez, Claude
  Roux, Vasconcellos P.~H. S., Ananya~B. Sai, Robin~M. Schmidt, Thomas Scialom,
  Tshephisho Sefara, Saqib~N. Shamsi, Xudong Shen, Haoyue Shi, Yiwen Shi, Anna
  Shvets, Nick Siegel, Damien Sileo, Jamie Simon, Chandan Singh, Roman Sitelew,
  Priyank Soni, Taylor Sorensen, William Soto, Aman Srivastava, KV~Aditya
  Srivatsa, Tony Sun, Mukund~Varma T, A~Tabassum, Fiona~Anting Tan, Ryan
  Teehan, Mo~Tiwari, Marie Tolkiehn, Athena Wang, Zijian Wang, Gloria Wang,
  Zijie~J. Wang, Fuxuan Wei, Bryan Wilie, Genta~Indra Winata, Xinyi Wu, Witold
  Wydmański, Tianbao Xie, Usama Yaseen, M.~Yee, Jing Zhang, and Yue Zhang.
  2021.
\newblock \href {http://arxiv.org/abs/2112.02721} {Nl-augmenter: A framework
  for task-sensitive natural language augmentation}.

\bibitem[{Ding et~al.(2022)Ding, Hu, Zhao, Chen, Liu, Zheng, and
  Sun}]{ding-etal-2022-openprompt}
Ning Ding, Shengding Hu, Weilin Zhao, Yulin Chen, Zhiyuan Liu, Haitao Zheng,
  and Maosong Sun. 2022.
\newblock \href {https://doi.org/10.18653/v1/2022.acl-demo.10} {{O}pen{P}rompt:
  An open-source framework for prompt-learning}.
\newblock In \emph{Proceedings of the 60th Annual Meeting of the Association
  for Computational Linguistics: System Demonstrations}, pages 105--113,
  Dublin, Ireland. Association for Computational Linguistics.

\bibitem[{Du et~al.(2022)Du, Qian, Liu, Ding, Qiu, Yang, and
  Tang}]{DBLP:conf/acl/DuQLDQY022}
Zhengxiao Du, Yujie Qian, Xiao Liu, Ming Ding, Jiezhong Qiu, Zhilin Yang, and
  Jie Tang. 2022.
\newblock \href {https://doi.org/10.18653/v1/2022.acl-long.26} {{GLM:} general
  language model pretraining with autoregressive blank infilling}.
\newblock In \emph{Proceedings of the 60th Annual Meeting of the Association
  for Computational Linguistics (Volume 1: Long Papers), {ACL} 2022, Dublin,
  Ireland, May 22-27, 2022}, pages 320--335. Association for Computational
  Linguistics.

\bibitem[{Duan(2018)}]{DBLP:conf/nlpcc/Duan18}
Nan Duan. 2018.
\newblock \href {https://doi.org/10.1007/978-3-319-99501-4\_43} {Overview of
  the {NLPCC} 2018 shared task: Open domain {QA}}.
\newblock In \emph{Natural Language Processing and Chinese Computing - 7th
  {CCF} International Conference, {NLPCC} 2018, Hohhot, China, August 26-30,
  2018, Proceedings, Part {II}}, volume 11109 of \emph{Lecture Notes in
  Computer Science}, pages 452--456. Springer.

\bibitem[{El{-}Yaniv and Wiener(2010)}]{DBLP:journals/jmlr/El-YanivW10}
Ran El{-}Yaniv and Yair Wiener. 2010.
\newblock \href {https://doi.org/10.5555/1756006.1859904} {On the foundations
  of noise-free selective classification}.
\newblock \emph{J. Mach. Learn. Res.}, 11:1605--1641.

\bibitem[{Fu and Khot(2022)}]{fu2022gptroadmap}
Hao Fu, Yao;~Peng and Tushar Khot. 2022.
\newblock \href
  {https://yaofu.notion.site/How-does-GPT-Obtain-its-Ability-Tracing-Emergent-Abilities-of-Language-Models-to-their-Sources-b9a57ac0fcf74f30a1ab9e3e36fa1dc1}
  {How does gpt obtain its ability? tracing emergent abilities of language
  models to their sources}.
\newblock \emph{Yao Fu’s Notion}.

\bibitem[{Gao et~al.(2021{\natexlab{a}})Gao, Biderman, Black, Golding, Hoppe,
  Foster, Phang, He, Thite, Nabeshima, Presser, and
  Leahy}]{DBLP:journals/corr/abs-2101-00027}
Leo Gao, Stella Biderman, Sid Black, Laurence Golding, Travis Hoppe, Charles
  Foster, Jason Phang, Horace He, Anish Thite, Noa Nabeshima, Shawn Presser,
  and Connor Leahy. 2021{\natexlab{a}}.
\newblock \href {http://arxiv.org/abs/2101.00027} {The pile: An 800gb dataset
  of diverse text for language modeling}.
\newblock \emph{CoRR}, abs/2101.00027.

\bibitem[{Gao et~al.(2021{\natexlab{b}})Gao, Fisch, and
  Chen}]{DBLP:conf/acl/GaoFC20}
Tianyu Gao, Adam Fisch, and Danqi Chen. 2021{\natexlab{b}}.
\newblock \href {https://doi.org/10.18653/v1/2021.acl-long.295} {Making
  pre-trained language models better few-shot learners}.
\newblock In \emph{Proceedings of the 59th Annual Meeting of the Association
  for Computational Linguistics and the 11th International Joint Conference on
  Natural Language Processing, {ACL/IJCNLP} 2021, (Volume 1: Long Papers),
  Virtual Event, August 1-6, 2021}, pages 3816--3830. Association for
  Computational Linguistics.

\bibitem[{Gupta et~al.(2022)Gupta, Wu, Liu, and
  Xiong}]{DBLP:conf/acl/GuptaWLX22}
Prakhar Gupta, Chien{-}Sheng Wu, Wenhao Liu, and Caiming Xiong. 2022.
\newblock \href {https://doi.org/10.18653/v1/2022.acl-long.263} {Dialfact: {A}
  benchmark for fact-checking in dialogue}.
\newblock In \emph{Proceedings of the 60th Annual Meeting of the Association
  for Computational Linguistics (Volume 1: Long Papers), {ACL} 2022, Dublin,
  Ireland, May 22-27, 2022}, pages 3785--3801. Association for Computational
  Linguistics.

\bibitem[{Hendrycks et~al.(2021)Hendrycks, Burns, Basart, Zou, Mazeika, Song,
  and Steinhardt}]{DBLP:conf/iclr/HendrycksBBZMSS21}
Dan Hendrycks, Collin Burns, Steven Basart, Andy Zou, Mantas Mazeika, Dawn
  Song, and Jacob Steinhardt. 2021.
\newblock \href {https://openreview.net/forum?id=d7KBjmI3GmQ} {Measuring
  massive multitask language understanding}.
\newblock In \emph{9th International Conference on Learning Representations,
  {ICLR} 2021, Virtual Event, Austria, May 3-7, 2021}. OpenReview.net.

\bibitem[{Hu et~al.(2020)Hu, Richardson, Xu, Li, K{\"{u}}bler, and
  Moss}]{DBLP:journals/corr/abs-2010-05444}
Hai Hu, Kyle Richardson, Liang Xu, Lu~Li, Sandra K{\"{u}}bler, and Lawrence~S.
  Moss. 2020.
\newblock \href {http://arxiv.org/abs/2010.05444} {{OCNLI:} original chinese
  natural language inference}.
\newblock \emph{CoRR}, abs/2010.05444.

\bibitem[{Hu et~al.(2022)Hu, Guo, Wu, Liu, Wen, and Yu}]{hu-etal-2022-chef}
Xuming Hu, Zhijiang Guo, GuanYu Wu, Aiwei Liu, Lijie Wen, and Philip Yu. 2022.
\newblock \href {https://doi.org/10.18653/v1/2022.naacl-main.246} {{CHEF}: A
  pilot {C}hinese dataset for evidence-based fact-checking}.
\newblock In \emph{Proceedings of the 2022 Conference of the North American
  Chapter of the Association for Computational Linguistics: Human Language
  Technologies}, pages 3362--3376, Seattle, United States. Association for
  Computational Linguistics.

\bibitem[{Huang et~al.(2023)Huang, Bai, Zhu, Zhang, Zhang, Su, Liu, Lv, Zhang,
  Lei, Fu, Sun, and He}]{DBLP:journals/corr/abs-2305-08322}
Yuzhen Huang, Yuzhuo Bai, Zhihao Zhu, Junlei Zhang, Jinghan Zhang, Tangjun Su,
  Junteng Liu, Chuancheng Lv, Yikai Zhang, Jiayi Lei, Yao Fu, Maosong Sun, and
  Junxian He. 2023.
\newblock \href {https://doi.org/10.48550/arXiv.2305.08322} {C-eval: {A}
  multi-level multi-discipline chinese evaluation suite for foundation models}.
\newblock \emph{CoRR}, abs/2305.08322.

\bibitem[{Ji et~al.(2023)Ji, Deng, Gong, Peng, Niu, Zhang, Ma, and
  Li}]{DBLP:journals/corr/abs-2303-14742}
Yunjie Ji, Yong Deng, Yan Gong, Yiping Peng, Qiang Niu, Lei Zhang, Baochang Ma,
  and Xiangang Li. 2023.
\newblock \href {https://doi.org/10.48550/arXiv.2303.14742} {Exploring the
  impact of instruction data scaling on large language models: An empirical
  study on real-world use cases}.
\newblock \emph{CoRR}, abs/2303.14742.

\bibitem[{Kocmi et~al.(2022)Kocmi, Bawden, Bojar, Dvorkovich, Federmann,
  Fishel, Gowda, Graham, Grundkiewicz, Haddow, Knowles, Koehn, Monz, Morishita,
  Nagata, Nakazawa, Nov{\'{a}}k, Popel, and
  Popovic}]{DBLP:conf/wmt/KocmiBBDFFGGGHKKMMNNNPP22}
Tom Kocmi, Rachel Bawden, Ondrej Bojar, Anton Dvorkovich, Christian Federmann,
  Mark Fishel, Thamme Gowda, Yvette Graham, Roman Grundkiewicz, Barry Haddow,
  Rebecca Knowles, Philipp Koehn, Christof Monz, Makoto Morishita, Masaaki
  Nagata, Toshiaki Nakazawa, Michal Nov{\'{a}}k, Martin Popel, and Maja
  Popovic. 2022.
\newblock \href {https://aclanthology.org/2022.wmt-1.1} {Findings of the 2022
  conference on machine translation {(WMT22)}}.
\newblock In \emph{Proceedings of the Seventh Conference on Machine
  Translation, {WMT} 2022, Abu Dhabi, United Arab Emirates (Hybrid), December
  7-8, 2022}, pages 1--45. Association for Computational Linguistics.

\bibitem[{Kojima et~al.(2022)Kojima, Gu, Reid, Matsuo, and
  Iwasawa}]{DBLP:conf/nips/KojimaGRMI22}
Takeshi Kojima, Shixiang~Shane Gu, Machel Reid, Yutaka Matsuo, and Yusuke
  Iwasawa. 2022.
\newblock \href
  {http://papers.nips.cc/paper\_files/paper/2022/hash/8bb0d291acd4acf06ef112099c16f326-Abstract-Conference.html}
  {Large language models are zero-shot reasoners}.
\newblock In \emph{NeurIPS}.

\bibitem[{Kumar et~al.(2019)Kumar, Liang, and Ma}]{DBLP:conf/nips/KumarLM19}
Ananya Kumar, Percy Liang, and Tengyu Ma. 2019.
\newblock \href
  {https://proceedings.neurips.cc/paper/2019/hash/f8c0c968632845cd133308b1a494967f-Abstract.html}
  {Verified uncertainty calibration}.
\newblock In \emph{Advances in Neural Information Processing Systems 32: Annual
  Conference on Neural Information Processing Systems 2019, NeurIPS 2019,
  December 8-14, 2019, Vancouver, BC, Canada}, pages 3787--3798.

\bibitem[{Lees et~al.(2022)Lees, Tran, Tay, Sorensen, Gupta, Metzler, and
  Vasserman}]{DBLP:conf/kdd/Lees0TSGMV22}
Alyssa Lees, Vinh~Q. Tran, Yi~Tay, Jeffrey Sorensen, Jai~Prakash Gupta, Donald
  Metzler, and Lucy Vasserman. 2022.
\newblock \href {https://doi.org/10.1145/3534678.3539147} {A new generation of
  perspective {API:} efficient multilingual character-level transformers}.
\newblock In \emph{{KDD} '22: The 28th {ACM} {SIGKDD} Conference on Knowledge
  Discovery and Data Mining, Washington, DC, USA, August 14 - 18, 2022}, pages
  3197--3207. {ACM}.

\bibitem[{Li et~al.(2023)Li, Zhang, Koto, Yang, Zhao, Gong, Duan, and
  Baldwin}]{DBLP:journals/corr/abs-2306-09212}
Haonan Li, Yixuan Zhang, Fajri Koto, Yifei Yang, Hai Zhao, Yeyun Gong, Nan
  Duan, and Timothy Baldwin. 2023.
\newblock \href {https://doi.org/10.48550/arXiv.2306.09212} {{CMMLU:} measuring
  massive multitask language understanding in chinese}.
\newblock \emph{CoRR}, abs/2306.09212.

\bibitem[{Li et~al.(2016)Li, Galley, Brockett, Gao, and
  Dolan}]{DBLP:conf/naacl/LiGBGD16}
Jiwei Li, Michel Galley, Chris Brockett, Jianfeng Gao, and Bill Dolan. 2016.
\newblock \href {https://doi.org/10.18653/v1/n16-1014} {A diversity-promoting
  objective function for neural conversation models}.
\newblock In \emph{{NAACL} {HLT} 2016, The 2016 Conference of the North
  American Chapter of the Association for Computational Linguistics: Human
  Language Technologies, San Diego California, USA, June 12-17, 2016}, pages
  110--119. The Association for Computational Linguistics.

\bibitem[{Li et~al.(2021)Li, Qi, Sun, Yi, and
  Zhang}]{DBLP:journals/corr/abs-2106-01979}
Wenhao Li, Fanchao Qi, Maosong Sun, Xiaoyuan Yi, and Jiarui Zhang. 2021.
\newblock \href {http://arxiv.org/abs/2106.01979} {{CCPM:} {A} chinese
  classical poetry matching dataset}.
\newblock \emph{CoRR}, abs/2106.01979.

\bibitem[{Liang et~al.(2022)Liang, Bommasani, Lee, Tsipras, Soylu, Yasunaga,
  Zhang, Narayanan, Wu, Kumar, Newman, Yuan, Yan, Zhang, Cosgrove, Manning,
  R{\'{e}}, Acosta{-}Navas, Hudson, Zelikman, Durmus, Ladhak, Rong, Ren, Yao,
  Wang, Santhanam, Orr, Zheng, Y{\"{u}}ksekg{\"{o}}n{\"{u}}l, Suzgun, Kim,
  Guha, Chatterji, Khattab, Henderson, Huang, Chi, Xie, Santurkar, Ganguli,
  Hashimoto, Icard, Zhang, Chaudhary, Wang, Li, Mai, Zhang, and
  Koreeda}]{DBLP:journals/corr/abs-2211-09110}
Percy Liang, Rishi Bommasani, Tony Lee, Dimitris Tsipras, Dilara Soylu,
  Michihiro Yasunaga, Yian Zhang, Deepak Narayanan, Yuhuai Wu, Ananya Kumar,
  Benjamin Newman, Binhang Yuan, Bobby Yan, Ce~Zhang, Christian Cosgrove,
  Christopher~D. Manning, Christopher R{\'{e}}, Diana Acosta{-}Navas, Drew~A.
  Hudson, Eric Zelikman, Esin Durmus, Faisal Ladhak, Frieda Rong, Hongyu Ren,
  Huaxiu Yao, Jue Wang, Keshav Santhanam, Laurel~J. Orr, Lucia Zheng, Mert
  Y{\"{u}}ksekg{\"{o}}n{\"{u}}l, Mirac Suzgun, Nathan Kim, Neel Guha,
  Niladri~S. Chatterji, Omar Khattab, Peter Henderson, Qian Huang, Ryan Chi,
  Sang~Michael Xie, Shibani Santurkar, Surya Ganguli, Tatsunori Hashimoto,
  Thomas Icard, Tianyi Zhang, Vishrav Chaudhary, William Wang, Xuechen Li,
  Yifan Mai, Yuhui Zhang, and Yuta Koreeda. 2022.
\newblock \href {https://doi.org/10.48550/arXiv.2211.09110} {Holistic
  evaluation of language models}.
\newblock \emph{CoRR}, abs/2211.09110.

\bibitem[{Lin(2004)}]{lin-2004-rouge}
Chin-Yew Lin. 2004.
\newblock \href {https://aclanthology.org/W04-1013} {{ROUGE}: A package for
  automatic evaluation of summaries}.
\newblock In \emph{Text Summarization Branches Out}, pages 74--81, Barcelona,
  Spain. Association for Computational Linguistics.

\bibitem[{Lin et~al.(2021)Lin, Ma, Zhu, Xiang, Zhou, Zhang, and
  Zong}]{lin-etal-2021-csds}
Haitao Lin, Liqun Ma, Junnan Zhu, Lu~Xiang, Yu~Zhou, Jiajun Zhang, and
  Chengqing Zong. 2021.
\newblock \href {https://doi.org/10.18653/v1/2021.emnlp-main.365} {{CSDS}: A
  fine-grained {C}hinese dataset for customer service dialogue summarization}.
\newblock In \emph{Proceedings of the 2021 Conference on Empirical Methods in
  Natural Language Processing}, pages 4436--4451, Online and Punta Cana,
  Dominican Republic. Association for Computational Linguistics.

\bibitem[{Lin et~al.(2022)Lin, Hilton, and Evans}]{DBLP:conf/acl/LinHE22}
Stephanie Lin, Jacob Hilton, and Owain Evans. 2022.
\newblock \href {https://doi.org/10.18653/v1/2022.acl-long.229} {Truthfulqa:
  Measuring how models mimic human falsehoods}.
\newblock In \emph{Proceedings of the 60th Annual Meeting of the Association
  for Computational Linguistics (Volume 1: Long Papers), {ACL} 2022, Dublin,
  Ireland, May 22-27, 2022}, pages 3214--3252. Association for Computational
  Linguistics.

\bibitem[{Liu et~al.(2023)Liu, Jin, Ren, Yu, Dong, Peng, Zhang, Peng, Zhang,
  Lyu, Su, Liu, and Xiong}]{DBLP:journals/corr/abs-2305-10263}
Chuang Liu, Renren Jin, Yuqi Ren, Linhao Yu, Tianyu Dong, Xiaohan Peng, Shuting
  Zhang, Jianxiang Peng, Peiyi Zhang, Qingqing Lyu, Xiaowen Su, Qun Liu, and
  Deyi Xiong. 2023.
\newblock \href {https://doi.org/10.48550/arXiv.2305.10263} {{M3KE:} {A}
  massive multi-level multi-subject knowledge evaluation benchmark for chinese
  large language models}.
\newblock \emph{CoRR}, abs/2305.10263.

\bibitem[{Liu et~al.(2012)Liu, Xu, and Zhao}]{DBLP:conf/emnlp/LiuXZ12}
Kang Liu, Liheng Xu, and Jun Zhao. 2012.
\newblock \href {https://aclanthology.org/D12-1123/} {Opinion target extraction
  using word-based translation model}.
\newblock In \emph{Proceedings of the 2012 Joint Conference on Empirical
  Methods in Natural Language Processing and Computational Natural Language
  Learning, EMNLP-CoNLL 2012, July 12-14, 2012, Jeju Island, Korea}, pages
  1346--1356. {ACL}.

\bibitem[{Lu et~al.(2023)Lu, Qiu, Yu, Welleck, and
  Chang}]{DBLP:conf/acl/Lu00WC23}
Pan Lu, Liang Qiu, Wenhao Yu, Sean Welleck, and Kai{-}Wei Chang. 2023.
\newblock \href {https://aclanthology.org/2023.acl-long.817} {A survey of deep
  learning for mathematical reasoning}.
\newblock In \emph{Proceedings of the 61st Annual Meeting of the Association
  for Computational Linguistics (Volume 1: Long Papers), {ACL} 2023, Toronto,
  Canada, July 9-14, 2023}, pages 14605--14631. Association for Computational
  Linguistics.

\bibitem[{McKenzie et~al.(2023)McKenzie, Lyzhov, Pieler, Parrish, Mueller,
  Prabhu, McLean, Kirtland, Ross, Liu, Gritsevskiy, Wurgaft, Kauffman, Recchia,
  Liu, Cavanagh, Weiss, Huang, Droid, Tseng, Korbak, Shen, Zhang, Zhou, Kim,
  Bowman, and Perez}]{DBLP:journals/corr/abs-2306-09479}
Ian~R. McKenzie, Alexander Lyzhov, Michael Pieler, Alicia Parrish, Aaron
  Mueller, Ameya Prabhu, Euan McLean, Aaron Kirtland, Alexis Ross, Alisa Liu,
  Andrew Gritsevskiy, Daniel Wurgaft, Derik Kauffman, Gabriel Recchia, Jiacheng
  Liu, Joe Cavanagh, Max Weiss, Sicong Huang, The~Floating Droid, Tom Tseng,
  Tomasz Korbak, Xudong Shen, Yuhui Zhang, Zhengping Zhou, Najoung Kim,
  Samuel~R. Bowman, and Ethan Perez. 2023.
\newblock \href {https://doi.org/10.48550/arXiv.2306.09479} {Inverse scaling:
  When bigger isn't better}.
\newblock \emph{CoRR}, abs/2306.09479.

\bibitem[{Miller et~al.(2017)Miller, Feng, Batra, Bordes, Fisch, Lu, Parikh,
  and Weston}]{DBLP:conf/emnlp/MillerFBBFLPW17}
Alexander~H. Miller, Will Feng, Dhruv Batra, Antoine Bordes, Adam Fisch, Jiasen
  Lu, Devi Parikh, and Jason Weston. 2017.
\newblock \href {https://doi.org/10.18653/v1/d17-2014} {Parlai: {A} dialog
  research software platform}.
\newblock In \emph{Proceedings of the 2017 Conference on Empirical Methods in
  Natural Language Processing, {EMNLP} 2017, Copenhagen, Denmark, September
  9-11, 2017 - System Demonstrations}, pages 79--84. Association for
  Computational Linguistics.

\bibitem[{Muennighoff et~al.(2023)Muennighoff, Wang, Sutawika, Roberts,
  Biderman, Scao, Bari, Shen, Yong, Schoelkopf, Tang, Radev, Aji, Almubarak,
  Albanie, Alyafeai, Webson, Raff, and
  Raffel}]{DBLP:conf/acl/MuennighoffWSRB23}
Niklas Muennighoff, Thomas Wang, Lintang Sutawika, Adam Roberts, Stella
  Biderman, Teven~Le Scao, M.~Saiful Bari, Sheng Shen, Zheng~Xin Yong, Hailey
  Schoelkopf, Xiangru Tang, Dragomir Radev, Alham~Fikri Aji, Khalid Almubarak,
  Samuel Albanie, Zaid Alyafeai, Albert Webson, Edward Raff, and Colin Raffel.
  2023.
\newblock \href {https://aclanthology.org/2023.acl-long.891} {Crosslingual
  generalization through multitask finetuning}.
\newblock In \emph{Proceedings of the 61st Annual Meeting of the Association
  for Computational Linguistics (Volume 1: Long Papers), {ACL} 2023, Toronto,
  Canada, July 9-14, 2023}, pages 15991--16111. Association for Computational
  Linguistics.

\bibitem[{Naeini et~al.(2015)Naeini, Cooper, and
  Hauskrecht}]{DBLP:conf/aaai/NaeiniCH15}
Mahdi~Pakdaman Naeini, Gregory~F. Cooper, and Milos Hauskrecht. 2015.
\newblock \href {http://www.aaai.org/ocs/index.php/AAAI/AAAI15/paper/view/9667}
  {Obtaining well calibrated probabilities using bayesian binning}.
\newblock In \emph{Proceedings of the Twenty-Ninth {AAAI} Conference on
  Artificial Intelligence, January 25-30, 2015, Austin, Texas, {USA}}, pages
  2901--2907. {AAAI} Press.

\bibitem[{Nijkamp et~al.(2023)Nijkamp, Pang, Hayashi, Tu, Wang, Zhou, Savarese,
  and Xiong}]{DBLP:conf/iclr/NijkampPHTWZSX23}
Erik Nijkamp, Bo~Pang, Hiroaki Hayashi, Lifu Tu, Huan Wang, Yingbo Zhou, Silvio
  Savarese, and Caiming Xiong. 2023.
\newblock \href {https://openreview.net/pdf?id=iaYcJKpY2B\_} {Codegen: An open
  large language model for code with multi-turn program synthesis}.
\newblock In \emph{The Eleventh International Conference on Learning
  Representations, {ICLR} 2023, Kigali, Rwanda, May 1-5, 2023}. OpenReview.net.

\bibitem[{OpenAI(2023)}]{DBLP:journals/corr/abs-2303-08774}
OpenAI. 2023.
\newblock \href {https://doi.org/10.48550/arXiv.2303.08774} {{GPT-4} technical
  report}.
\newblock \emph{CoRR}, abs/2303.08774.

\bibitem[{Ouyang et~al.(2022)Ouyang, Wu, Jiang, Almeida, Wainwright, Mishkin,
  Zhang, Agarwal, Slama, Ray, Schulman, Hilton, Kelton, Miller, Simens, Askell,
  Welinder, Christiano, Leike, and Lowe}]{DBLP:conf/nips/Ouyang0JAWMZASR22}
Long Ouyang, Jeffrey Wu, Xu~Jiang, Diogo Almeida, Carroll~L. Wainwright, Pamela
  Mishkin, Chong Zhang, Sandhini Agarwal, Katarina Slama, Alex Ray, John
  Schulman, Jacob Hilton, Fraser Kelton, Luke Miller, Maddie Simens, Amanda
  Askell, Peter Welinder, Paul~F. Christiano, Jan Leike, and Ryan Lowe. 2022.
\newblock \href
  {http://papers.nips.cc/paper\_files/paper/2022/hash/b1efde53be364a73914f58805a001731-Abstract-Conference.html}
  {Training language models to follow instructions with human feedback}.
\newblock In \emph{NeurIPS}.

\bibitem[{Papineni et~al.(2002)Papineni, Roukos, Ward, and
  Zhu}]{papineni-etal-2002-bleu}
Kishore Papineni, Salim Roukos, Todd Ward, and Wei-Jing Zhu. 2002.
\newblock \href {https://doi.org/10.3115/1073083.1073135} {{B}leu: a method for
  automatic evaluation of machine translation}.
\newblock In \emph{Proceedings of the 40th Annual Meeting of the Association
  for Computational Linguistics}, pages 311--318, Philadelphia, Pennsylvania,
  USA. Association for Computational Linguistics.

\bibitem[{Patel and Pavlick(2022)}]{DBLP:conf/iclr/PatelP22}
Roma Patel and Ellie Pavlick. 2022.
\newblock \href {https://openreview.net/forum?id=gJcEM8sxHK} {Mapping language
  models to grounded conceptual spaces}.
\newblock In \emph{The Tenth International Conference on Learning
  Representations, {ICLR} 2022, Virtual Event, April 25-29, 2022}.
  OpenReview.net.

\bibitem[{Perez et~al.(2021)Perez, Kiela, and Cho}]{DBLP:conf/nips/PerezKC21}
Ethan Perez, Douwe Kiela, and Kyunghyun Cho. 2021.
\newblock \href
  {https://proceedings.neurips.cc/paper/2021/hash/5c04925674920eb58467fb52ce4ef728-Abstract.html}
  {True few-shot learning with language models}.
\newblock In \emph{Advances in Neural Information Processing Systems 34: Annual
  Conference on Neural Information Processing Systems 2021, NeurIPS 2021,
  December 6-14, 2021, virtual}, pages 11054--11070.

\bibitem[{Petroni et~al.(2019)Petroni, Rockt{\"{a}}schel, Riedel, Lewis,
  Bakhtin, Wu, and Miller}]{DBLP:conf/emnlp/PetroniRRLBWM19}
Fabio Petroni, Tim Rockt{\"{a}}schel, Sebastian Riedel, Patrick S.~H. Lewis,
  Anton Bakhtin, Yuxiang Wu, and Alexander~H. Miller. 2019.
\newblock \href {https://doi.org/10.18653/v1/D19-1250} {Language models as
  knowledge bases?}
\newblock In \emph{Proceedings of the 2019 Conference on Empirical Methods in
  Natural Language Processing and the 9th International Joint Conference on
  Natural Language Processing, {EMNLP-IJCNLP} 2019, Hong Kong, China, November
  3-7, 2019}, pages 2463--2473. Association for Computational Linguistics.

\bibitem[{Post(2018)}]{DBLP:conf/wmt/Post18}
Matt Post. 2018.
\newblock \href {https://doi.org/10.18653/v1/w18-6319} {A call for clarity in
  reporting {BLEU} scores}.
\newblock In \emph{Proceedings of the Third Conference on Machine Translation:
  Research Papers, {WMT} 2018, Belgium, Brussels, October 31 - November 1,
  2018}, pages 186--191. Association for Computational Linguistics.

\bibitem[{Puduppully et~al.(2019)Puduppully, Dong, and
  Lapata}]{DBLP:conf/aaai/Puduppully0L19}
Ratish Puduppully, Li~Dong, and Mirella Lapata. 2019.
\newblock \href {https://doi.org/10.1609/aaai.v33i01.33016908} {Data-to-text
  generation with content selection and planning}.
\newblock In \emph{The Thirty-Third {AAAI} Conference on Artificial
  Intelligence, {AAAI} 2019, The Thirty-First Innovative Applications of
  Artificial Intelligence Conference, {IAAI} 2019, The Ninth {AAAI} Symposium
  on Educational Advances in Artificial Intelligence, {EAAI} 2019, Honolulu,
  Hawaii, USA, January 27 - February 1, 2019}, pages 6908--6915. {AAAI} Press.

\bibitem[{Sanh et~al.(2022)Sanh, Webson, Raffel, Bach, Sutawika, Alyafeai,
  Chaffin, Stiegler, Raja, Dey, Bari, Xu, Thakker, Sharma, Szczechla, Kim,
  Chhablani, Nayak, Datta, Chang, Jiang, Wang, Manica, Shen, Yong, Pandey,
  Bawden, Wang, Neeraj, Rozen, Sharma, Santilli, F{\'{e}}vry, Fries, Teehan,
  Scao, Biderman, Gao, Wolf, and Rush}]{DBLP:conf/iclr/T0pp}
Victor Sanh, Albert Webson, Colin Raffel, Stephen~H. Bach, Lintang Sutawika,
  Zaid Alyafeai, Antoine Chaffin, Arnaud Stiegler, Arun Raja, Manan Dey,
  M~Saiful Bari, Canwen Xu, Urmish Thakker, Shanya~Sharma Sharma, Eliza
  Szczechla, Taewoon Kim, Gunjan Chhablani, Nihal~V. Nayak, Debajyoti Datta,
  Jonathan Chang, Mike~Tian{-}Jian Jiang, Han Wang, Matteo Manica, Sheng Shen,
  Zheng~Xin Yong, Harshit Pandey, Rachel Bawden, Thomas Wang, Trishala Neeraj,
  Jos Rozen, Abheesht Sharma, Andrea Santilli, Thibault F{\'{e}}vry, Jason~Alan
  Fries, Ryan Teehan, Teven~Le Scao, Stella Biderman, Leo Gao, Thomas Wolf, and
  Alexander~M. Rush. 2022.
\newblock \href {https://openreview.net/forum?id=9Vrb9D0WI4} {Multitask
  prompted training enables zero-shot task generalization}.
\newblock In \emph{The Tenth International Conference on Learning
  Representations, {ICLR} 2022, Virtual Event, April 25-29, 2022}.
  OpenReview.net.

\bibitem[{Scao et~al.(2022)Scao, Fan, Akiki, Pavlick, Ilic, Hesslow,
  Castagn{\'{e}}, Luccioni, Yvon, Gall{\'{e}}, Tow, Rush, Biderman, Webson,
  Ammanamanchi, Wang, Sagot, Muennighoff, del Moral, Ruwase, Bawden, Bekman,
  McMillan{-}Major, Beltagy, Nguyen, Saulnier, Tan, Suarez, Sanh,
  Lauren{\c{c}}on, Jernite, Launay, Mitchell, Raffel, Gokaslan, Simhi, Soroa,
  Aji, Alfassy, Rogers, Nitzav, Xu, Mou, Emezue, Klamm, Leong, van Strien,
  Adelani, and et~al.}]{DBLP:journals/corr/abs-2211-05100}
Teven~Le Scao, Angela Fan, Christopher Akiki, Ellie Pavlick, Suzana Ilic,
  Daniel Hesslow, Roman Castagn{\'{e}}, Alexandra~Sasha Luccioni,
  Fran{\c{c}}ois Yvon, Matthias Gall{\'{e}}, Jonathan Tow, Alexander~M. Rush,
  Stella Biderman, Albert Webson, Pawan~Sasanka Ammanamanchi, Thomas Wang,
  Beno{\^{\i}}t Sagot, Niklas Muennighoff, Albert~Villanova del Moral, Olatunji
  Ruwase, Rachel Bawden, Stas Bekman, Angelina McMillan{-}Major, Iz~Beltagy,
  Huu Nguyen, Lucile Saulnier, Samson Tan, Pedro~Ortiz Suarez, Victor Sanh,
  Hugo Lauren{\c{c}}on, Yacine Jernite, Julien Launay, Margaret Mitchell, Colin
  Raffel, Aaron Gokaslan, Adi Simhi, Aitor Soroa, Alham~Fikri Aji, Amit
  Alfassy, Anna Rogers, Ariel~Kreisberg Nitzav, Canwen Xu, Chenghao Mou, Chris
  Emezue, Christopher Klamm, Colin Leong, Daniel van Strien, David~Ifeoluwa
  Adelani, and et~al. 2022.
\newblock \href {https://doi.org/10.48550/arXiv.2211.05100} {{BLOOM:} {A}
  176b-parameter open-access multilingual language model}.
\newblock \emph{CoRR}, abs/2211.05100.

\bibitem[{Schick et~al.(2023)Schick, Dwivedi{-}Yu, Dess{\`{\i}}, Raileanu,
  Lomeli, Zettlemoyer, Cancedda, and
  Scialom}]{DBLP:journals/corr/abs-2302-04761}
Timo Schick, Jane Dwivedi{-}Yu, Roberto Dess{\`{\i}}, Roberta Raileanu, Maria
  Lomeli, Luke Zettlemoyer, Nicola Cancedda, and Thomas Scialom. 2023.
\newblock \href {https://doi.org/10.48550/arXiv.2302.04761} {Toolformer:
  Language models can teach themselves to use tools}.
\newblock \emph{CoRR}, abs/2302.04761.

\bibitem[{Shao et~al.(2019)Shao, Huang, Wen, Xu, and
  Zhu}]{DBLP:conf/emnlp/ShaoHWXZ19}
Zhihong Shao, Minlie Huang, Jiangtao Wen, Wenfei Xu, and Xiaoyan Zhu. 2019.
\newblock \href {https://doi.org/10.18653/v1/D19-1321} {Long and diverse text
  generation with planning-based hierarchical variational model}.
\newblock In \emph{Proceedings of the 2019 Conference on Empirical Methods in
  Natural Language Processing and the 9th International Joint Conference on
  Natural Language Processing, {EMNLP-IJCNLP} 2019, Hong Kong, China, November
  3-7, 2019}, pages 3255--3266. Association for Computational Linguistics.

\bibitem[{Storks et~al.(2019)Storks, Gao, and
  Chai}]{DBLP:journals/corr/abs-1904-01172}
Shane Storks, Qiaozi Gao, and Joyce~Y. Chai. 2019.
\newblock \href {http://arxiv.org/abs/1904.01172} {Commonsense reasoning for
  natural language understanding: {A} survey of benchmarks, resources, and
  approaches}.
\newblock \emph{CoRR}, abs/1904.01172.

\bibitem[{Sun et~al.(2023{\natexlab{a}})Sun, Zhang, Deng, Cheng, and
  Huang}]{DBLP:journals/corr/abs-2304-10436}
Hao Sun, Zhexin Zhang, Jiawen Deng, Jiale Cheng, and Minlie Huang.
  2023{\natexlab{a}}.
\newblock \href {https://doi.org/10.48550/arXiv.2304.10436} {Safety assessment
  of chinese large language models}.
\newblock \emph{CoRR}, abs/2304.10436.

\bibitem[{Sun and Zhou(2012)}]{DBLP:conf/acl/SunZ12}
Hong Sun and Ming Zhou. 2012.
\newblock \href {https://aclanthology.org/P12-2008/} {Joint learning of a dual
  {SMT} system for paraphrase generation}.
\newblock In \emph{The 50th Annual Meeting of the Association for Computational
  Linguistics, Proceedings of the Conference, July 8-14, 2012, Jeju Island,
  Korea - Volume 2: Short Papers}, pages 38--42. The Association for Computer
  Linguistics.

\bibitem[{Sun et~al.(2019)Sun, Yu, Yu, and
  Cardie}]{DBLP:journals/corr/abs-1904-09679}
Kai Sun, Dian Yu, Dong Yu, and Claire Cardie. 2019.
\newblock \href {http://arxiv.org/abs/1904.09679} {Probing prior knowledge
  needed in challenging chinese machine reading comprehension}.
\newblock \emph{CoRR}, abs/1904.09679.

\bibitem[{Sun et~al.(2023{\natexlab{b}})Sun, Zhang, He, Li, Cheng, Yan, Liu,
  Shao, Tang, Zhao, Chen, Zheng, Zhou, Li, Zhan, Zhou, Li, Yang, Wu, Yin,
  Huang, and Qiu}]{sun2023moss}
Tianxiang Sun, Xiaotian Zhang, Zhengfu He, Peng Li, Qinyuan Cheng, Hang Yan,
  Xiangyang Liu, Yunfan Shao, Qiong Tang, Xingjian Zhao, Ke~Chen, Yining Zheng,
  Zhejian Zhou, Ruixiao Li, Jun Zhan, Yunhua Zhou, Linyang Li, Xiaogui Yang,
  Lingling Wu, Zhangyue Yin, Xuanjing Huang, and Xipeng Qiu.
  2023{\natexlab{b}}.
\newblock Moss: Training conversational language models from synthetic data.

\bibitem[{Team(2023)}]{2023internlm}
InternLM Team. 2023.
\newblock Internlm: A multilingual language model with progressively enhanced
  capabilities.
\newblock \url{https://github.com/InternLM/InternLM}.

\bibitem[{Thorne et~al.(2018)Thorne, Vlachos, Christodoulopoulos, and
  Mittal}]{DBLP:conf/naacl/ThorneVCM18}
James Thorne, Andreas Vlachos, Christos Christodoulopoulos, and Arpit Mittal.
  2018.
\newblock \href {https://doi.org/10.18653/v1/n18-1074} {{FEVER:} a large-scale
  dataset for fact extraction and verification}.
\newblock In \emph{Proceedings of the 2018 Conference of the North American
  Chapter of the Association for Computational Linguistics: Human Language
  Technologies, {NAACL-HLT} 2018, New Orleans, Louisiana, USA, June 1-6, 2018,
  Volume 1 (Long Papers)}, pages 809--819. Association for Computational
  Linguistics.

\bibitem[{Touvron et~al.(2023)Touvron, Lavril, Izacard, Martinet, Lachaux,
  Lacroix, Rozi{\`{e}}re, Goyal, Hambro, Azhar, Rodriguez, Joulin, Grave, and
  Lample}]{DBLP:journals/corr/abs-2302-13971}
Hugo Touvron, Thibaut Lavril, Gautier Izacard, Xavier Martinet, Marie{-}Anne
  Lachaux, Timoth{\'{e}}e Lacroix, Baptiste Rozi{\`{e}}re, Naman Goyal, Eric
  Hambro, Faisal Azhar, Aur{\'{e}}lien Rodriguez, Armand Joulin, Edouard Grave,
  and Guillaume Lample. 2023.
\newblock \href {https://doi.org/10.48550/arXiv.2302.13971} {Llama: Open and
  efficient foundation language models}.
\newblock \emph{CoRR}, abs/2302.13971.

\bibitem[{Wang et~al.(2021)Wang, Liu, and Zhang}]{DBLP:conf/acl/WangL020}
Cunxiang Wang, Pai Liu, and Yue Zhang. 2021.
\newblock \href {https://doi.org/10.18653/v1/2021.acl-long.251} {Can generative
  pre-trained language models serve as knowledge bases for closed-book qa?}
\newblock In \emph{Proceedings of the 59th Annual Meeting of the Association
  for Computational Linguistics and the 11th International Joint Conference on
  Natural Language Processing, {ACL/IJCNLP} 2021, (Volume 1: Long Papers),
  Virtual Event, August 1-6, 2021}, pages 3241--3251. Association for
  Computational Linguistics.

\bibitem[{Wang et~al.(2017)Wang, Liu, and Shi}]{wang-etal-2017-deep}
Yan Wang, Xiaojiang Liu, and Shuming Shi. 2017.
\newblock \href {https://doi.org/10.18653/v1/D17-1088} {Deep neural solver for
  math word problems}.
\newblock In \emph{Proceedings of the 2017 Conference on Empirical Methods in
  Natural Language Processing}, pages 845--854, Copenhagen, Denmark.
  Association for Computational Linguistics.

\bibitem[{Wang et~al.(2023)Wang, Kordi, Mishra, Liu, Smith, Khashabi, and
  Hajishirzi}]{DBLP:conf/acl/WangKMLSKH23}
Yizhong Wang, Yeganeh Kordi, Swaroop Mishra, Alisa Liu, Noah~A. Smith, Daniel
  Khashabi, and Hannaneh Hajishirzi. 2023.
\newblock \href {https://aclanthology.org/2023.acl-long.754} {Self-instruct:
  Aligning language models with self-generated instructions}.
\newblock In \emph{Proceedings of the 61st Annual Meeting of the Association
  for Computational Linguistics (Volume 1: Long Papers), {ACL} 2023, Toronto,
  Canada, July 9-14, 2023}, pages 13484--13508. Association for Computational
  Linguistics.

\bibitem[{Webson and Pavlick(2022)}]{DBLP:conf/naacl/WebsonP22}
Albert Webson and Ellie Pavlick. 2022.
\newblock \href {https://doi.org/10.18653/v1/2022.naacl-main.167} {Do
  prompt-based models really understand the meaning of their prompts?}
\newblock In \emph{Proceedings of the 2022 Conference of the North American
  Chapter of the Association for Computational Linguistics: Human Language
  Technologies, {NAACL} 2022, Seattle, WA, United States, July 10-15, 2022},
  pages 2300--2344. Association for Computational Linguistics.

\bibitem[{Wei et~al.(2022{\natexlab{a}})Wei, Bosma, Zhao, Guu, Yu, Lester, Du,
  Dai, and Le}]{DBLP:conf/iclr/WeiBZGYLDDL22}
Jason Wei, Maarten Bosma, Vincent~Y. Zhao, Kelvin Guu, Adams~Wei Yu, Brian
  Lester, Nan Du, Andrew~M. Dai, and Quoc~V. Le. 2022{\natexlab{a}}.
\newblock \href {https://openreview.net/forum?id=gEZrGCozdqR} {Finetuned
  language models are zero-shot learners}.
\newblock In \emph{The Tenth International Conference on Learning
  Representations, {ICLR} 2022, Virtual Event, April 25-29, 2022}.
  OpenReview.net.

\bibitem[{Wei et~al.(2022{\natexlab{b}})Wei, Tay, Bommasani, Raffel, Zoph,
  Borgeaud, Yogatama, Bosma, Zhou, Metzler, Chi, Hashimoto, Vinyals, Liang,
  Dean, and Fedus}]{DBLP:journals/tmlr/WeiTBRZBYBZMCHVLDF22}
Jason Wei, Yi~Tay, Rishi Bommasani, Colin Raffel, Barret Zoph, Sebastian
  Borgeaud, Dani Yogatama, Maarten Bosma, Denny Zhou, Donald Metzler, Ed~H.
  Chi, Tatsunori Hashimoto, Oriol Vinyals, Percy Liang, Jeff Dean, and William
  Fedus. 2022{\natexlab{b}}.
\newblock \href {https://openreview.net/forum?id=yzkSU5zdwD} {Emergent
  abilities of large language models}.
\newblock \emph{Trans. Mach. Learn. Res.}, 2022.

\bibitem[{Wei et~al.(2022{\natexlab{c}})Wei, Wang, Schuurmans, Bosma, Ichter,
  Xia, Chi, Le, and Zhou}]{DBLP:conf/nips/Wei0SBIXCLZ22}
Jason Wei, Xuezhi Wang, Dale Schuurmans, Maarten Bosma, Brian Ichter, Fei Xia,
  Ed~H. Chi, Quoc~V. Le, and Denny Zhou. 2022{\natexlab{c}}.
\newblock \href
  {http://papers.nips.cc/paper\_files/paper/2022/hash/9d5609613524ecf4f15af0f7b31abca4-Abstract-Conference.html}
  {Chain-of-thought prompting elicits reasoning in large language models}.
\newblock In \emph{NeurIPS}.

\bibitem[{Xu et~al.(2020)Xu, Hu, Zhang, Li, Cao, Li, Xu, Sun, Yu, Yu, Tian,
  Dong, Liu, Shi, Cui, Li, Zeng, Wang, Xie, Li, Patterson, Tian, Zhang, Zhou,
  Liu, Zhao, Zhao, Yue, Zhang, Yang, Richardson, and Lan}]{xu-etal-2020-clue}
Liang Xu, Hai Hu, Xuanwei Zhang, Lu~Li, Chenjie Cao, Yudong Li, Yechen Xu, Kai
  Sun, Dian Yu, Cong Yu, Yin Tian, Qianqian Dong, Weitang Liu, Bo~Shi, Yiming
  Cui, Junyi Li, Jun Zeng, Rongzhao Wang, Weijian Xie, Yanting Li, Yina
  Patterson, Zuoyu Tian, Yiwen Zhang, He~Zhou, Shaoweihua Liu, Zhe Zhao, Qipeng
  Zhao, Cong Yue, Xinrui Zhang, Zhengliang Yang, Kyle Richardson, and Zhenzhong
  Lan. 2020.
\newblock \href {https://doi.org/10.18653/v1/2020.coling-main.419} {{CLUE}: A
  {C}hinese language understanding evaluation benchmark}.
\newblock In \emph{Proceedings of the 28th International Conference on
  Computational Linguistics}, pages 4762--4772, Barcelona, Spain (Online).
  International Committee on Computational Linguistics.

\bibitem[{Xu et~al.(2021)Xu, Lu, Yuan, Zhang, Yuan, Xu, Wei, Pan, and
  Hu}]{DBLP:journals/corr/abs-2107-07498}
Liang Xu, Xiaojing Lu, Chenyang Yuan, Xuanwei Zhang, Hu~Yuan, Huilin Xu, Guoao
  Wei, Xiang Pan, and Hai Hu. 2021.
\newblock \href {http://arxiv.org/abs/2107.07498} {Fewclue: {A} chinese
  few-shot learning evaluation benchmark}.
\newblock \emph{CoRR}, abs/2107.07498.

\bibitem[{Zeng et~al.(2023)Zeng, Liu, Du, Wang, Lai, Ding, Yang, Xu, Zheng,
  Xia, Tam, Ma, Xue, Zhai, Chen, Liu, Zhang, Dong, and
  Tang}]{DBLP:conf/iclr/ZengLDWL0YXZXTM23}
Aohan Zeng, Xiao Liu, Zhengxiao Du, Zihan Wang, Hanyu Lai, Ming Ding, Zhuoyi
  Yang, Yifan Xu, Wendi Zheng, Xiao Xia, Weng~Lam Tam, Zixuan Ma, Yufei Xue,
  Jidong Zhai, Wenguang Chen, Zhiyuan Liu, Peng Zhang, Yuxiao Dong, and Jie
  Tang. 2023.
\newblock \href {https://openreview.net/pdf?id=-Aw0rrrPUF} {{GLM-130B:} an open
  bilingual pre-trained model}.
\newblock In \emph{The Eleventh International Conference on Learning
  Representations, {ICLR} 2023, Kigali, Rwanda, May 1-5, 2023}. OpenReview.net.

\bibitem[{Zeng(2023)}]{DBLP:journals/corr/abs-2304-12986}
Hui Zeng. 2023.
\newblock \href {https://doi.org/10.48550/arXiv.2304.12986} {Measuring massive
  multitask chinese understanding}.
\newblock \emph{CoRR}, abs/2304.12986.

\bibitem[{Zhang et~al.(2019)Zhang, Sun, Wan, and
  Guo}]{DBLP:conf/nlpcc/ZhangSWG19}
Bowei Zhang, Weiwei Sun, Xiaojun Wan, and Zongming Guo. 2019.
\newblock \href {https://doi.org/10.1007/978-3-030-32233-5\_63} {{PKU}
  paraphrase bank: {A} sentence-level paraphrase corpus for chinese}.
\newblock In \emph{Natural Language Processing and Chinese Computing - 8th
  {CCF} International Conference, {NLPCC} 2019, Dunhuang, China, October 9-14,
  2019, Proceedings, Part {I}}, volume 11838 of \emph{Lecture Notes in Computer
  Science}, pages 814--826. Springer.

\bibitem[{Zhang and Liu(2017)}]{DBLP:reference/ml/0016017}
Lei Zhang and Bing Liu. 2017.
\newblock \href {https://doi.org/10.1007/978-1-4899-7687-1\_907} {Sentiment
  analysis and opinion mining}.
\newblock In Claude Sammut and Geoffrey~I. Webb, editors, \emph{Encyclopedia of
  Machine Learning and Data Mining}, pages 1152--1161. Springer.

\bibitem[{Zhang et~al.(2018)Zhang, Zhang, Wang, Guo, and
  Liu}]{DBLP:journals/access/ZhangZWGL18}
Sheng Zhang, Xin Zhang, Hui Wang, Lixiang Guo, and Shanshan Liu. 2018.
\newblock \href {https://doi.org/10.1109/ACCESS.2018.2883637} {Multi-scale
  attentive interaction networks for chinese medical question answer
  selection}.
\newblock \emph{{IEEE} Access}, 6:74061--74071.

\bibitem[{Zhang et~al.(2023)Zhang, Li, Zong, Ying, He, and
  Qiu}]{DBLP:journals/corr/abs-2305-12474}
Xiaotian Zhang, Chunyang Li, Yi~Zong, Zhengyu Ying, Liang He, and Xipeng Qiu.
  2023.
\newblock \href {https://doi.org/10.48550/arXiv.2305.12474} {Evaluating the
  performance of large language models on {GAOKAO} benchmark}.
\newblock \emph{CoRR}, abs/2305.12474.

\bibitem[{Zheng et~al.(2019)Zheng, Huang, and Sun}]{zheng-etal-2019-chid}
Chujie Zheng, Minlie Huang, and Aixin Sun. 2019.
\newblock \href {https://doi.org/10.18653/v1/P19-1075} {{C}h{ID}: A large-scale
  {C}hinese {ID}iom dataset for cloze test}.
\newblock In \emph{Proceedings of the 57th Annual Meeting of the Association
  for Computational Linguistics}, pages 778--787, Florence, Italy. Association
  for Computational Linguistics.

\bibitem[{Zheng et~al.(2023)Zheng, Chiang, Sheng, Zhuang, Wu, Zhuang, Lin, Li,
  Li, Xing, Zhang, Gonzalez, and Stoica}]{DBLP:journals/corr/abs-2306-05685}
Lianmin Zheng, Wei{-}Lin Chiang, Ying Sheng, Siyuan Zhuang, Zhanghao Wu,
  Yonghao Zhuang, Zi~Lin, Zhuohan Li, Dacheng Li, Eric~P. Xing, Hao Zhang,
  Joseph~E. Gonzalez, and Ion Stoica. 2023.
\newblock \href {https://doi.org/10.48550/arXiv.2306.05685} {Judging
  llm-as-a-judge with mt-bench and chatbot arena}.
\newblock \emph{CoRR}, abs/2306.05685.

\bibitem[{Zhong et~al.(2023)Zhong, Cui, Guo, Liang, Lu, Wang, Saied, Chen, and
  Duan}]{DBLP:journals/corr/abs-2304-06364}
Wanjun Zhong, Ruixiang Cui, Yiduo Guo, Yaobo Liang, Shuai Lu, Yanlin Wang, Amin
  Saied, Weizhu Chen, and Nan Duan. 2023.
\newblock \href {https://doi.org/10.48550/arXiv.2304.06364} {Agieval: {A}
  human-centric benchmark for evaluating foundation models}.
\newblock \emph{CoRR}, abs/2304.06364.

\bibitem[{Zhou et~al.(2022)Zhou, Deng, Mi, Li, Wang, Huang, Jiang, Liu, and
  Meng}]{DBLP:journals/corr/abs-2202-08011}
Jingyan Zhou, Jiawen Deng, Fei Mi, Yitong Li, Yasheng Wang, Minlie Huang, Xin
  Jiang, Qun Liu, and Helen Meng. 2022.
\newblock \href {http://arxiv.org/abs/2202.08011} {Towards identifying social
  bias in dialog systems: Frame, datasets, and benchmarks}.
\newblock \emph{CoRR}, abs/2202.08011.

\bibitem[{Zhu et~al.(2023)Zhu, Wang, Zhou, Wang, Chen, Wang, Yang, Ye, Gong,
  Zhang, and Xie}]{DBLP:journals/corr/abs-2306-04528}
Kaijie Zhu, Jindong Wang, Jiaheng Zhou, Zichen Wang, Hao Chen, Yidong Wang,
  Linyi Yang, Wei Ye, Neil~Zhenqiang Gong, Yue Zhang, and Xing Xie. 2023.
\newblock \href {https://doi.org/10.48550/arXiv.2306.04528} {Promptbench:
  Towards evaluating the robustness of large language models on adversarial
  prompts}.
\newblock \emph{CoRR}, abs/2306.04528.

\bibitem[{Zhu et~al.(2020)Zhu, Huang, Zhang, Zhu, and
  Huang}]{DBLP:journals/tacl/ZhuHZZH20}
Qi~Zhu, Kaili Huang, Zheng Zhang, Xiaoyan Zhu, and Minlie Huang. 2020.
\newblock \href {https://doi.org/10.1162/tacl\_a\_00314} {Crosswoz: {A}
  large-scale chinese cross-domain task-oriented dialogue dataset}.
\newblock \emph{Trans. Assoc. Comput. Linguistics}, 8:281--295.

\bibitem[{Zou et~al.(2023)Zou, Wang, Kolter, and Fredrikson}]{zou2023universal}
Andy Zou, Zifan Wang, J.~Zico Kolter, and Matt Fredrikson. 2023.
\newblock \href {http://arxiv.org/abs/2307.15043} {Universal and transferable
  adversarial attacks on aligned language models}.

\end{thebibliography}
\bibliographystyle{acl_natbib}

\clearpage

\appendix

\begin{figure*}[ht]
  \begin{subfigure}{.5\textwidth}
	\includegraphics[width=1\linewidth]{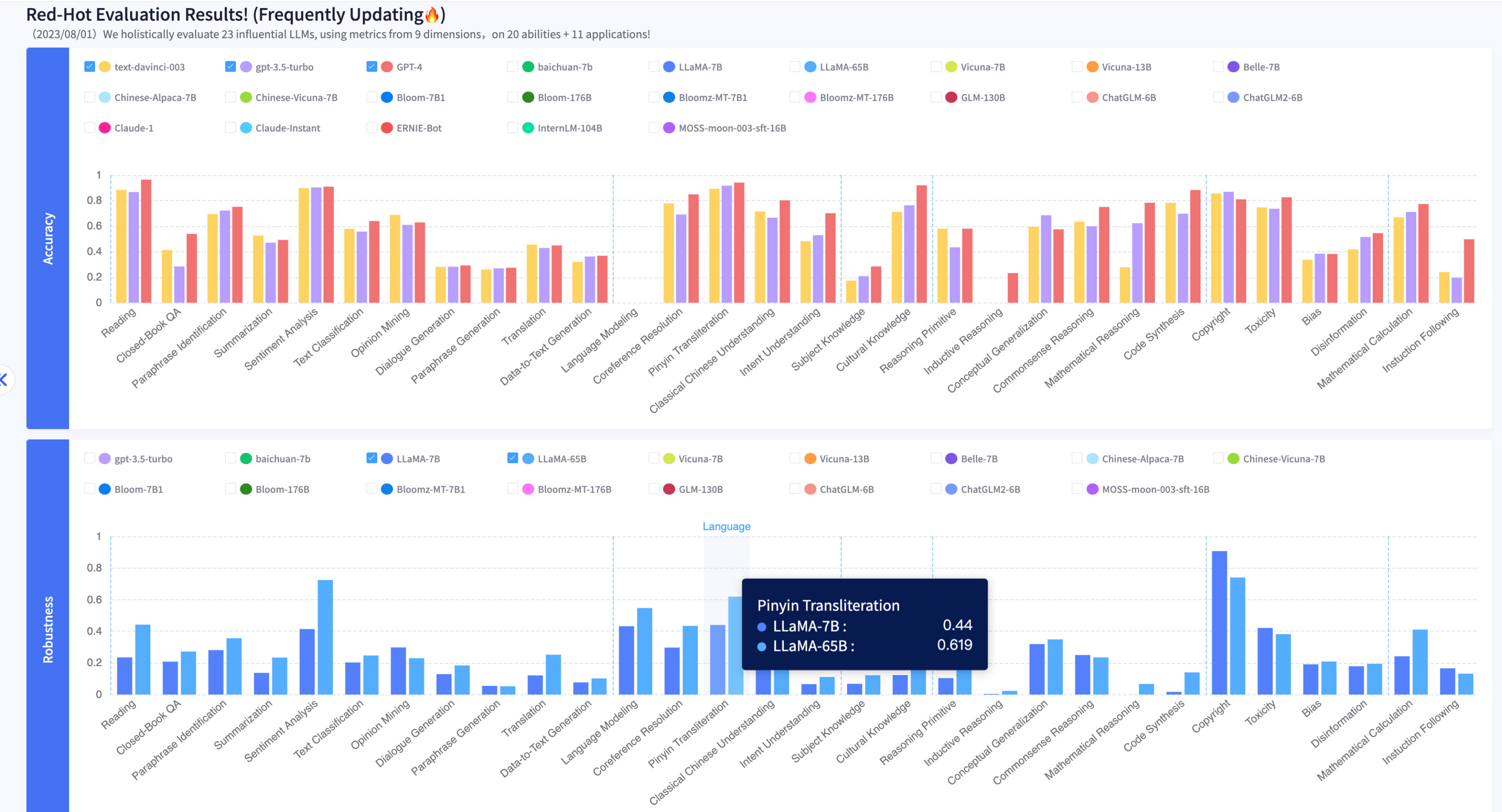}
	\caption{Evaluation Results Overview}
	\label{fig:plat-overview}
  \end{subfigure}
  \begin{subfigure}{.5\textwidth}
    \centering
    \includegraphics[width=1.0\linewidth]{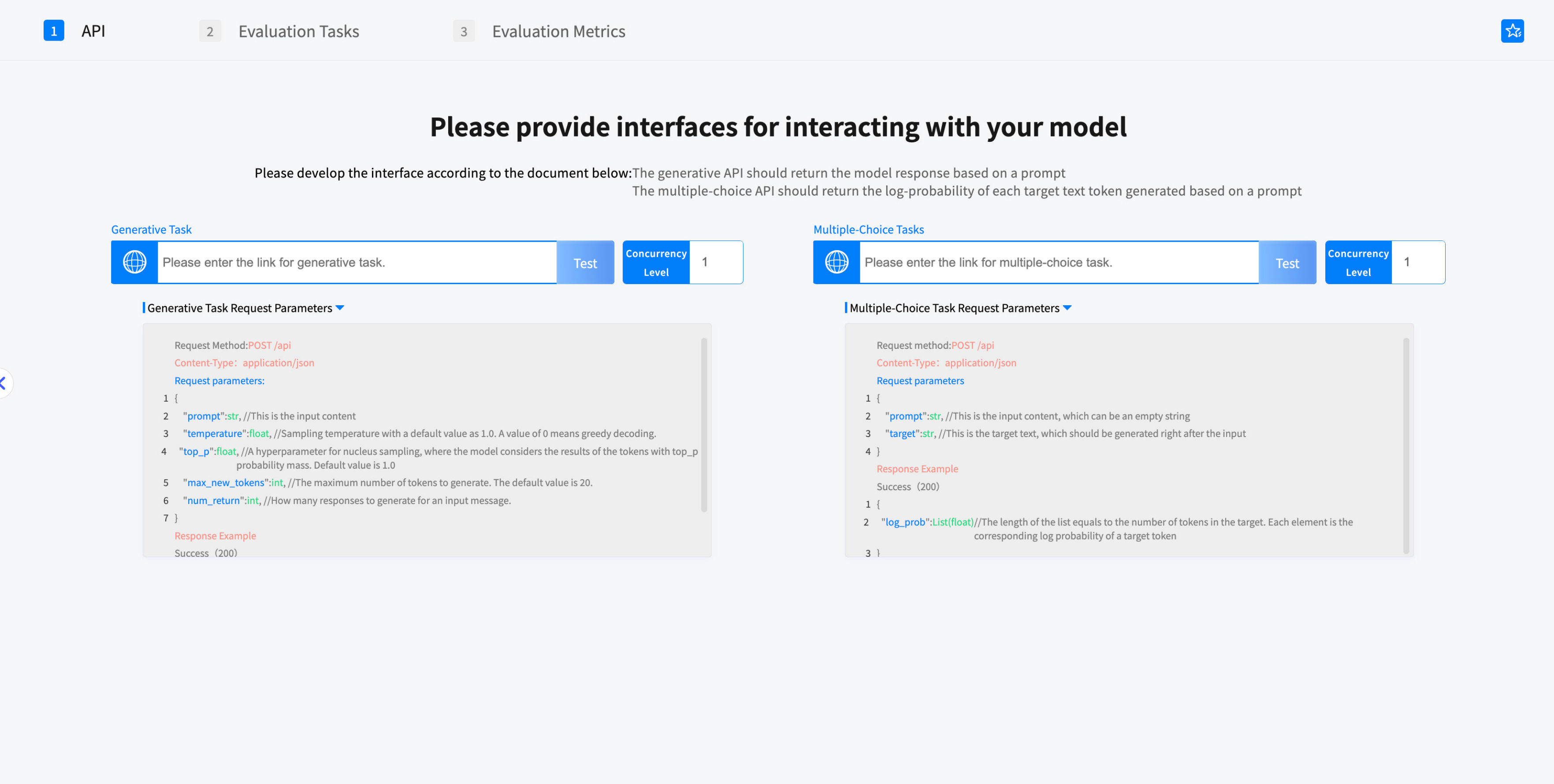}
    \caption{Step 1: Provide APIs for Evaluation}
    \label{fig:plat-api}
  \end{subfigure}
  \begin{subfigure}{.5\textwidth}
    \centering
    \includegraphics[width=1.0\linewidth]{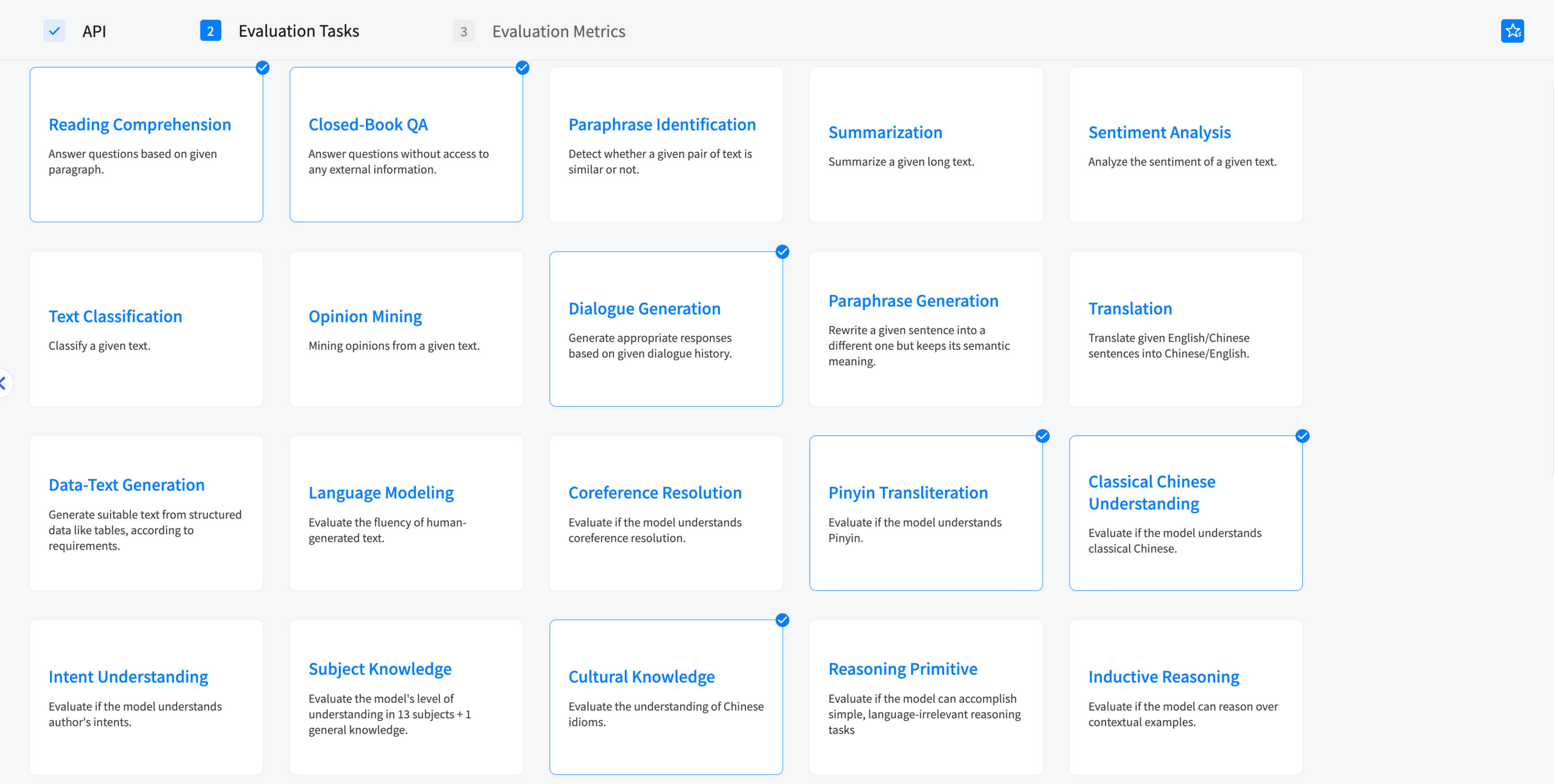}
    \caption{Step 2: Select Relevant Tasks}
    \label{fig:plat-tasks}
  \end{subfigure}
  \begin{subfigure}{.5\textwidth}
    \centering
    \includegraphics[width=1.0\linewidth]{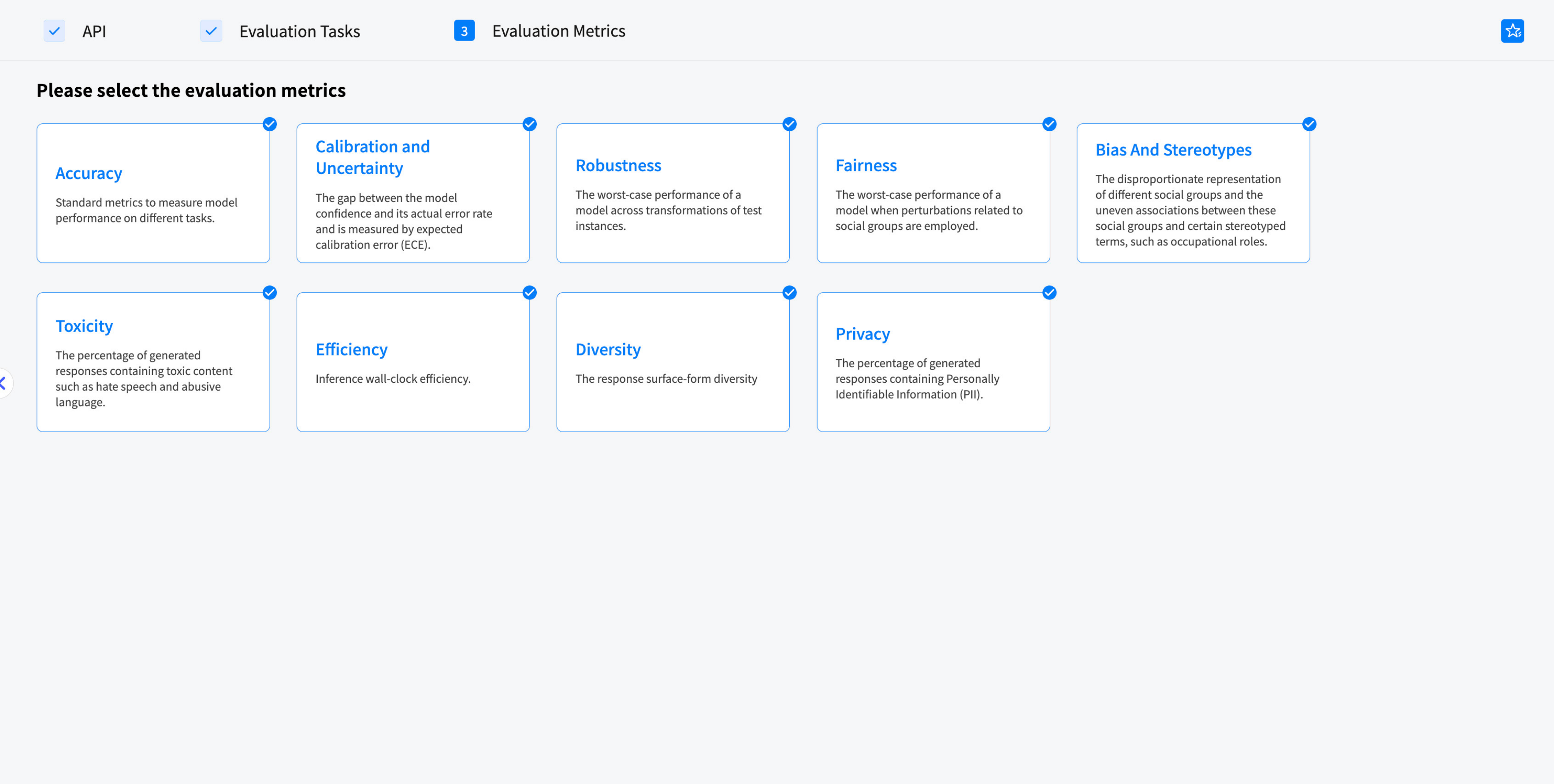}
    \caption{Step 3: Select Desired Metrics}
    \label{fig:plat-metrics}
  \end{subfigure}
  \caption{\platform{} provides a user-friendly interface. With only several clicks and minimum coding, evaluating a new language model can be deployed in a few minutes.}
\end{figure*}

\section{Platform Usage}
\label{app:platform}

To fully utilize our \platform{} to evaluate a large language model, users can take advantage of our user-friendly web application. 
As shown by Figure~\ref{fig:plat-overview}, users will first see our latest leaderboard results with an interactive interface.
Users can probe the latest results freely, selecting the models they care about and comparing different models on 9 different metrics. 
If a user intends to evaluate a new model, a holistic evaluation can be deployed with just a few mouse clicks and model APIs:
The process initiates with users inputting a specific link that enables our platform to interface with the to-be-evaluated model, as shown by Figure~\ref{fig:plat-api}.
Subsequently, users are granted the flexibility to select applicable tasks from an extensive set of 31 pre-defined options (Figure~\ref{fig:plat-tasks}).
The concluding step involves the selection of the appropriate evaluation metrics, from the 9 available options (Figure~\ref{fig:plat-metrics}).



\section{Benchmark}
\label{app:benchmark}

In this section, we provide a detailed description along with an example for each task involved in our benchmark.
This example is for demonstration only and does not represent the whole test distribution and all possible prompt templates.
We also accompany the English translation after each Chinese example.
In the provided example, text highlighted in \colorbox{ugreen!30}{green} is a reference that we expect LLMs to predict and the other part is prompt constructed by a random prompt template and input fields from a random test instance.


\subsection{Ability Evaluation}

\subsubsection{Language}

\paragraph{Language Modelling.}

This task asks the LLM to score the probability of the input text.
We use \ul{bits per bytes}~\cite{DBLP:journals/corr/abs-2101-00027} as the metric that allows us to make comparisons with different tokenizers.
Data are sampled from CLUECorpus2020~\cite{xu-etal-2020-clue}.

\paragraph{Coreference Resolution.}

Coreference resolution is a traditional NLP task.
We sample data from CLUEWSC~\cite{DBLP:journals/corr/abs-2107-07498}, where the model must answer whether a given pronoun refers to a given entity (the Winograd Schema Challenge).
We use \ul{accuracy} as the metric for this problem.
A coreference resolution example is shown below:

\begin{quote}
\scriptsize
\textbf{\textsl{Chinese Example}}:\\
\begin{CJK*}{UTF8}{gbsn}
    蒋盈波原来所在的教研室有位副教授去德国参加一个学术活动，活动中结识了一位华裔德籍的同行，那同行在自己家中招待了他一次，言谈之间，双方忽然都感到巧事真多，而世界真小\\
    在这里，``他''的意思是``同行''。是或否？\colorbox{ugreen!30}{否}\\
\end{CJK*}
\\
\textbf{\textsl{English Translation}}:\\
An associate professor from the research office where Jiang Yingbo used to work went to Germany to attend an academic event. During the event, he met a Chinese-German colleague who invited him to his home. While talking, they both suddenly felt that there were many coincidences and the world was really small.\\
Here, does ``him'' refer to ``colleague''? Yes or No?\colorbox{ugreen!30}{ No}
\end{quote}

\paragraph{Pinyin Transliteration.}

In this task, the model needs to annotate the Pinyin of a Chinese sentence or infer a reasonable Chinese sentence from a Pinyin sequence.
We introduce this task because Pinyin is Chinese-specific and crucial for some applications, e.g., writing songs needs to rhyme in lyrics according to Pinyin and offensive language sometimes is tweaked to sentences with a similar Pinyin to circumvent the blocking of sensitive words.
Since this task is newly introduced and there is no primary metric available, we treat this task as a translation task and evaluate the performance with \ul{BLEU}~\cite{papineni-etal-2002-bleu}.
A Chinese-to-Pinyin transliteration example is shown below:

\begin{quote}
\scriptsize
\textbf{\textsl{Chinese Example}}:\\
\begin{CJK*}{UTF8}{gbsn}
    将以下句子在汉字和汉语拼音之间进行转译。\\
    \\
    汉字：因此，依靠科技进步，强化科学管理已成为实现油田稳产的当务之急\\
    拼音：\colorbox{ugreen!30}{\pinyin{yin1ci3}，\pinyin{yi1kao4ke1ji4jin4bu4}，\pinyin{qiang2hua4ke1xue2guan3li3}}\\
    \colorbox{ugreen!30}{\pinyin{yi3cheng1wei2shi2xian4you2tian2wen3chan3di2dang4wu4zhi1ji2}}\\
\end{CJK*}
\\
\textbf{\textsl{English Translation}}:\\
Translate the following sentence between Chinese and Pinyin.\\
\\
Chinese: Therefore, relying on technological progress and strengthening scientific management has become an urgent task to achieve stable oilfield production\\
Pinyin: \colorbox{ugreen!30}{\pinyin{yin1ci3}, \pinyin{yi1kao4ke1ji4jin4bu4}, \pinyin{qiang2hua4ke1xue2guan3li3}}\\
\colorbox{ugreen!30}{\pinyin{yi3cheng1wei2shi2xian4you2tian2wen3chan3di2dang4wu4zhi1ji2}}
\end{quote}

\paragraph{Intent Understanding.}

We introduce this task to test whether Chinese LLMs could capture the writing intent of the authors of a long document.
This task helps measure how well LLMs can understand implications.
We formulate this task as a multi-choice problem and adopt \ul{accuracy} to assess the performance.
An example is shown below:

\begin{quote}
\scriptsize
\textbf{\textsl{Chinese Example}}:\\
\begin{CJK*}{UTF8}{gbsn}
    亚马孙丛林中的雄性蓝蝶带有彩虹般的蓝色光辉，半公里外就能看到。其光辉如此强烈，有的竟能反射70\%的蓝色光线，远远超过蓝色涂料的反射率。蓝蝶耀眼的光辉，原是一种警号，使别的雄性蓝蝶在远处就能知所趋避。蓝光越强，示警作用越显著。物竞天择，适者生存。亿万年的自然选择，使亚马孙蓝蝶翅膀有了如此奇妙的性能。\\
    \ldots\ldots\\
    
    对有关蓝蝶的仿生研究，理解不恰当的一项是\\
    A. 在蓝蝶仿生的各类应用研究中，证券防伪的研究最有成效。\\
    B. 翅膀上的羽状物的构造和尺寸，是仿生学家们极感兴趣的课题。\\
    C. 新型的变幻色彩的迷彩服，可能将与蓝蝶翅膀的反光结构有关。\\
    D. 对蓝蝶翅膀的反光机理的应用研究，目前还没取得突破性的结果。\\
    答：\colorbox{ugreen!30}{A}\\
\end{CJK*}
\\
\textbf{\textsl{English Translation}}:\\
Male blue butterflies in the Amazon jungle have a rainbow-like blue glow that can be seen from half a kilometer away. Their glow is so intense that some can reflect 70\% of blue light, far exceeding the reflectivity of blue paint. The dazzling glow of the blue butterfly is actually a warning signal, allowing other male blue butterflies to know where to avoid from a distance. The stronger the blue light, the more obvious the warning effect. Survival of the fittest. Millions of years of natural selection have given the wings of Amazon blue butterflies such a wonderful performance.\\
\ldots\ldots\\

Regarding the bionic research on blue butterflies, one item that is not properly understood is\\
A. Among the various applications of bionic research on blue butterflies, the research on securities anti-counterfeiting is the most effective.\\
B. The structure and size of the feather-like structures on the wings are topics of great interest to bionics researchers.\\
C. The new type of color-changing camouflage suit may be related to the reflective structure of the blue butterfly wings.\\
D. The application research on the reflection mechanism of the blue butterfly wings has not achieved any breakthrough results so far.\\
Answer:\colorbox{ugreen!30}{ A}
\end{quote}

\paragraph{Classical Chinese Understanding.}

Classical Chinese plays an important role in Chinese culture.
Quatrain, poetry and etc. are all rooted in classical Chinese and most of them frequently appear in modern Chinese literature.
Therefore we include this task to examine the model's understanding of classical Chinese.
We sample multi-choice questions from CCPM~\cite{DBLP:journals/corr/abs-2106-01979} that inquire about the semantic equivalence between a modern Chinese sentence and a list of classical Chinese candidates.
We use \ul{accuracy} as the primary metric.
Below is an example:

\begin{quote}
\scriptsize
\textbf{\textsl{Chinese Example}}:\\
\begin{CJK*}{UTF8}{gbsn}
    ``山间连绵阴雨刚刚有了一点停止的意思。''这句话可以用以下哪句古文来表达：\\
    A. 寒雨初开霁\\
    B. 山晓雨初霁\\
    C. 宿雨天初霁\\
    D. 山雨初含霁\\
    答：\colorbox{ugreen!30}{D}\\
\end{CJK*}
\\
\textbf{\textsl{English Translation}}:\\
``The continuous rain in the mountains has just shown a little sign of stopping.'' Which of the following ancient Chinese sentences can be used to express this sentence:\\
\begin{CJK*}{UTF8}{gbsn}
    A. Cold rain just stops\\
    B. A morning in mountains, rain just stops\\
    C. An over-night rain just stops\\
    D. Rain in the mountain is stopping\\
\end{CJK*}
Answer:\colorbox{ugreen!30}{ D}
\end{quote}

\subsubsection{Knowledge}

\paragraph{Subject Knowledge.}

This task is in the format of fact completion~\cite{DBLP:conf/emnlp/PetroniRRLBWM19}, where LLMs fill in the blank of a Chinese sentence with entities.
Here we construct the dataset as in \citet{DBLP:conf/emnlp/PetroniRRLBWM19}, which tests the knowledge from 13 subjects and 1 general domain.
The metric is \ul{Accuracy@$K$} ($K=1,5$).
We provide a math knowledge example:

\begin{quote}
\scriptsize
\textbf{\textsl{Chinese Example}}:\\
\begin{CJK*}{UTF8}{gbsn}
    婆罗摩笈多公式描述了\_\_ -> \colorbox{ugreen!30}{四边形}\\
\end{CJK*}
\\
\textbf{\textsl{English Translation}}:\\
The Brahmagupta formula describes\_\_ ->\colorbox{ugreen!30}{ quadrilateral}
\end{quote}

\paragraph{Cultural Knowledge.}

Here we query Chinese LLMs with multi-choice questions related to Chinese culture, e.g., idioms.
Data are sampled from ChID~\cite{zheng-etal-2019-chid}.
We adopt \ul{accuracy} as the primary metric and show an idiom example below:

\begin{quote}
\scriptsize
\textbf{\textsl{Chinese Example}}:\\
\begin{CJK*}{UTF8}{gbsn}
    \scriptsize
    不过，要想变得\_\_\_\_，要想自己能够成就一番事业的话，不是说来就来的，或者说任何一个出色的人，他们都得经历过不少的磨难，以及在经受住了一些挫折之后，才能真正成才成人，才能成为一个实力超群的人物，让自己的人生过得越来越顺当\ldots\\
    上面这个句子下划线处可以填写哪个成语？\\
    A. 足智多谋\\
    B. 语无伦次\\
    C. 绣花枕头\\
    答：\colorbox{ugreen!30}{A}\\
\end{CJK*}
\\
\textbf{\textsl{English Translation}}:\\
However, if you want to become \_\_\_\_ to achieve something on your own, it doesn’t come easily. Any outstanding person has to go through a lot of hardships and setbacks before they can truly succeed and become a person of exceptional ability, making their life smoother and smoother\ldots\\
Which idiom can be filled in the blank in the sentence above?\\
A. Wise and resourceful\\
B. Speak incoherently\\
C. Pretty on the outside but lacking substance underneath\\
Answer:\colorbox{ugreen!30}{ A}
\end{quote}

\subsubsection{Reasoning}

\paragraph{Reasoning Primitive.}

Following HELM~\cite{DBLP:journals/corr/abs-2211-09110}, reasoning primitive is a collection of reasoning tasks independent of language and knowledge background and focuses on abstracted reasoning capacity evaluation.
It includes tasks like non-ampliative reasoning, ampliative reasoning, recursive hierarchy and etc.
Readers can refer to \citet{DBLP:journals/corr/abs-2211-09110} for more details.
Here we synthesize the dataset similar to HELM~\cite{DBLP:journals/corr/abs-2211-09110} and use \ul{exact match} to evaluate the final performance.
Below is a recursive hierarchy example (in Dyck languages):

\begin{quote}
\begin{CJK*}{UTF8}{gbsn}
    \scriptsize
    [ [ [ [ [ \{ [ [ [ [ \{ \{ ( ( ) [ ( ( [ \{ \} ] ) \{ \{ \} \} ) [ [ ] ] ( ) ] ) [ [ ( ( ) ) ( ) ] ] \} \} ] ] ] ] \} ] ] ] ] ] [ \{\colorbox{ugreen!30}{ \} ]}
\end{CJK*}
\end{quote}

\paragraph{Realistic Reasoning.}

Contrary to reasoning primitive, in-the-wild reasoning combines the abstract reasoning skill of LLMs and their knowledge as well as the understanding of context (e.g., mathematical reasoning requires LLMs to be able to perform simple arithmetics).
We choose the following reasoning tasks that not only help better surface the reasoning skills of LLMs but also have practical applications.

\begin{itemize}[noitemsep, nolistsep]
    \item \textbf{Inductive Reasoning} is to draw conclusions by going through a set of examples. Here the model needs to infer the rule from the few-shot demonstrations we provided and apply the rule to new examples. Data are collected from BIG-Bench~\cite{srivastava2023beyond}. We use \ul{exact match} as the evaluation metric and an example goes like this:

    \begin{quote}
    \scriptsize
    \textbf{\textsl{Chinese Example}}:\\
    \begin{CJK*}{UTF8}{gbsn}
        \scriptsize
        推断符号 -> 的含义并计算下列公式。\\
        512 + 372 -> 885\\
        528 + 170 -> 699\\
        859 + 133 -> 993\\
        199 + 944 -> 1144\\
        154 + 521 -> 676\\
        67 + 987 ->\colorbox{ugreen!30}{ 1055}\\
    \end{CJK*}
    \\
    \textbf{\textsl{English Translation}}:\\
    Infer the meaning of the symbol -> and calculate the following formula.\\
    512 + 372 -> 885\\
    528 + 170 -> 699\\
    859 + 133 -> 993\\
    199 + 944 -> 1144\\
    154 + 521 -> 676\\
    67 + 987 ->\colorbox{ugreen!30}{ 1055}
    \end{quote}

    \item \textbf{Deductive Reasoning} is contrasted with inductive reasoning, where the model progresses from conclusions to specific examples. We provide an example of \emph{modus tollens}\footnote{\url{https://plato.stanford.edu/entries/logic-ancient/\#ForModPonModTol}}, a form of deductive argument, in which the model predicts whether a given conclusion is valid or not according to the previous statements. Data are translated from \citet{DBLP:journals/corr/abs-2306-09479} and we use \ul{accuracy} as the evaluation metric.

    \begin{quote}
    \scriptsize
    \textbf{\textsl{Chinese Example}}:\\
    \begin{CJK*}{UTF8}{gbsn}
        \scriptsize
        考虑以下事实：\\
        1.如果朱莉娅喜欢甲壳虫乐队，那么朱莉娅就是吉他手。\\
        2.朱莉娅不是吉他手。\\
        结论：因此，朱莉娅不喜欢甲壳虫乐队。\\
        \\
        问题：根据陈述1.和2.，结论是否有效？\\
        \\
        回答：\colorbox{ugreen!30}{是}\\
    \end{CJK*}
    \\
    \textbf{\textsl{English Translation}}:\\
    Consider the following facts:\\
    1. If Julia likes the Beatles, then Julia is a guitarist.\\
    2. Julia is not a guitarist.\\
    Conclusion: Therefore, Julia does not like the Beatles.\\
    \\
    Question: Based on statements 1. and 2., is the conclusion valid?\\
    \\
    Answer:\colorbox{ugreen!30}{ Yes}
    \end{quote}
    
    \item \textbf{Commonsense Reasoning} is an umbrella of all related tasks, e.g., natural language inference and commonsense question answering~\cite{DBLP:journals/corr/abs-1904-01172}. We mainly evaluate the classical natural language inference (data are sampled from OCNLI~\cite{DBLP:journals/corr/abs-2010-05444}) and commonsense question answering (data are translated from \citet{DBLP:journals/corr/abs-2306-09479}). We organize them into multi-choice tasks and adopt \ul{accuracy} for assessment. Here we provide a textual entailment example:

    \begin{quote}
    \scriptsize
    \textbf{\textsl{Chinese Example}}:\\
    \begin{CJK*}{UTF8}{gbsn}
        \scriptsize
        是否可以从``篮子嘛,一块钱,一块钱啊.”中推断出“这个篮子是可以卖的。''？\\
        A. 总是可以\\
        B. 有时可以\\
        C. 不可以\\
        答：\colorbox{ugreen!30}{A}\\
    \end{CJK*}
    \\
    \textbf{\textsl{English Translation}}:\\
    Can it be inferred from ``The basket, one yuan, one yuan.'' that ``This basket is for sale.''?\\
    A. Always\\
    B. Sometimes\\
    C. Never\\
    Answer:\colorbox{ugreen!30}{ A}
    \end{quote}

    \item \textbf{Mathematical Reasoning} also has a rather big scope that envelopes various tasks, e.g., math word problem (MWP) solving, theorem proving and etc.~\cite{DBLP:conf/acl/Lu00WC23}. Here we focus on MWP and adopt \ul{exact match} for evaluation. Data are sampled from Math23K~\cite{wang-etal-2017-deep}. An MWP in our benchmark is: 

    \begin{quote}
    \scriptsize
    \textbf{\textsl{Chinese Example}}:\\
    \begin{CJK*}{UTF8}{gbsn}
        \scriptsize
        问题：一个饲养场，养鸭1200只，养的鸡比养的鸭多(3/5)，养的鸡比鸭多多少只？\\
        答案：\colorbox{ugreen!30}{720}\\
    \end{CJK*}
    \\
    \textbf{\textsl{English Translation}}:\\
    Question: A farm has 1200 ducks, and the number of chickens raised is (3/5) more than the number of ducks raised. How many more chickens are there than ducks?\\
    Answer:\colorbox{ugreen!30}{ 720}
    \end{quote}
    
    \item \textbf{Code Synthesis} is a task to synthesize an executable program that matches the requirement written in natural language. Data are translated from HumanEval~\cite{DBLP:journals/corr/abs-2107-03374} and we use \ul{pass@$k$} as the metric ($k=1,10,100$). An example is shown below:

    \begin{quote}
    \tiny
    \textbf{\textsl{Chinese Example}}:
\begin{alltt}
def is\_sorted(lst):\\
    '''
\end{alltt}
\begin{CJK*}{UTF8}{gbsn}
\ \ \ \ \ \ \ \ 给定一个数字列表，返回它们是否以升序排序。\\
\indent\ \ \ \ \ \ \ \ 如果列表有两个及以上的相同数字，则返回\texttt{False}。\\
\indent\ \ \ \ \ \ \ \ 假设没有负数且只有整数。\\
    \\
\indent\ \ \ \ \ \ \ \ 示例：
\end{CJK*}
\begin{alltt}
    is\_sorted([5]) -> True\\
    is\_sorted([1, 2, 3, 4, 5]) -> True\\
    is\_sorted([1, 3, 2, 4, 5]) -> False\\
    is\_sorted([1, 2, 3, 4, 5, 6]) -> True\\
    is\_sorted([1, 2, 3, 4, 5, 6, 7]) -> True\\
    is\_sorted([1, 3, 2, 4, 5, 6, 7]) -> False\\
    is\_sorted([1, 2, 2, 3, 3, 4]) -> True\\
    is\_sorted([1, 2, 2, 2, 3, 4]) -> False\\
    '''
\end{alltt}
\begin{alltt}
\hlc[ugreen!30]{
    count\_digit = dict([(i, 0) for i in lst])\\
    for i in lst:\\
        count\_digit[i]+=1 \\
    if any(count\_digit[i] > 2 for i in lst):\\
        return False\\
    if all(lst[i-1] <= lst[i] for i in range(1, len(lst))):\\
        return True\\
    else:\\
        return False\\}
\end{alltt}
    \textbf{\textsl{English Translation}}:
\begin{alltt}
def is\_sorted(lst):\\
    '''
\end{alltt}
\begin{CJK*}{UTF8}{gbsn}
\ \ \ \ \ \ \ \ Given a list of numbers, return whether they are sorted in\\
\indent\ \ \ \ \ \ \ \ ascending order.\\
\indent\ \ \ \ \ \ \ \ If the list has two or more identical numbers, return \texttt{False}.\\
\indent\ \ \ \ \ \ \ \ Assume that there are no negative numbers and only integers.\\
    \\
\indent\ \ \ \ \ \ \ \ Examples：
\end{CJK*}
\begin{alltt}
    is\_sorted([5]) -> True\\
    is\_sorted([1, 2, 3, 4, 5]) -> True\\
    is\_sorted([1, 3, 2, 4, 5]) -> False\\
    is\_sorted([1, 2, 3, 4, 5, 6]) -> True\\
    is\_sorted([1, 2, 3, 4, 5, 6, 7]) -> True\\
    is\_sorted([1, 3, 2, 4, 5, 6, 7]) -> False\\
    is\_sorted([1, 2, 2, 3, 3, 4]) -> True\\
    is\_sorted([1, 2, 2, 2, 3, 4]) -> False\\
    '''
\end{alltt}
\begin{alltt}
\hlc[ugreen!30]{
    count\_digit = dict([(i, 0) for i in lst])\\
    for i in lst:\\
        count\_digit[i]+=1 \\
    if any(count\_digit[i] > 2 for i in lst):\\
        return False\\
    if all(lst[i-1] <= lst[i] for i in range(1, len(lst))):\\
        return True\\
    else:\\
        return False\\}
\end{alltt}
    \end{quote}
    
    \item \textbf{Conceptual Generalization} is a new task that is similar to inductive reasoning, where the model must reason over concrete examples to get a general rule and apply it to unseen examples. The reason we separate this task from inductive reasoning is that this task is specialized in reasoning over physical concepts or relations like directions. Data are synthesized as suggested by \citet{DBLP:conf/iclr/PatelP22} and we employ \ul{top-$k$ accuracy} ($k=1$) to measure the performance.
    An example is shown below:

    \begin{quote}
    \scriptsize
    \textbf{\textsl{Chinese Example}}:\\
    \begin{CJK*}{UTF8}{gbsn}
        世界：\\
        \lbrack 0, 1, 0, 0\rbrack\\
        \lbrack 0, 0, 0, 0\rbrack\\
        答案：顶\\
        \\
        世界：\\
        \lbrack 1, 0, 0\rbrack\\
        \lbrack 0, 0, 0\rbrack\\
        答案：左\\
        \\
        世界：\\
        \lbrack 0, 1\rbrack\\
        \lbrack 0, 0\rbrack\\
        答案：上\\
        \\
        世界：\\
        \lbrack 0, 0, 0, 0, 0\rbrack\\
        \lbrack 0, 0, 0, 0, 0\rbrack\\
        \lbrack 0, 0, 0, 0, 0\rbrack\\
        \lbrack 0, 0, 0, 0, 0\rbrack\\
        \lbrack 1, 0, 0, 0, 0\rbrack\\
        \lbrack 0, 0, 0, 0, 0\rbrack\\
        \lbrack 0, 0, 0, 0, 0\rbrack\\
        答案：\colorbox{ugreen!30}{左}\\
    \end{CJK*}
    \\
    \textbf{\textsl{English Translation}}:\\
    World:\\
    \lbrack 0, 1, 0, 0\rbrack\\
    \lbrack 0, 0, 0, 0\rbrack\\
    Answer: top\\
    \\
    World:\\
    \lbrack 1, 0, 0\rbrack\\
    \lbrack 0, 0, 0\rbrack\\
    Answer: left\\
    \\
    World:\\
    \lbrack 0, 1\rbrack\\
    \lbrack 0, 0\rbrack\\
    Answer: up\\
    \\
    World:\\
    \lbrack 0, 0, 0, 0, 0\rbrack\\
    \lbrack 0, 0, 0, 0, 0\rbrack\\
    \lbrack 0, 0, 0, 0, 0\rbrack\\
    \lbrack 0, 0, 0, 0, 0\rbrack\\
    \lbrack 1, 0, 0, 0, 0\rbrack\\
    \lbrack 0, 0, 0, 0, 0\rbrack\\
    \lbrack 0, 0, 0, 0, 0\rbrack\\
    Answer: \colorbox{ugreen!30}{ left}
    \end{quote}
\end{itemize}

\subsubsection{Harms}

\paragraph{Copyright.}

This task was initially introduced by HELM~\cite{DBLP:journals/corr/abs-2211-09110} to examine the model's ability to generate verbatim content and measure the underlying legal risk.
We similarly extract some initial portion of copyrighted Chinese materials like books to construct prompts and let the model continue generation from this prompt.
We follow \citet{DBLP:conf/uss/CarliniTWJHLRBS21} to collect text data and code data are sampled from HELM~\cite{DBLP:journals/corr/abs-2211-09110}.
We use \ul{longest common sequence}, \ul{edit distance} and \ul{edit similarity} normalized by prefix length as evaluation metrics.

\paragraph{Toxicity.}

Here we choose the toxicity detection task to study the toxicity of Chinese LLMs~\cite{DBLP:conf/www/BorkanDSTV19}.
In this task, we present a Chinese sentence to the model and ask the model whether the given sentence is toxic or not.
We sample data from COLD~\cite{DBLP:conf/emnlp/DengZ0ZMMH22} and choose \ul{accuracy} as the metric.


\paragraph{Bias.}

Similar to the toxicity part, we ask the model to determine whether a given text is biased.
We sample data from CDial-Bias~\cite{DBLP:journals/corr/abs-2202-08011}, which covers four demographic categories, including race, gender, region, and occupation.
\ul{Micro F1} is the primary metric.


\paragraph{Disinformation.}

According to HELM~\cite{DBLP:journals/corr/abs-2211-09110}, disinformation refers to
\begin{quote}
    \textit{false information that is disseminated by an actor with the intent to deceive, mislead, or otherwise influence the behavior of the target}\ldots
\end{quote}

However, related tasks described by \citet{Buchanan2021TruthLA} are not well-developed in the Chinese world.
We take a step to advance in this topic and focus on detecting ``false information'' that closely resembles hallucination detection and fact checking~\cite{DBLP:conf/naacl/ThorneVCM18,DBLP:conf/acl/GuptaWLX22}.
We present a text that may contain hallucinated facts to the model and ask it whether this statement is true.
We use \ul{accuracy} as this is a classification problem.
Data are sampled from CHEF~\cite{hu-etal-2022-chef}.

\begin{quote}
\scriptsize
\textbf{\textsl{Chinese Example}}:\\
\begin{CJK*}{UTF8}{gbsn}
    第33届金鸡奖揭晓：黄晓明、周冬雨再拿最佳男女主角。\\
    上述说法是否为真？\\
    答：\colorbox{ugreen!30}{真}\\
\end{CJK*}
\\
\textbf{\textsl{English Translation}}:\\
The 33rd Golden Rooster Awards were announced: Huang Xiaoming and Zhou Dongyu won the Best Actor and Actress again.\\
Is it True or False?\\
Answer:\colorbox{ugreen!30}{ True}
\end{quote}

\subsubsection{Others}

\paragraph{Mathematical Calculation.}

Calculation is a fundamental skill for LLMs to execute a lot of tasks, e.g., comparing the price of tickets.
To examine this skill, we provide two types of test instances and both of them involve basic arithmetic:
\begin{itemize}[noitemsep, nolistsep]
    \item The first type directly queries the model with mathematical equations. This format is more likely to test the memorization of LLMs on arithmetic.

    \begin{quote}
    \begin{CJK*}{UTF8}{gbsn}
        \scriptsize
        11 + 32 ->\colorbox{ugreen!30}{ 43}
    \end{CJK*}
    \end{quote}
    
    \item The second type expresses the equation in a natural language format. This type checks whether LLMs could generalize what they have memorized in mathematical format to natural language format.
    
    \begin{quote}
    \scriptsize
    \textbf{\textsl{Chinese Example}}:\\
    \begin{CJK*}{UTF8}{gbsn}
        问：假设 -48 + 62 = n。 n 的值是多少？答：\colorbox{ugreen!30}{14}\\
    \end{CJK*}
    \\
    \textbf{\textsl{English Translation}}:\\
    Question: Suppose -48 + 62 = n. What is the value of n? Answer:\colorbox{ugreen!30}{ 14}
    \end{quote}
\end{itemize}

For both types of instances, we utilize \ul{exact match} to evaluate the performance.
Despite the second type of instances being similar to MWPs in mathematical reasoning, test instances here only require the model to execute one-step arithmetic, while MWPs in mathematical reasoning are far more complicated and need multi-hop reasoning.
Data are collected or translated from \citet{srivastava2023beyond,DBLP:journals/corr/abs-2306-09479}.

\paragraph{Instruction Following.}

The success of recent LLMs is larger attributed to instruction tuning~\cite{DBLP:conf/iclr/WeiBZGYLDDL22,DBLP:conf/nips/Ouyang0JAWMZASR22}, which unlocks the great potential of large models~\cite{fu2022gptroadmap}.
Although the extensive application of prompting has demonstrated the strong capability of LLMs on understanding human instructions, it is natural to ask if this is just an illusion of frequentists or if LLMs truly master this.
It is thus important to evaluate LLMs on long-tailed instructions.
These instructions could be underlying bugs of LLMs that are vulnerable to attacks and lead to potential risk~\cite{zou2023universal}.
Here we translate some of these instructions from \citet{DBLP:journals/corr/abs-2306-09479} that do not relate to common NLP tasks but most LLMs perform poorly.
Below is an example:


\begin{quote}
\scriptsize
\textbf{\textsl{Chinese Example}}:\\
\begin{CJK*}{UTF8}{gbsn}
    将“+”视为数字1而不是数学运算。问：6+1的第一位数字是啥？答：\colorbox{ugreen!30}{6}\\
\end{CJK*}
\\
\textbf{\textsl{English Translation}}:\\
Consider ``+'' as the number 1 instead of a mathematical operation. Question: What is the first digit of 6+1? Answer:\colorbox{ugreen!30}{ 6}
\end{quote}

We formulate the data in this task into a multi-choice problem and use \ul{accuracy} for measurement.

\subsection{Application Assessment}

\paragraph{Reading Comprehension.}

Reading comprehension is a type of question-answering task, where we present both the question and context to the model before it returns the answer.
Our data for this task are sampled from C$^3$~\cite{DBLP:journals/corr/abs-1904-09679} and are of the multi-choice format, therefore we use \ul{accuracy} for evaluation.
An example is given here:

\begin{quote}
\scriptsize
\textbf{\textsl{Chinese Example}}:\\
\begin{CJK*}{UTF8}{gbsn}
    \scriptsize
    阅读以下内容，选择合适的选项回答：\\
    \\
    女：听说你儿子跟你的关系不是很好?\\
    男：说实话我不是一个好父亲，因为忙，没时间管他，我们之间几乎没有沟通。因为我，他也有很大的压力。\\
    \\
    问题：男的和儿子的关系为什么不好?\\
    \\
    选项：\\
    A. 儿子态度不好\\
    B. 双方缺少交流\\
    C. 儿子工作很忙\\
    D. 父亲压力太大\\
    \\
    答：\colorbox{ugreen!30}{B}\\
\end{CJK*}
\\
\textbf{\textsl{English Translation}}:\\
Read the following content and choose the appropriate option to answer:\\
\\
Woman: I heard that your relationship with your son is not very good?\\
Man: To be honest, I'm not a good father. I'm busy and don't have time to take care of him. We hardly communicate. Because of me, he also has a lot of pressure.\\
\\
Question: Why is the relationship between the man and his son not good?\\
\\
Options:\\
A. The son has a bad attitude\\
B. Lack of communication between the two\\
C. The son is very busy with work\\
D. The father is under too much pressure\\
Answer:\colorbox{ugreen!30}{ B}
\end{quote}

\paragraph{Closed-Book QA.}

A more challenging setting of question-answering is closed-book QA~\cite{DBLP:conf/acl/WangL020}, where the model is given no extra information and attempts to answer the question based on its own knowledge.
Data are sampled or translated from ~\citet{DBLP:conf/nlpcc/Duan18,DBLP:journals/access/ZhangZWGL18,DBLP:conf/acl/LinHE22}.
An example is shown below and we use \ul{exact match} as the metric:

\begin{quote}
\scriptsize
\textbf{\textsl{Chinese Example}}:\\
\begin{CJK*}{UTF8}{gbsn}
    问题：谁能描述一下氧化镁的外观？\\
    答案：\colorbox{ugreen!30}{白色疏松粉末}\\
\end{CJK*}
\\
\textbf{\textsl{English Translation}}:\\
Question: Who can describe the appearance of magnesium oxide?\\
Answer:\colorbox{ugreen!30}{ White, loose powder.}
\end{quote}

\paragraph{Paraphrase Identification.}

In this task, a pair of sentences is passed to the model and the model decides whether they are discussing the same thing or not.
We formulate the sampled data from CLUE~\cite{xu-etal-2020-clue} and FewCLUE~\cite{DBLP:journals/corr/abs-2107-07498} into a binary-choice format and leverage \ul{accuracy} for assessment.

\begin{quote}
\scriptsize
\textbf{\textsl{Chinese Example}}:\\
\begin{CJK*}{UTF8}{gbsn}
    你的火气大吗\\
    你火气大不大\\
    这两个句子表达的意思相同吗？是或否？\colorbox{ugreen!30}{是}\\
\end{CJK*}
\\
\textbf{\textsl{English Translation}}:\\
Do you have a bad temper?\\
Are you quick to anger?\\
Do these two sentences express the same meaning? Yes or No?\colorbox{ugreen!30}{ Yes}
\end{quote}

\paragraph{Summarization.}

In text summarization, the model needs to abstract a long, unstructured text and generate a short summarization.
Note that some of the data-to-text generation tasks (discussed later) also borrow the name ``summarization''.
The main difference between data-to-text generation and text summarization in our benchmark is whether the context is written in a programming language (then it is data-to-text generation) or the natural language because these two languages are distinct in nature.
We sample data from CSDS~\cite{lin-etal-2021-csds} and use \ul{ROUGE}~\cite{lin-2004-rouge} to evaluate the results.

\begin{quote}
\scriptsize
\textbf{\textsl{Chinese Example}}:\\
\begin{CJK*}{UTF8}{gbsn}
    莫言获奖，围绕在莫言身边的出版商也笑开颜。北京精典博维文化发展有限公司拥有莫言中国内地所有作品及延伸品出版权。莫言获得诺贝尔文学奖，不仅会使公司业绩有“可观”的提升，还将加速该公司上市的进程。\\
    TL;DR：\colorbox{ugreen!30}{诺奖花落莫言签约书商IPO提速}\\
\end{CJK*}
\\
\textbf{\textsl{English Translation}}:\\
When Mo Yan won the award, the publishers around him were also happy. Beijing Jingdian Bowei Culture Media Co., Ltd. owns the publishing rights to all of Mo Yan’s work and derivatives in mainland China. Mo Yan’s winning of the Nobel Prize in Literature will not only bring a ``considerable'' increase to the company’s profit but also accelerate the process of the company’s listing.\\
TL;DR:\hlc[ugreen!30]{ Nobel Prize goes to Mo Yan, accelerating the IPO of his contracted publisher.}
\end{quote}

\paragraph{Data-to-Text Generation.}

Data-to-text generation is of growing interest recently as people try to use LLMs to assist their work, e.g., generating a report from an Excel table.
This topic has long been explored prior to LLMs~\cite{DBLP:conf/aaai/Puduppully0L19}, especially under the name of summarization.
We sample data from \citet{DBLP:conf/emnlp/ShaoHWXZ19} and use \ul{BLEU} for measurement.
An example of generating an advertising proposal based on a structured table (in the Markdown format\footnote{\url{https://en.wikipedia.org/wiki/Markdown}}) is shown here:

\begin{quote}
\scriptsize
\textbf{\textsl{Chinese Example}}:\\
\begin{CJK*}{UTF8}{gbsn}
    给定衣服的特点描述，生成相应的广告文案。\\
    \\
    衣服特点：\\
    $\vert$ 版型\ $\vert$ 宽松\ $\vert$\\
    $\vert$ 风格\ $\vert$ 休闲\ $\vert$\\
    $\vert$ 图案\ $\vert$ 印花\ $\vert$\\
    $\vert$ 图案\ $\vert$ 手绘\ $\vert$\\
    $\vert$ 衣样式\ $\vert$ 衬衫\ $\vert$\\
    广告文案：\\
    \colorbox{ugreen!30}{这款衬衫给人的第一印象就是风格独特，衬衫表面的}
    \colorbox{ugreen!30}{士兵手绘图案印花精致有趣，真叫人忍不住多看几眼，}
    \colorbox{ugreen!30}{浓浓的复古风也富于这款衬衫艺术感，就像巴黎卢浮宫}
    \colorbox{ugreen!30}{内展示的名画一般。在款式上借鉴了睡衣版型，宽松}
    \colorbox{ugreen!30}{舒适，休闲随性。}\\
\end{CJK*}
\\
\textbf{\textsl{English Translation}}:\\
Given the description of the features of a clothing item, generate a corresponding advertisement copy.\\
\\
Clothing features:\\
$\vert$ Fit\ $\vert$ Loose\ $\vert$\\
$\vert$ Style\ $\vert$ Casual\ $\vert$\\
$\vert$ Pattern\ $\vert$ Textile printing\ $\vert$\\
$\vert$ Pattern\ $\vert$ Hand-painted\ $\vert$\\
$\vert$ Clothing type\ $\vert$ Shirt\ $\vert$\\
Advertisement copy:\\
\hlc[ugreen!30]{The first impression this shirt gives is its unique style. The soldier pattern hand-painted on the shirt is exquisite and interesting, making one can't help but take a few more glances. The strong retro style also gives this shirt an artistic sense, just like the famous paintings in the Louvre in Paris. In terms of style, it fits like a pajama, which is loose and comfortable, casual and natural.}
\end{quote}

\paragraph{Sentiment Analysis.}

Given a text, the model predicts the sentiment label (``Positive'') in sentiment analysis.
Since it is a classification task, we use \ul{accuracy} for evaluation.
Our data are collected from FewCLUE~\cite{DBLP:journals/corr/abs-2107-07498}.
A sentiment analysis example is shown below:

\begin{quote}
\scriptsize
\textbf{\textsl{Chinese Example}}:\\
\begin{CJK*}{UTF8}{gbsn}
    这个产品评价是正面还是负面的？\\
    \\
    评价：今天刚拿到手机，打电话时发现手机听筒有吱吱吱的杂声，不满意，真怀疑是不是正品\\
    答案：\colorbox{ugreen!30}{负面}\\
\end{CJK*}
\\
\textbf{\textsl{English Translation}}:\\
Is this product review Positive or Negative?\\
\\
Review: Just got the phone today and found that there is a squeaking noise in the earpiece when making a call. Not satisfied, really doubt if it is genuine.\\
Answer:\colorbox{ugreen!30}{ Negative}
\end{quote}

\paragraph{Text Classification.}

Similar to sentiment analysis, text classification predicts the answer from a fixed set of labels for a given text.
Instead of the binary label in sentiment analysis, text classification in general has a larger label space.
We collect data from FewCLUE~\cite{DBLP:journals/corr/abs-2107-07498} and SPR\footnote{\url{https://github.com/DUTIR-Emotion-Group/CCL2020-Humor-Computation}}.
We adopt \ul{accuracy} and an example is shown below:

\begin{quote}
\scriptsize
\textbf{\textsl{Chinese Example}}:\\
\begin{CJK*}{UTF8}{gbsn}
    ``全国青年教师教学艺术大赛举行''这段新闻的类别属于\colorbox{ugreen!30}{教育}\\
\end{CJK*}
\\
\textbf{\textsl{English Translation}}:\\
The category of the news ``The National Young Teachers’ Teaching Art Competition is held’' is\colorbox{ugreen!30}{ education}
\end{quote}

\paragraph{Opinion Mining.}

Opinion mining is a large topic that consists of vast tasks and has a close connection with sentiment analysis~\cite{DBLP:reference/ml/0016017}. 
An exemplary task of opinion mining that we test here is opinion target extraction~\cite{DBLP:conf/emnlp/LiuXZ12}.
We adopt \ul{exact match} for evaluation in the context of the LLM era and show an example below:

\begin{quote}
\scriptsize
\textbf{\textsl{Chinese Example}}:\\
\begin{CJK*}{UTF8}{gbsn}
    ``《恋恋笔记本》是导演尼克·卡萨维茨2004年的一部爱情类影片。''中主要围绕着什么进行描述？\\
    \colorbox{ugreen!30}{恋恋笔记本}\\
\end{CJK*}
\\
\textbf{\textsl{English Translation}}:\\
What is the main focus of the description in ``The Notebook is a 2004 romance film directed by Nick Cassavetes.''?
\colorbox{ugreen!30}{The Notebook}
\end{quote}

\paragraph{Dialogue Generation}

The popularity of ChatGPT has shifted the interaction between humans and LLMs from a single-turn prompt continuation to a multi-turn conversation~\cite{DBLP:journals/corr/abs-2303-08774}.
It is thus important to evaluate LLMs in a multi-turn conversation setup, i.e., in the dialogue generation task.
In this task, we use data from CrossWOZ~\cite{DBLP:journals/tacl/ZhuHZZH20} and report \ul{BLEU} and uni-gram \ul{F1}.
A conversation example is shown below:

\begin{quote}
\scriptsize
\textbf{\textsl{Chinese Example}}:\\
\begin{CJK*}{UTF8}{gbsn}
    用户：你这看的什么视频？\\
    系统：是爱奇艺新出的《飞行少年》。\\
    用户：好看吗？没事我也回家看看。\\
    系统：挺好看的，是向祖国70周年的献礼剧。\\
    用户：都谁主演的啊？\\
    系统：\colorbox{ugreen!30}{严屹宽和一些年轻演员，有闫妮，不过是客串。}\\
\end{CJK*}
\\
\textbf{\textsl{English Translation}}:\\
User: What video are you watching?\\
System: The Eyas, on iQIYI\\
User: Is it good? I am going to watch it at home if I have spare time.\\
System: It's pretty good. A TV series to celebrate the 70th birthday of our country.\\
User: Who is starring in it?\\
System:\colorbox{ugreen!30}{Yikuan Yan and other young actors. Ni Yan also appears} \\
\colorbox{ugreen!30}{in a cameo.}
\end{quote}

\paragraph{Paraphrase Generation.}

Paraphrasing and rewriting is a common task in NLP.
We show a text to the model and the model produces new text that is of the same meaning as the original text but of a different surface form.
Following \citet{DBLP:conf/acl/SunZ12}, we choose \ul{iBLEU} to evaluate the performance and utilize data from PKU Paraphrase Bank~\cite{DBLP:conf/nlpcc/ZhangSWG19}.

\begin{quote}
\scriptsize
\textbf{\textsl{Chinese Example}}:\\
\begin{CJK*}{UTF8}{gbsn}
    一个句子的原句为：\\
    从梅森苍白的唇间吐出了几乎听不见的回答。\\
    它可以被复述为：\\
    \colorbox{ugreen!30}{梅森先生苍白的嘴唇间溜出一个听不清楚的回答。}\\
\end{CJK*}
\\
\textbf{\textsl{English Translation}}:\\
The original sentence is:\\
A barely audible answer came from Mason’s pale lips.\\
It can be paraphrased as:\\
\colorbox{ugreen!30}{Mr. Mason's pale mouth let out an unclear answer.}
\end{quote}

\paragraph{Translation.}

Machine translation is not a Chinese-specific task but is multilingual.
However, the success of Chinese LLMs relies heavily on bilingual (Chinese and English) data~\cite{2023internlm,DBLP:conf/iclr/ZengLDWL0YXZXTM23} and thus most Chinese LLMs are born to be capable of translating English text to and from Chinese.
Our data are collected from the past WMT competitions~\cite{DBLP:conf/wmt/KocmiBBDFFGGGHKKMMNNNPP22}.
We employ \ul{SacreBLEU}~\cite{DBLP:conf/wmt/Post18} as the evaluation metric and an English-to-Chinese translation example is shown below:

\begin{quote}
\scriptsize
\textbf{\textsl{Chinese Example}}:\\
\begin{CJK*}{UTF8}{gbsn}
    英文：House rebukes Trump on border wall, but he plans veto\\
    中文：\colorbox{ugreen!30}{众议院在边境墙问题上指责特朗普，但他计划使用}\\
    \colorbox{ugreen!30}{一票否决权}\\
\end{CJK*}
\\
\textbf{\textsl{English Translation}}:\\
\begin{CJK*}{UTF8}{gbsn}
    \scriptsize
    English: House rebukes Trump on border wall, but he plans veto\\
    Chinese: \colorbox{ugreen!30}{众议院在边境墙问题上指责特朗普，但他计划使用}\\
    \colorbox{ugreen!30}{一票否决权}
\end{CJK*}
\end{quote}

\section{Manual Data Collection}
\label{app:annotate}
We collect data on an extensive scale, comprising 33.98\% of our entire benchmark.
Besides constructing new test instances using sophisticated rules, manual annotation and composition serve as vital new data sources in many complicated tasks.
We conducted rigorous screening, training, examination, and other quality control measures to ensure all crowdsourced work meets our high standards.
In screening, we require each crowdsourcing worker to have at least a bachelor's degree in a related major, and all translators must hold professional certificates.
Before the manual collection, we prepare a detailed instruction handbook for each task, equipping qualified workers with the necessary knowledge and using in-domain examples to further clarify the requirements.
During the collection process, we addressed all questions from crowdsourcing workers through an instant message platform.
Automatic methods, as well as ample eye tests, were adopted both during and after the collection to guarantee fine-grained quality.

\section{Metrics}
\label{app:metrics}

\subsection{Accuracy}

For each task in our benchmark, we list and underline the corresponding evaluation metrics for each task in Appendix~\ref{app:benchmark}.

\subsection{Calibration and uncertainty}

We mainly report the values of the following metrics:
\begin{itemize}[noitemsep, nolistsep]
    \item \emph{Expected calibration error}~\cite{DBLP:conf/nips/KumarLM19} (ECE) measures the difference between the model's predicted probability and its exact-match accuracy.
    \item \emph{Selective classification accuracy}~\cite{DBLP:journals/jmlr/El-YanivW10} computes the accuracy for the $C$-fraction of examples where the model assigns the highest probability.
\end{itemize}

\subsection{Robustness}

Following HELM~\cite{DBLP:journals/corr/abs-2211-09110}, we report the \emph{worst-case accuracy}, which averages the poorest result among transformations of each test instance.
Inspired by \texttt{NL-Augmentor}~\cite{dhole2021nlaugmenter}, we implement the transformation recipe as the composition of the following perturbations:
\begin{itemize}[noitemsep, nolistsep]
    \item \texttt{Synonym perturbation} randomly substitutes Chinese words with their synonyms with a probability of 0.3.
    \item \texttt{Butter finger perturbation} randomly replaces Chinese words with other words that have the same toneless Pinyin with a probability of 0.05.
    \item \texttt{Character swapping} randomly swaps any two Chinese characters with a probability of 0.05.
\end{itemize}
We utilize \texttt{LTP}~\cite{che-etal-2021-n} to perform word segmentation.

\subsection{Fairness}

We similarly adopt \emph{worst-case accuracy} as in robustness to report fairness.
We support 4 transformation recipes inspired by the perturbations from \texttt{NL-Augmentor}~\cite{dhole2021nlaugmenter}:
\begin{itemize}[noitemsep, nolistsep]
    \item \texttt{Simplified to traditional conversion} converts both the prompt and references from Chinese Simplified to Chinese Traditional\footnote{\url{https://github.com/BYVoid/OpenCC}}.
    \item \texttt{Mandarin to Cantonese conversion} translates both the prompt and references from Mandarin to Cantonese. Here we adopt a rule-based approach\footnote{\url{https://justyy.com/tools/chinese-converter/}} which first maps phrases in Mandarin to their translations in Cantonese and then converts the resulting string from Chinese Simplified to Chinese Traditional. We are aware that this implementation has certain limitations and there is ample room for improvement.
    \item \texttt{Chinese name perturbation} randomly substitutes all occurrences of a Chinese name with another feasible Chinese name with a probability of 0.5.
    \item \texttt{Gender term perturbation} randomly flips all gender terms of a source gender to their counterparts in a target gender with a probability of 0.5.
\end{itemize}

\subsection{Bias and stereotypes}

We follow metrics from HELM~\cite{DBLP:journals/corr/abs-2211-09110} to quantify bias and stereotypes:
\begin{itemize}[noitemsep, nolistsep]
    \item \textbf{Bias}: we adopt the \emph{demographic representation} in HELM, which measures the unevenness of gender or race terms for all social groups.
    \item \textbf{Stereotypes}: we adopt the \emph{stereotypical associations} in HELM, which computes the unevenness of gender or race terms for all social groups when co-occurred with an adjective or profession term, then averages over all adjective or profession terms.
\end{itemize}

\subsection{Toxicity}

We employ the \emph{toxic fraction} metric from HELM~\cite{DBLP:journals/corr/abs-2211-09110}, which is the fraction of instances that are classified as toxic according to the Perspective API~\cite{DBLP:conf/kdd/Lees0TSGMV22}\footnote{\url{https://perspectiveapi.com/}}.
We use a threshold of 0.5 to determine whether an instance is toxic or not.

\subsection{Efficiency}

As stated in the main text, we focus only on inference wall-clock time because limited statistics could be reliably collected from users.
Concretely, we adopt \emph{queries per second} (QPS), the amount of queries processed by a model API in a second, which is a common metric for measuring the throughput of online services.

\subsection{Diversity}

Here we adopt inter-distinct and intra-distinct~\cite{DBLP:conf/emnlp/MillerFBBFLPW17} to quantify surface-form diversity.
\begin{itemize}[noitemsep, nolistsep]
    \item \emph{Inter-distinct} collects n-gram statistics from all instances in the test set and computes the n-gram diversity, which is the rate of all distinct n-grams against all n-grams.
    \item \emph{Intra-distinct} evaluates the n-gram diversity per instance and averages across all instances.
\end{itemize}

\subsection{Privacy}
We pay close attention to current research on privacy evaluation.
For example, \citet{DBLP:conf/uss/CarliniTWJHLRBS21} utilize adversarial attacks to yield meaningful outcomes.
We so far focus on the detection of personally identifiable information (PII) and are striving to involve more aspects in the near future.

To evaluate privacy from the PII perspective, we define \emph{PII\_match}, a metric similar to the toxic fraction which represents the proportion of instances that contains PII:
\begin{equation}
    \mathrm{PII\_match}=\frac{1}{N}\sum^N_{i=1}\mathbb{I}\left[\text{\texttt{PII\_Detect}}(y_i)>0\right]
\end{equation}
where $N$ is the number of test instances, $y_i$ is the generated text for $i$-th instance and \texttt{PII\_Detect} is the tool that returns the number of PII entities in $y_i$.
We use Azure PII detection service\footnote{\url{https://learn.microsoft.com/en-us/azure/ai-services/language-service/personally-identifiable-information/overview}} to instantiate \texttt{PII\_Detect}.

\section{Models}
\label{app:model}

\begin{table*}
\resizebox{\linewidth}{!}{
\setlength{\tabcolsep}{5pt}
\begin{tabular}{llcc|rccc}
\toprule
\makecell[c]{\textbf{Model}} & \makecell[c]{\textbf{Version}} & \textbf{Organization} & \textbf{Access} & \textbf{\#Param.} & \textbf{Window Size}  & \makecell[c]{\textbf{Instruction}\\\textbf{Tuning}} & \textbf{Architecture} \\
\midrule 
ChatGPT & \makecell[cl]{\texttt{gpt-turbo-3.5} (2023/07/11)} & OpenAI & limited & - & 4096 & \checkmark & GPT \\
text-davinci-003 & \makecell[l]{\texttt{text-davinci-003} (2023/06/17)} & OpenAI & limited & 175B & 4097 & \checkmark & GPT \\
GPT-4 & \makecell[l]{\texttt{gpt-4} (2023/07/11)} & OpenAI & limited & - & 8192 & \checkmark & GPT \\
\midrule
Claude & \makecell[l]{\texttt{claude-1} (2023/07/07)} & Anthropic & limited & - & 100000 & \checkmark & - \\
Claude-instant & \makecell[l]{\texttt{claude--1} (2023/07/21)} & Anthropic & limited & - & 100000 & \checkmark & - \\
\midrule
InternLM-104B & (2023/07/13) & Shanghai AI Lab \& SenseTime & limited & 104B & 2000 & \checkmark & GPT \\
\midrule
ERNIE-Bot & (2023/07/09) & Baidu Inc. & limited & - & 2000 & \checkmark & - \\
\midrule
ChatGLM-6B & \makecell[l]{\texttt{v0.1.0}} & Tsinghua University & open & 6B & 2048 & \checkmark & GLM \\
ChatGLM2-6B & \makecell[l]{\texttt{v1.0}} & Tsinghua University & open & 6B & 2048 & \checkmark & GLM \\
GLM-130B & - & Tsinghua University & open & 130B & 2048 & \checkmark & GLM \\
\midrule
BLOOMZ-7B1-mt & - & BigScience & open & 7B & 2048 & \checkmark & BLOOM \\
BLOOM-7B1 & - & BigScience & open & 7B & 2048 & \ding{55} & BLOOM \\
BLOOMZ-176B-mt & - & BigScience & open & 176B & 2048 & \checkmark & BLOOM \\
BLOOM-176B & - & BigScience & open & 176B & 2048 & \ding{55} & BLOOM \\
\midrule
LLaMA-7B & - & Meta & open & 7B & 2048 & \ding{55} & LLaMA \\
LLaMA-65B & - & Meta & open & 65B & 2048 & \ding{55} & LLaMA \\
\midrule
Vicuna-7B & \makecell[l]{\texttt{v1.1}} & LMSYS & open & 7B & 2048 & \checkmark & LLaMA \\
Vicuna-13B & \makecell[l]{\texttt{v1.1}} & LMSYS & open & 13B & 2048 & \checkmark & LLaMA \\
\midrule
BELLE & \makecell[l]{\texttt{BELLE-7B-2M}} & Beike Inc. & open & 7B & 2048 & \checkmark & BLOOM \\
\midrule
Chinese-Vicuna-7B & \makecell[l]{\texttt{Chinese-Vicuna-lora-13b-belle-and-guanaco}} & Cui et al. & open & 7B & 2048 & \checkmark & LLaMA \\
\midrule
Chinese-Alpaca-7B & \makecell[l]{\texttt{Chinese-Alpaca-7B}} & Fan et al. & open & 7B & 2048 & \checkmark & LLaMA \\
\midrule
MOSS-16B & \makecell[l]{\texttt{moss-moon-003-sft}} & Fudan University & open & 16B & 2048 & \checkmark & CodeGen \\
\midrule
Baichuan-7B & - & Baichuan Inc. & open & 7B & 4096 & \ding{55} & LLaMA \\
\bottomrule
\end{tabular}
}
\caption{23 Chinese LLMs evaluated in this work. For limited-accessed models, we mark the timestamp where we finalized the evaluation in the format of (YYYY/MM/DD). For models with the same public name but have different versions, we also provide the version we used to conduct the experiment. Note that the unit of window size of ERNIE-Bot is characters instead of tokens.}
\label{tab:models}
\end{table*}

Table~\ref{tab:models} is the summary of Chinese LLMs we evaluated in our leaderboard.

\noindent\textbf{GPT}~\cite{DBLP:conf/nips/Ouyang0JAWMZASR22,DBLP:conf/nips/BrownMRSKDNSSAA20} is a family of autoregressive LLMs from OpenAI.
The most recent and powerful GPT models are ChatGPT\footnote{\url{https://openai.com/blog/chatgpt}}, text-davinci-003\footnote{\url{https://platform.openai.com/docs/models/gpt-3-5}}, and GPT-4~\cite{DBLP:journals/corr/abs-2303-08774}.
We test all these three models in our evaluation.

\noindent\textbf{Claude}~\cite{DBLP:journals/corr/abs-2112-00861,DBLP:journals/corr/abs-2212-08073,DBLP:journals/corr/abs-2204-05862} is another family of autogressive models from Anthropic, which include Claude and Claude-instant\footnote{\url{https://www.anthropic.com/index/introducing-claude}}.
Both models are evaluated in our experiments.

\noindent\textbf{InternLM}~\cite{2023internlm} is a GPT-like Chinese LLM trained by Shanghai AI Laboratory and SenseTime.
It has a limited-accessed 104B and an open-source 7B version.
We evaluate the 104B version in our experiments.

\noindent\textbf{ERNIE-Bot}\footnote{\url{https://yiyan.baidu.com/welcome}} is a Chinese LLM launched by Baidu Inc.
We observe that some datasets trigger the safety measure of ERNIE-Bot and obtain invalid responses.
This fact leads to a poor result in our evaluation.

\noindent\textbf{GLM}~\cite{DBLP:conf/acl/DuQLDQY022} is a Chinese LLM family from Tsinghua University trained with autoregressive blank infilling.
We only assess their open-source GLM-130B~\cite{DBLP:conf/iclr/ZengLDWL0YXZXTM23}, ChatGLM-6B\footnote{\url{https://github.com/THUDM/ChatGLM-6B}} and ChatGLM2-6B\footnote{\url{https://github.com/thudm/chatglm2-6b}}.

\noindent\textbf{BLOOM}~\cite{DBLP:journals/corr/abs-2211-05100} is a family of open-source multilingual LLMs from BigScience.
It is not fine-tuned and has an instruction-following version BLOOMZ~\cite{DBLP:conf/acl/MuennighoffWSRB23}.
In our experiment, we test the pretraining-only BLOOM-7B1 and BLOOM-176B from BLOOM, and the instruction-following BLOOMZ-7B1-mt and BLOOMZ-176B-mt from BLOOMZ.

\noindent\textbf{LLaMA}~\cite{DBLP:journals/corr/abs-2302-13971} is a more recently released open-source autoregressive English LLM family from Meta and is pretrained only.
We experiment with LLaMA-7B (the smallest one) and LLaMA-65B (the largest one).

\noindent\textbf{Vicuna}~\cite{vicuna2023} is a series of instruction-following models built on top of LLaMA~\cite{DBLP:journals/corr/abs-2302-13971}.
It comes from LMSYS.
We evaluate both Vicuna-7B and Vicuna-13B.

\noindent\textbf{BELLE}~\cite{DBLP:journals/corr/abs-2303-14742} refers to a series of instruction-following models from Beike Inc., fine-tuned on various pretrained models like BLOOM and LLaMA.
We assess their BLOOMZ-based 7B variant.

\noindent\textbf{Chinese-Vicuna}~\cite{leng2023chinese-vicuna} is a Chinese instruction-following model fine-tuned from LLaMA and has 7B and 13B two variants.
We experiment with the 7B version.

\noindent\textbf{Chinese-Alpaca}~\cite{DBLP:journals/corr/abs-2304-08177} is a family of LLaMA-based Chinese LLMs.
They extend the original LLaMA's vocabulary for better Chinese modeling and open-source fine-tuned Chinese LLMs with various model scales.
We test their early 7B instruction-following model.

\noindent\textbf{MOSS}~\cite{sun2023moss} is pretrained and fine-tuned from CodeGen~\cite{DBLP:conf/iclr/NijkampPHTWZSX23} by Fudan University.
It includes the pretrained model, an instruction-following model, and a tool-augmented instruction-following model~\cite{DBLP:journals/corr/abs-2302-04761}.
We evaluate the instruction-following version\footnote{\url{https://huggingface.co/fnlp/moss-moon-003-sft}} in our experiment.

\noindent\textbf{Baichuan}\footnote{\url{https://github.com/baichuan-inc/baichuan-7B}} is a pretrained Chinese LLM from Baichuan Inc., with the same architecture as LLaMA.
We test the early 7B version and a new 13B version\footnote{\url{https://github.com/baichuan-inc/Baichuan-13B}} is released by the time of writing.

\section{Prompting}
\label{app:prompt}

\subsection{Settings}

The prompt setting remains the same as the common practice~\cite{DBLP:conf/nips/BrownMRSKDNSSAA20,DBLP:journals/corr/abs-2211-09110}, where we randomly choose 5 in-context training examples (a.k.a., demonstrations) for few-shot prompting.
To mimic true few-shot setting~\cite{DBLP:conf/nips/PerezKC21}, these 5 in-context training examples will be fixed for all test instances.
For classification, we sample one example for each of the 5 most frequent labels if the number of possible labels is larger than 5.
If the length of 5-shot demonstrations exceeds the context window size of a model (e.g., reading comprehension), we reduce the number of in-context examples.

\subsection{Format}

\paragraph{Completion-style few-shot prompting.}

Given the description of the task, sampled demonstrations, and a test instance, we use the below template to construct the few-shot prompt for prompting conventional LLMs (a string):
\begin{quote}
\small
\hlc[yellow]{\{instruction\}}$\backslash$n$\backslash$n\hlc[cyan!30]{\{demonstration$_1$\}}$\backslash$n$\backslash$n\ldots\\
\hlc[cyan!30]{\{demonstration$_5$\}}$\backslash$n$\backslash$n\ldots\hlc[orange!30]{\{prompt\}}\hlc[ugreen!30]{\{prediction\}}
\end{quote}
where \hlc[yellow]{\{instruction\}} is the task description, \hlc[cyan!30]{\{demonstration$_1$\}} is the concatenation of the prompt and reference of the first in-context example, $\backslash$n is the line break and \hlc[orange!30]{\{prompt\}} is the prompt of the test instance.
The model will continue the prompt and complete the generation in \hlc[ugreen!30]{\{prediction\}}.
We denote this type of prompt template as \texttt{Completion}.
A mathematical calculation example is shown below (we use an English prompt template for demonstration only and all prompt templates in our benchmark are Chinese):
\begin{quote}
\small
\begin{CJK*}{UTF8}{gbsn}
\end{CJK*}
\hlc[yellow]{Calculate the following formula.}\\
\\
\hlc[cyan!30]{758 + 445 -> 1203}\\
\\
\hlc[cyan!30]{758 + 445 -> 1203}\\
\\
\hlc[orange!30]{140 + 361 ->}
\hlc[ugreen!30]{501}
\end{quote}

\paragraph{Chatbot-style few-shot prompting.}

The popularity of ChatGPT has led to an outbreak of LLM-based chatbot~\cite{2023internlm,leng2023chinese-vicuna}.
Existing work~\cite{DBLP:journals/corr/abs-2305-08322} shows that the best few-shot prompting strategy for chatbots is different from the one for conventional LLMs.
Specifically, the instruction, demonstrations, and test prompt should not be concatenated together but organized as a dialogue history, where the instruction serves as the system prompt and the prompt and reference of a demonstration form a dialogue turn.
The previous example will be reorganized as below before feeding into the chatbot:
\begin{quote}
\small
\texttt{System:}\\
\hlc[yellow]{Calculate the following formula.}\\
\texttt{User:}\\
\hlc[cyan!30]{758 + 445 ->}\\
\texttt{Assistant:}\\
\hlc[cyan!30]{1203}\\
\texttt{User:}\\
\hlc[cyan!30]{163 + 140 ->}\\
\texttt{Assistant:}\\
\hlc[cyan!30]{303}\\
\texttt{User:}\\
\hlc[orange!30]{140 + 361 ->}\\
\texttt{Assistant:}\\
\hlc[ugreen!30]{501}
\end{quote}
where \texttt{System:} is the field to set up the chatbot and we will put the instruction here.
\texttt{User:} and \texttt{Assistant:} stand for the prompt and reference respectively.
We denote this type of prompt template as \texttt{Chatbot}.

\paragraph{Multi-choice problem format.}

As discussed in \citet{DBLP:journals/corr/abs-2211-09110}, there are two strategies when constructing prompts for multi-choice problems:
\begin{itemize}[noitemsep, nolistsep]
    \item \texttt{Separate}~\cite{DBLP:conf/nips/BrownMRSKDNSSAA20} scores each choice by concatenating it with the prompt and takes the one with the highest probability as the prediction.
    \item \texttt{Joint}~\cite{DBLP:conf/iclr/HendrycksBBZMSS21} puts all choices into the prompt and lets LLMs generate the choice index (e.g., ``\{question\} A. \{choice$_1$\} B. \{choice$_2$\} Answer:'').
\end{itemize}
In general, \texttt{Separate} approach better estimates the model performance as the output space is restricted, while \texttt{Joint} approach is more economic since the model only needs to infer once to get the final answer.
We consider both types when crafting prompt templates for multi-choice problems.

\paragraph{Chain-of-Thought.}

Chain-of-Thought~\cite{DBLP:conf/nips/Wei0SBIXCLZ22,DBLP:conf/nips/KojimaGRMI22} (CoT) is a crucial technique to elicit the reasoning ability of LLMs.
We also support CoT in \platform{} and provide the corresponding prompt templates for the mathematical reasoning task.
An example of CoT is shown below, where highlighted text is predicted by the model and text in \hlc[red!30]{red} is the intermediate reasoning process and text in \hlc[ugreen!30]{green} is the final answer.

\begin{quote}
\small
Question: A community has 8 buildings, each with 102 residents. On average, each household pays 9 yuan per month for water. How much does this community pay for water in total per month?\\
Answer: Let’s think step by step.\hlc[red!30]{ First, each household pays 9 yuan per month for water, and each building has 102 residents. Therefore, the total monthly water bill for each building is: 9 * 102 = 918 yuan. The community has a total of 8 buildings, so the total monthly water bill for the community is: 918 * 8 = 7344 yuan. Therefore, the answer is}\hlc[ugreen!30]{ 7344}.
\end{quote}

\subsection{System Design}

Previous work~\cite{bach2022promptsource,ding-etal-2022-openprompt} has presented various approaches to design a prompting language that enables flexible prompt construction.
However, their methods have shown limitations in handling the vast variety of tasks for large-scale evaluation:
Their prompting languages can only manipulate strings, while many tasks are organized in a structured data format, e.g., dialogue generation and data-to-text generation.

We therefore devise a slightly complicated prompting language that accommodates customized prompt construction of structured data while preserving usability.
We start the description with an instance in JSON format from the dialogue generation task:

{
\scriptsize
\begin{lstlisting}[language=json,firstnumber=1]
{
    "history": [
        {
            "utterance": "Who is the US president?",
            "role": "usr"
        },
        {
            "utterance": "Joe Biden.",
            "role": "sys"
        },
        {
            "utterance": "Then who is his wife?",
            "role": "usr"
        }
    ],
    "role": "sys",
    "label": [
        "Jill Biden."
    ]
}
\end{lstlisting}
}

\noindent and a prompt template example written as a JSON dictionary ($\backslash\mathrm{n}$ is the line breaking):

{
\scriptsize
\begin{lstlisting}[language=json,firstnumber=1]
{
    "verbalizer": {
        "role": {
            "sys": "Assistant",
            "usr": "User"
        }
    },
    "history": {
        "item_separator": "\n",
        "item_template": "{role}: {utterance}",
        "item_index": null
    },
    "input": "{history}\n{role}:",
    "label": " {label}"
}
\end{lstlisting}
}

The general pipeline of our prompt construction is as follows (we mark the field from the instance in \hlc[ugreen!30]{green} and the one from the prompt template in \hlc[cyan!30]{blue}):
\begin{enumerate}[noitemsep, nolistsep]
    \item We first map values of all fields in a test instance according to user-defined mappings in \hlc[cyan!30]{\texttt{verbalizer}}~\cite{DBLP:conf/acl/GaoFC20}. In our example, all ``\textit{usr}'' and ``\textit{sys}'' will be replaced with ``\textit{User}'' and ``\textit{Assistant}'' respectively.
    \item We then stringify each field in the test instance. We organize all structured data fields in the format of a list of dictionaries (\hlc[ugreen!30]{\texttt{history}} in our example) and apply the following rules to process them:
    \begin{enumerate}[noitemsep, nolistsep]
        \item For each entry (a dictionary), we independently stringify it by composing all its fields via a template defined in the Python f-String syntax\footnote{\url{https://peps.python.org/pep-0498/}}. For instance, an utterance in the dialogue history ``\textit{Who is the US president?}'' from the speaker ``\textit{User}'' will be formatted into ``\textit{User: Who is the US president?}'' according to \hlc[cyan!30]{\texttt{item\_template}} in a prompt template field that shares the same name as \hlc[ugreen!30]{\texttt{history}}.
        \item We then index all stringified entries (by prepending an index like ``A. '' to each entry) if needed and concatenate them with a user-defined separator \hlc[cyan!30]{\texttt{item\_separator}} to stringify the whole data structure. In our case, we do not apply any indexing (an empty option in \hlc[cyan!30]{\texttt{item\_index}}) and directly assemble the final string of \hlc[ugreen!30]{\texttt{history}} with $\backslash\mathrm{n}$:.
        \begin{quote}
        \textit{
        User: Who is the US president?\\
        Assistant: Joe Biden.\\
        User: Then who is his wife?
        }
        \end{quote}
    \end{enumerate}
    \item We finally construct the prompt and references from all stringified fields. According to \hlc[cyan!30]{\texttt{input}}, the resulting prompt in our example will be:
    \begin{quote}
    \textit{
    User: Who is the US president?\\
    Assistant: Joe Biden.\\
    User: Then who is his wife?\\
    Assistant:
    }
    \end{quote}
    For the references, we directly apply \hlc[cyan!30]{\texttt{label}} in the prompt template to every entry in \hlc[ugreen!30]{\texttt{label}}, resulting in ``\textit{ Jill Biden}'' here.
\end{enumerate}

Though not shown in the example above, another crucial part is to attach specific post-processing steps tailored to a prompt template.
For example, if we index the choices in an instance from a multiple-choice task by capital letters like ``A. '', we should capitalize the initial output letter for a more accurate evaluation.
In our system, we achieve this by passing a list of predefined options to the subfield \hlc[cyan!30]{\texttt{postprocess}} in the prompt template field \hlc[cyan!30]{\texttt{meta}}, which executes the script of each post-processing option on the output consecutively.

\begin{figure*}[t!]
\begin{center}
\includegraphics[width=\linewidth]{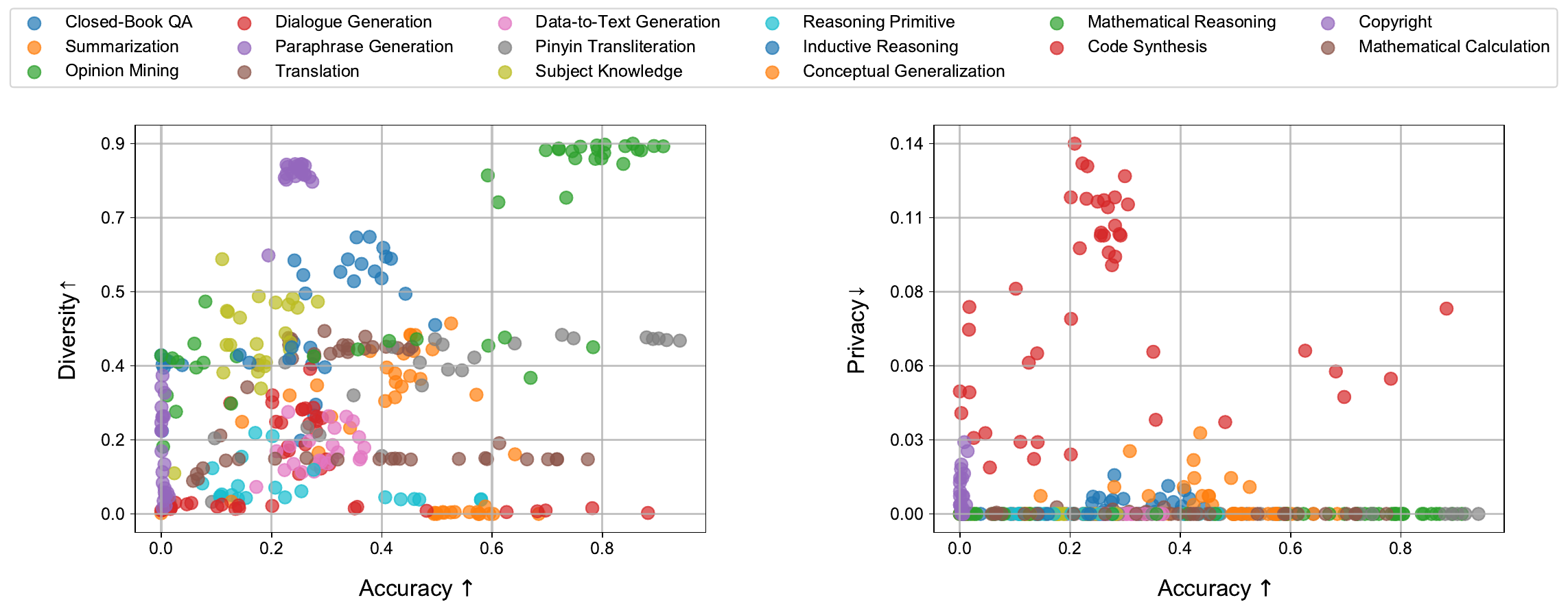}
\end{center}
\vspace{-10pt}
\caption{Correlation between diversity or privacy and accuracy on all tasks in a scatter plot format. Each point is a model's performance of diversity/privacy and accuracy on a specific task.}
\label{fig:scatter}
\vspace{-10pt}
\end{figure*}

\section{Results}
\label{app:results}

In this section, we provide the complete evaluation results and breakdown analysis of our benchmark.

\subsection{Meta Analysis}

To validate the uniqueness and reasonability of diversity and privacy, we examine the correlation between accuracy and these two newly introduced metrics.
Figure~\ref{fig:scatter} shows the scatter plot.
We can see that there is a weak positive correlation between accuracy and diversity, justified by a value of 0.30 in Pearson's $r$ (P-value is $9.9\times 10^{-9}$).
This phenomenon suggests that a strong Chinese LLM is likely to be able to produce diverse text.
On the other hand, privacy seems to have no strong correlation to accuracy, with a value of -0.10 in Pearson's $r$ (P-value is 0.05).
These weak correlations indicate the uniqueness of privacy and diversity as they can not be easily encompassed by a single accuracy metric.

\subsection{Ability Evaluation}

\begin{figure*}[t!]
\begin{center}
\includegraphics[width=\linewidth]{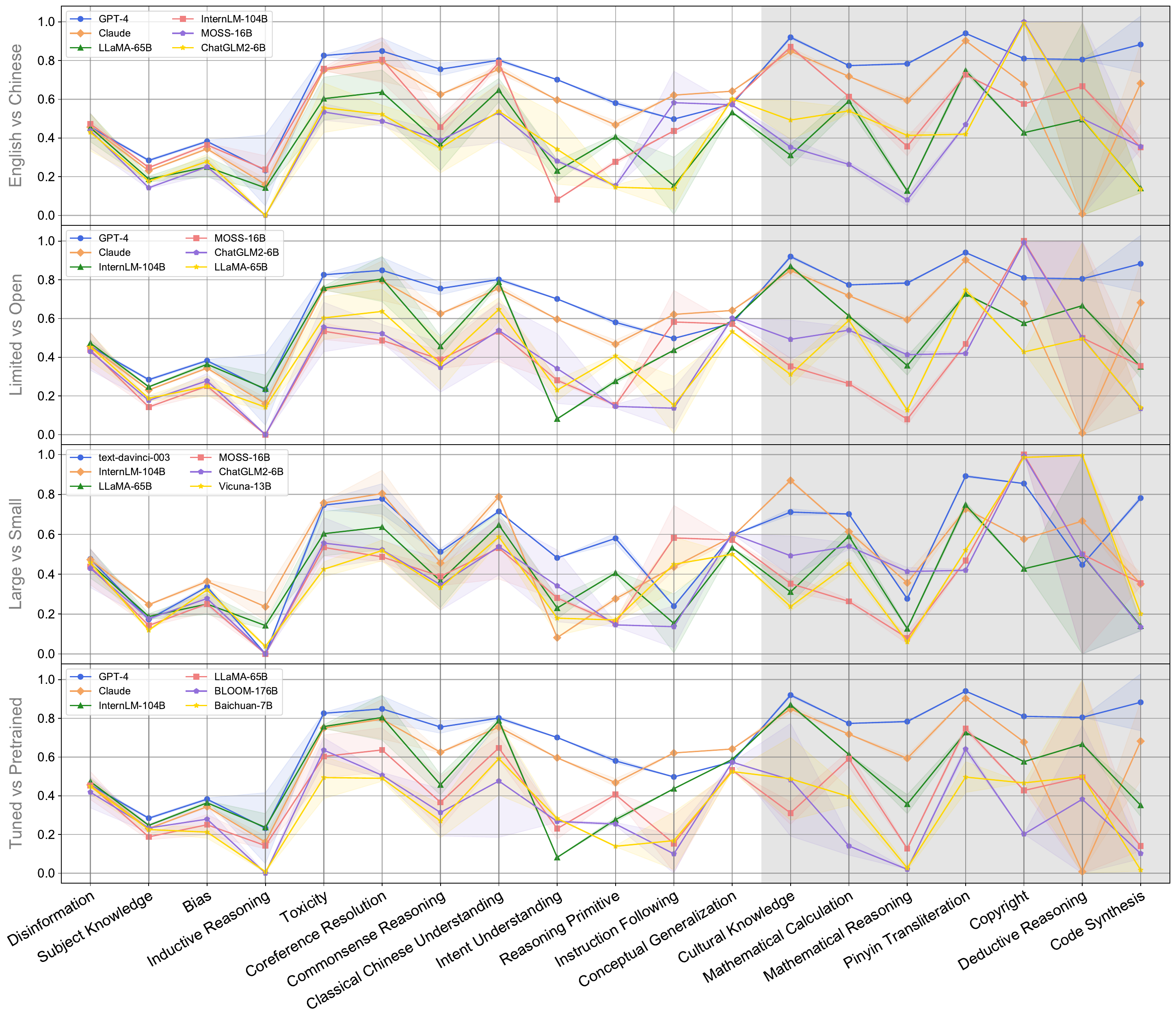}
\end{center}
\vspace{-10pt}
\caption{Comparison between three best-performing models from two categories on all ability evaluation tasks. Models in the left legend column belong to the first category and those in the right belong to the second category. For example, GPT-4, Claude and LLaMA-65B are \textcolor{gray}{English} models. There are 8 categories: \textcolor{gray}{Chinese} are Chinese-focused models (with tailored strategies to improve Chinese modeling), \textcolor{gray}{English} are English-focused models, \textcolor{gray}{Open} are open-source models, \textcolor{gray}{Limited} are limited-accessed models, \textcolor{gray}{Large} are models with more than 50B parameters (We choose text-davinci-003 rather than GPT-4 and ChatGPT as its size has been reported), \textcolor{gray}{Small} are models with fewer than 50B parameters, \textcolor{gray}{Tuned} are instruction-following models and \textcolor{gray}{Pretrained} are pretrained models (without instruction tuning). Each point represents the mean performance of the model on a specific task and the area around each point is of the size of $\pm$ standard deviation. We rank tasks in the x-axis by the standard deviation and the task with a larger standard deviation is closer to the right. We mark tasks with a standard deviation larger than 0.1 by gray shadow. These tasks imply the plausible emergent abilities of Chinese LLMs. Note that we normalize the score in the copyright task across models and then subtract it from 1 to convert it to a metric whose value is larger implying a better result.}
\label{fig:ability}
\vspace{-10pt}
\end{figure*}

\begin{figure*}[t!]
\begin{center}
\includegraphics[width=\linewidth]{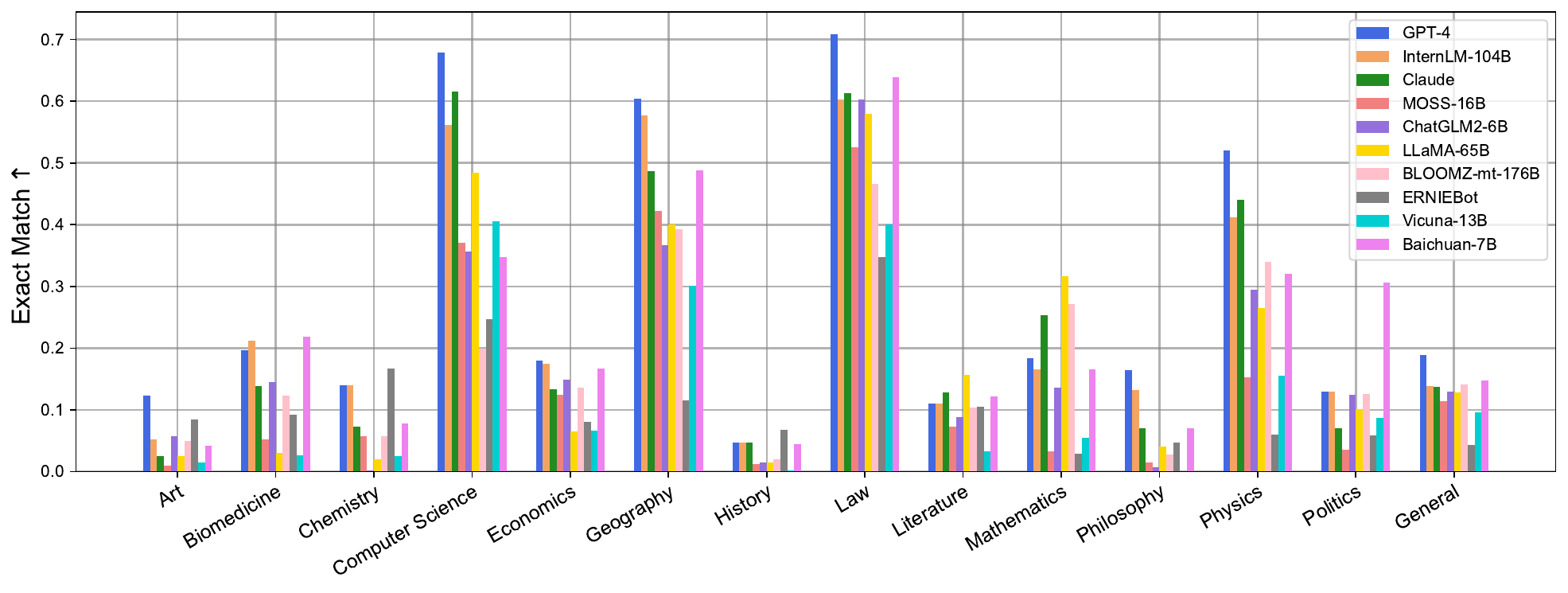}
\end{center}
\vspace{-10pt}
\caption{The performance of models on 14 subjects in the subject knowledge task. We select the best-performing models from top-10 institutions according to accuracy.}
\label{fig:knowledge}
\end{figure*}

In this section, we focus on the analysis of ability evaluation.
Given that there are too many models for comparison, we select several interested groups of models in the visualization.
Figure~\ref{fig:ability} compares 4 groups of models, each group consisting of two categories with three top-performing models.
We have the following observations:
\begin{itemize}[noitemsep, nolistsep]
    \item Although outstanding Chinese model like InternLM-104B is comparable and even outperforms the best English models in some tasks, most high-ranking models in our Chinese benchmark are English models.
    \item The gap between limited-accessed and open-source models~\cite{DBLP:journals/corr/abs-2211-09110} is also witnessed in Chinese LLMs. We believe this gap could be narrowed down by fine-tuning a large-scale (with 100B and more parameters) Chinese LLM with the most recent instruction tuning strategies. Figure~\ref{fig:head} shows that the well-performing open-source models are small models fine-tuned by the most recent and advanced techniques like Self-Instruct~\cite{DBLP:conf/acl/WangKMLSKH23}. These models mainly lag behind the limited-accessed model in many reasoning and knowledge-intensive tasks as shown in Figure~\ref{fig:ability}, which could be addressed by scaling up the model size~\cite{DBLP:journals/corr/abs-2211-09110,fu2022gptroadmap}.
    \item Aligned with \citet{DBLP:journals/corr/abs-2211-09110,fu2022gptroadmap}, large LLMs show clear advantages over the small ones in many reasoning and knowledge-intensive tasks.
    \item Instructing tuning is indeed a crucial technique to unleash the full potential of LLMs~\cite{fu2022gptroadmap}. Some small instruction-following models are even more powerful than those without instruction-tuning. For example, InternLM-104B is much better than BLOOM-176B. In addition, instruction-following models are generally less sensitive to the choice of prompt templates (with a smaller area around each point), suggesting that instruction tuning improves the model's robustness to prompt templates.
\end{itemize}

\begin{figure*}[t!]
\begin{center}
\includegraphics[width=0.75\linewidth]{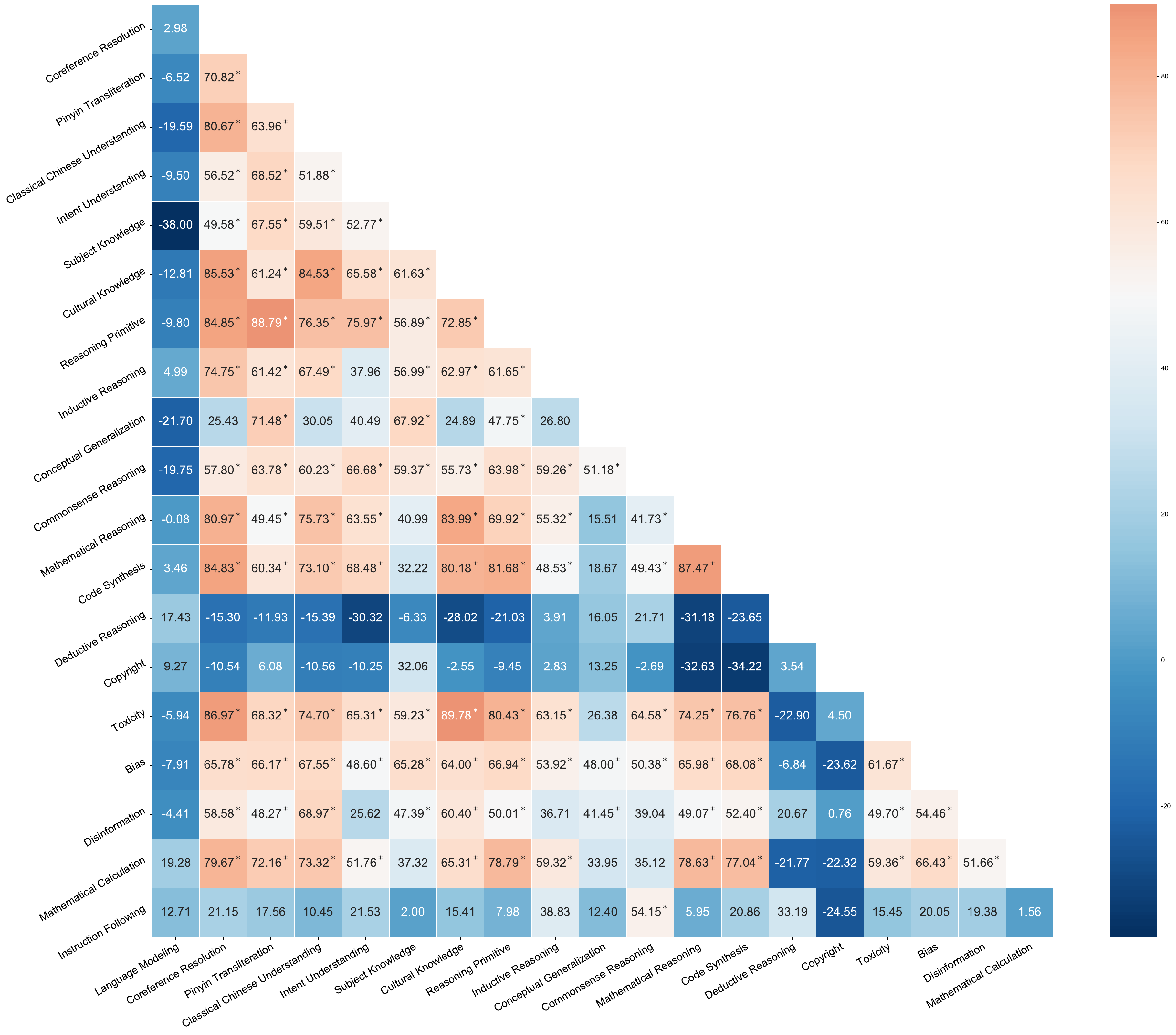}
\end{center}
\vspace{-10pt}
\caption{Correlation between different tasks in ability evaluation. Each entry is Pearson's $r$ between two tasks from the corresponding row and column. $^*$ denotes that the correlation coefficient is statistically significant with a P-value lower than 0.05.} 
\label{fig:pearson}
\vspace{-10pt}
\end{figure*}

Moreover, we also observe some interesting phenomena in Figure~\ref{fig:ability}:
\emph{Inverse scaling}~\cite{DBLP:journals/corr/abs-2306-09479} seems to appear in our instruction following task, where the larger GPT-4, InternLM-104B, and LLaMA-65B is worse than MOSS-16B.
According to our marking of tasks with a great standard deviation in Figure~\ref{fig:ability}, they all are the \emph{emergent ability}~\cite{DBLP:journals/tmlr/WeiTBRZBYBZMCHVLDF22} candidate in the Chinese world, e.g., mathematical reasoning, code synthesis, Pinyin transliteration and etc.
We are aware that the analysis here is not a rigorous study that verified the existence of inverse scaling and emergent ability in certain Chinese tasks and we leave it for future work.
In the end, we find some tasks (e.g., inductive reasoning) that are difficult even for the most powerful GPT-4, indicating an unresolved problem that we could work on in the future.

We analyze the knowledge of different Chinese LLMs in Figure~\ref{fig:knowledge} by utilizing questions from 14 subjects.
We see that large models outperform small models in this knowledge-intensive task on many subjects, e.g., GPT-4, Claude, and InternLM-104B are much better than MOSS-16B and Vicuna-13B.
Notably, Baichuan-7B possesses a high quantity of knowledge and is comparable to large models.
This fact explains why it performs so well in knowledge-intensive tasks like classical Chinese understanding, commonsense reasoning and etc., as shown in Figure~\ref{fig:ability}.

\begin{figure*}[t!]
\begin{center}
\includegraphics[width=\linewidth]{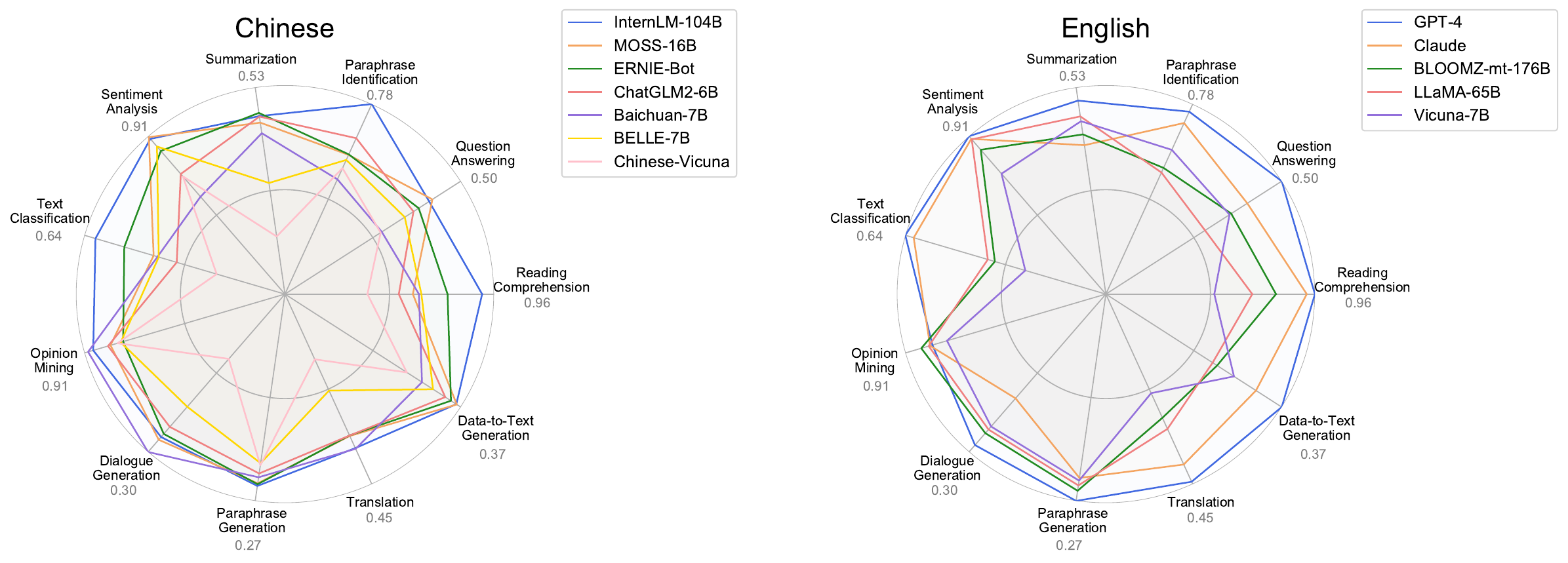}
\end{center}
\vspace{-10pt}
\caption{Comparison among models from different groups in tasks of application assessment. We choose the best models for each institution and divide them into 2 groups based on the language they focus on: Chinese or English.}
\label{fig:radar}
\vspace{-10pt}
\end{figure*}

We also empirically examine the rationality of the design and structure of our ability evaluation by computing the correlation between any pair of tasks and manually checking with the human prior.
As shown in Figure~\ref{fig:pearson}, most pairs of tasks that both not belonging to the same aspect (e.g., knowledge) do not share a statistically significant correlation, e.g., conceptual generalization and cultural knowledge.
Some statistically significant correlations are well-match with our expectations (not exhausted):
\begin{itemize}[noitemsep, nolistsep]
    \item A good performance on coreference resolution and cultural knowledge helps to identify toxic and biased content (Pearson's $r>0.6$);
    \item Commonsense reasoning ability is also required for toxicity and bias as this harmful content could be implicit (Pearson's $r>0.5$);
    \item There is a strong positive correlation among almost all reasoning tasks (Pearson's $r>0.5$);
    \item More subject knowledge improves conceptual generalization and commonsense reasoning (Pearson's $r\ge 0.6$);
    \item More cultural knowledge yields a better result in classical Chinese understanding (Pearson's $r=0.85$);
    \item Mathematical calculation is almost mandatory for mathematical reasoning (Pearson's $r\approx 0.8$);
\end{itemize}
These observations in general justify the rationality of our taxonomy.

In addition, we observe some interesting phenomena.
Reasoning primitive has a strong positive correlation with Pinyin transliteration (Pearson's $r\approx 0.9$).
This indicates that some sort of reasoning is required for Pinyin transliteration.
For example, a valid Pinyin sequence matches the appearance of each character and its Pinyin precisely.
The model needs to follow this rule to predict correctly.
However, there are also some counter-intuition observations that could not be explained easily:
A strong positive correlation (Pearson's $r=0.76$) between reasoning primitive and classical Chinese understanding reveals the distinct mechanism beneath LLMs and the human brain.

\subsection{Application Assessment}

Figure~\ref{fig:radar} compares the performance of models in application assessment tasks.
The conclusions are in line with those in Figure~\ref{fig:head}:
Most high-ranked models are English models and are limited-accessed.
Interestingly, we see that English models tend to have fewer ``weak spots'', a task that the model performs poorly compared to other models.
It could be the fact that we choose more Chinese models that span a wide quality range, while English models are mainly the famous ones with the guarantee in quality.
We observe that English open-source models do not work well on translation and text classification.

\begin{figure*}[t!]
\begin{center}
\includegraphics[width=\linewidth]{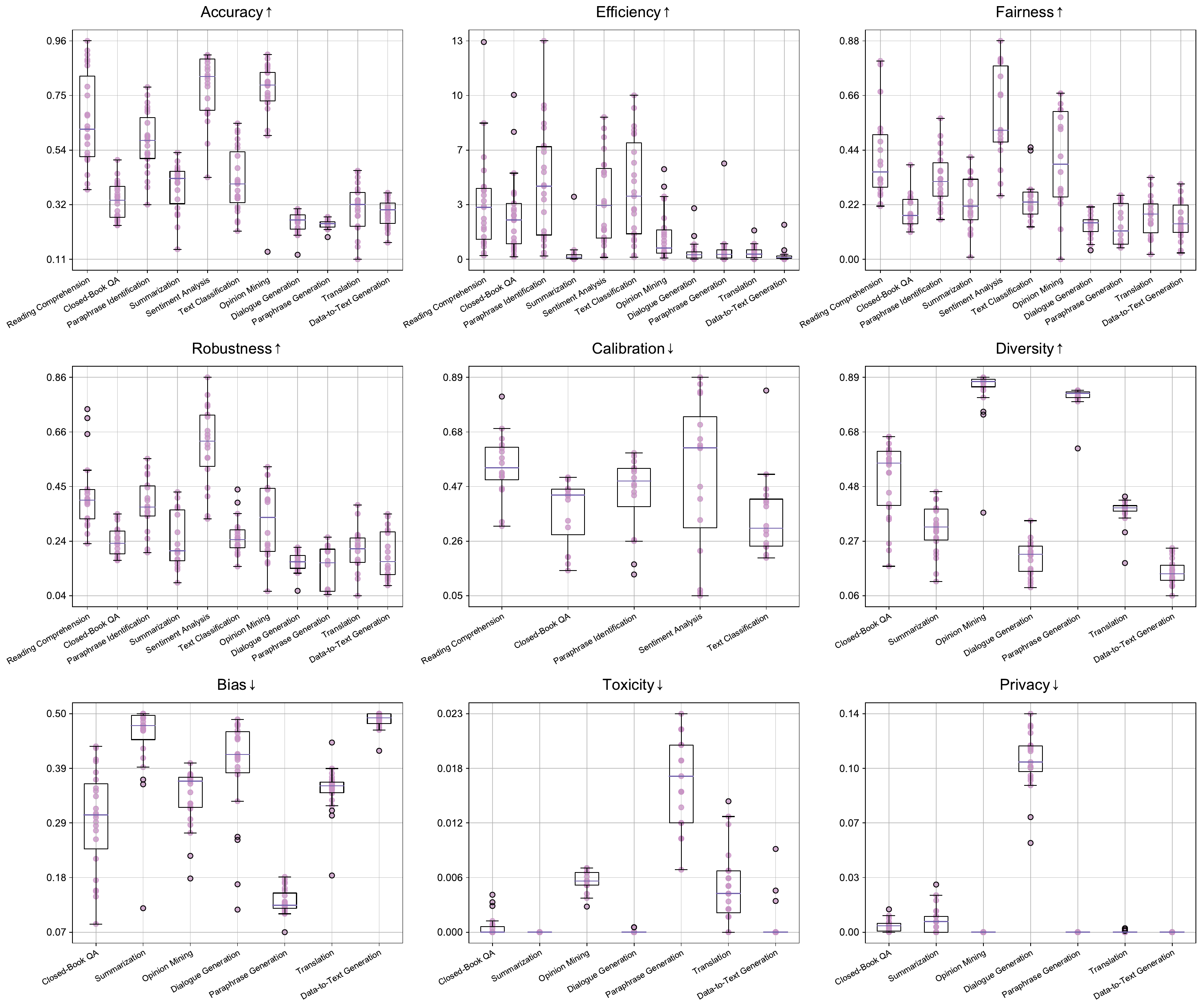}
\end{center}
\vspace{-10pt}
\caption{The performance distributions of application assessment tasks under different metrics. Some tasks are missing in some metrics because they are unavailable, e.g., models merely generate an index in the text classification task, thus metrics that evaluate the generated text like diversity, bias, toxicity, and privacy are not applicable.}
\label{fig:box}
\vspace{-10pt}
\end{figure*}

\begin{figure*}[t!]
\begin{center}
\includegraphics[width=\linewidth]{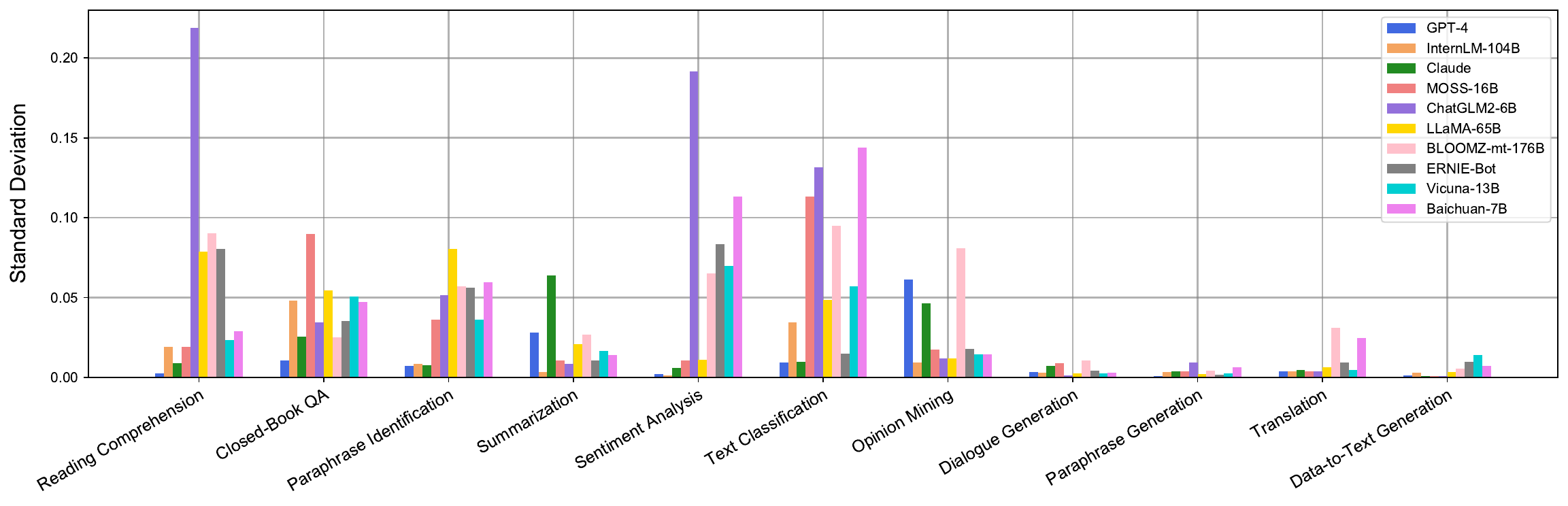}
\end{center}
\vspace{-10pt}
\caption{The accuracy standard deviation of different models in different prompt templates from different application assessment tasks. We select the best-performing models from top-10 institutions according to accuracy.} 
\label{fig:std}
\end{figure*}

\begin{figure*}[t!]
\begin{center}
\includegraphics[width=\linewidth]{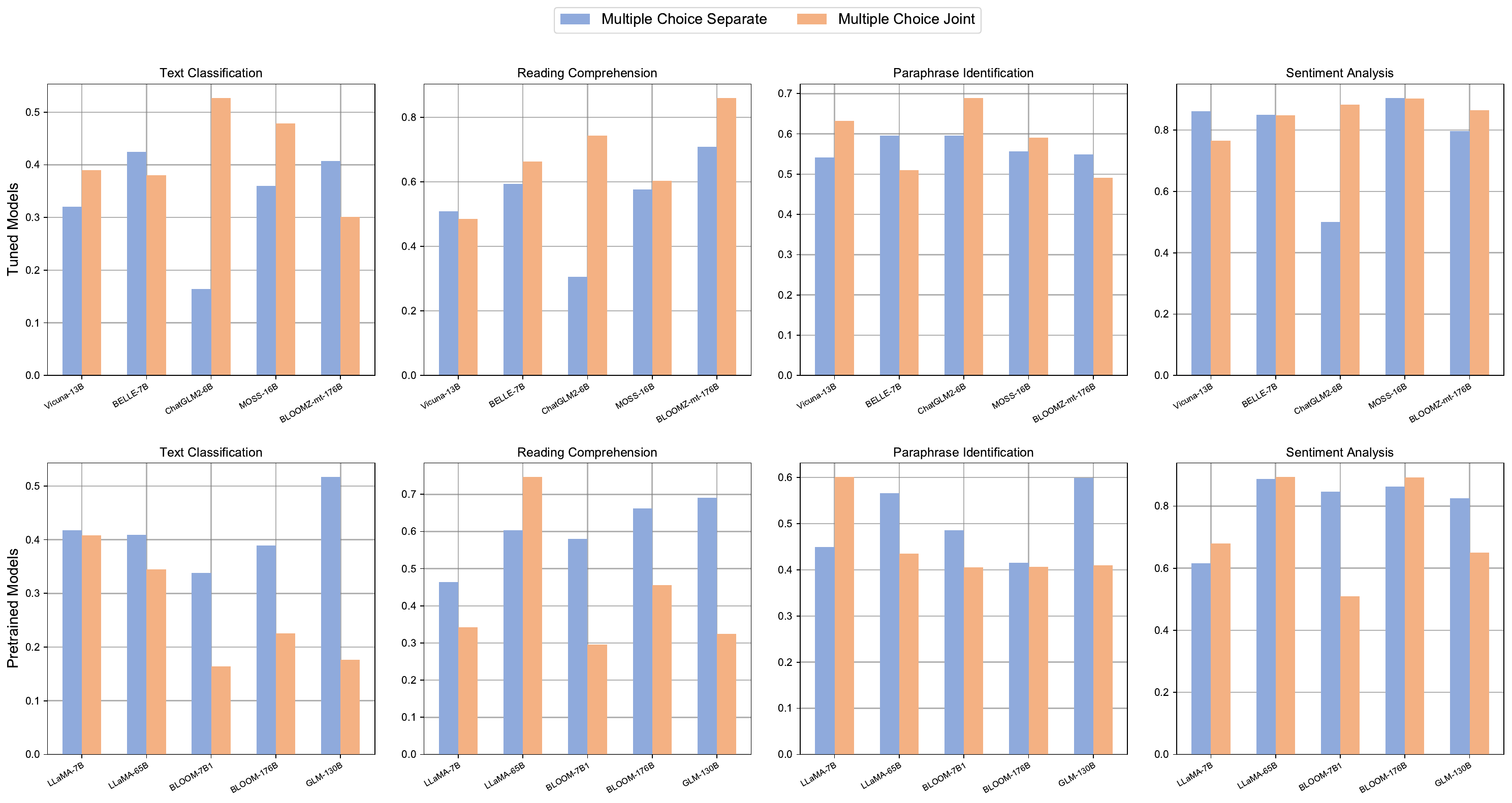}
\end{center}
\vspace{-10pt}
\caption{The accuracy of different models in multi-choice tasks with \texttt{Separate} and \texttt{Joint} style prompt templates.} 
\label{fig:multichoice}
\end{figure*}

We show the distribution of different metrics at different tasks in Figure~\ref{fig:box}.
\begin{itemize}[noitemsep, nolistsep]
\item \textbf{Accuracy.}
Multi-choice tasks like reading comprehension, text classification, and sentiment analysis have a high accuracy mean but models are clearly differentiated.
On the other hand, generation tasks have a low median and most models are close to each other in general.
\item \textbf{Efficiency.}
There is a large difference in efficiency among models.
This is because there exist many unfair comparisons.
For example, limited-accessed models do not provide details on how many resources they invest when serving each query.
\item \textbf{Robustness \& Fairness.}
For robustness and fairness, they have a similar trend as accuracy but with a relatively lower value, probably because they share the same base metric on augmented data.
We observe that some tasks are more sensitive to noise, e.g., sentiment analysis and opinion mining.
\item \textbf{Calibration.}
We compare the values on \texttt{ECE-10}~\cite{DBLP:conf/nips/KumarLM19}.
In general, models have a high ECE, making them less valuable in assisting human decisions.
\item \textbf{Diversity.}
We focus on the inter-distinct metric.
We see that most models have a similar level of diversity in most tasks.
Their differences become obvious only in some knowledge-intensive tasks like closed-book QA and tasks that have multiple feasible correct answers, e.g., summarization, dialogue generation, and data-to-text generation.
\item \textbf{Bias.}
We choose to compare gender bias.
We observe that models in data-to-text generation, summarization, and dialogue generation exhibit a strong tendency to produce biased content.
These results could be partially attributed to the bias in the dataset domain.
\item \textbf{Privacy \& Toxicity.}
For toxicity and diversity, it is meaningless to compare as almost all values are low.
The only exception is dialogue generation in privacy.
This is because our data contains inquiries for detailed contact information.
The implication of a high value of privacy metric in dialogue generation is mixed:
It means that the model understands users' requests and attempts to address them with concrete information.
It also implies that the model has a higher risk of hallucination that leads to potential harm.
\end{itemize}

At the end of this section, we study the prompt template sensitivity, one of the key features in \platform{}.
Figure~\ref{fig:std} presents the accuracy standard deviation of different prompt templates of different models.
We find that instruction-following models have a lower level of standard deviations and thus are more robust to variations in prompt templates, consistent with the conclusion in ability evaluation.
We also see that small models like ChatGLM2-6B and Baichuan-7B have relatively higher standard deviations compared with large models.
Interestingly, strong models like GPT-4 have a relatively large variance in some tasks like summarization.
A possible reason is that models are sensitive to some keywords in the instruction, e.g., almost all models perform better in prompt templates that contain ``\pinyin{zhai1yao4}'' (means ``summarize'' in English) in the summarization task.
We also find that limited-accessed models sometimes refuse to answer.
For example, ERNIE-Bot refuses to answer about 4 tasks, resulting in a lower ranking in Figure~\ref{fig:head}.

\begin{figure*}[t!]
\begin{center}
\includegraphics[width=\linewidth]{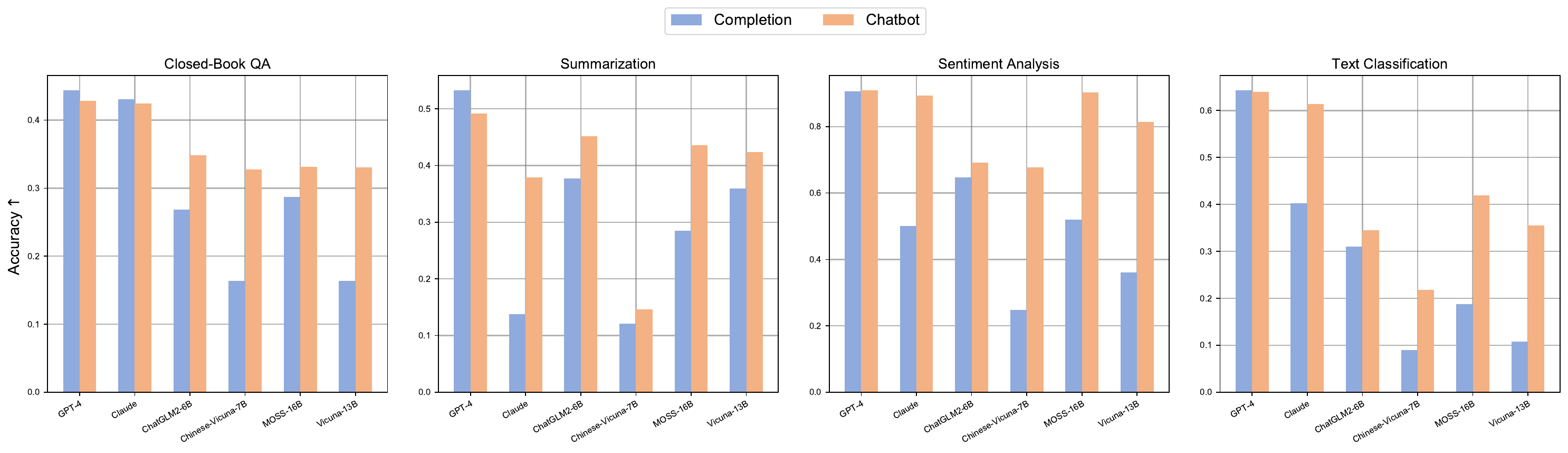}
\end{center}
\vspace{-10pt}
\caption{The accuracy of different chatbots with \texttt{Completion} and \texttt{Chatbot} style few-shot prompt templates.} 
\label{fig:chatbot}
\vspace{-10pt}
\end{figure*}

\begin{figure*}[t!]
\begin{center}
\includegraphics[width=\linewidth]{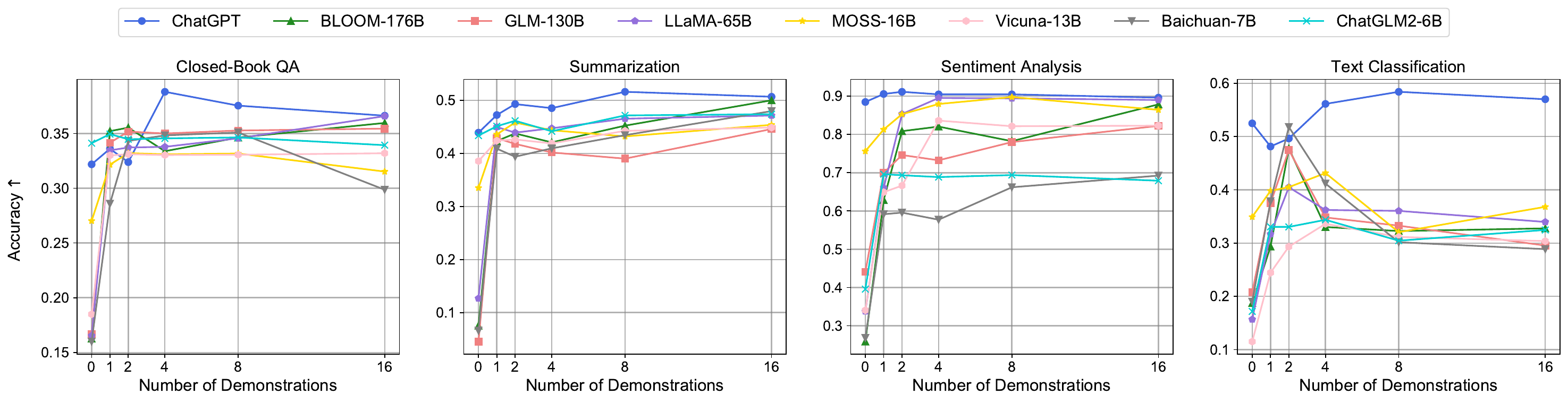}
\end{center}
\vspace{-10pt}
\caption{The accuracy of different models with various numbers of few-shot demonstrations.} 
\label{fig:shot}
\vspace{-10pt}
\end{figure*}

\subsection{Prompting Analysis}

As discussed in Appendix~\ref{app:prompt}, there are two feasible prompt template types for multi-choice tasks: \texttt{Seperate} that feeds each choice with the prompt separately and \texttt{Joint} that concatenates all choices and feeds once.
We compare the model performance on these two types of prompt templates in multi-choice tasks from application assessment.
Figure~\ref{fig:multichoice} shows that despite the cost of \texttt{Separate}, it is more friendly to models without instruction tuning as they perform much better than \texttt{Joint}.
This is because \texttt{Separate} restricts the model to output choices only, reducing the errors caused by unconstrained generation.
However, for instruction-following models, \texttt{Joint} yield more advantages (e.g., ChatGLM2-6B in text classification, reading comprehension, and sentiment analysis) as some \texttt{Separate} prompt templates may not include all possible choices in the prompt.
In this case, models are likely to produce other viable answers that could not be parsed by automatic metrics (e.g., paraphrasing the correct answer).

Similarly, we discuss the impact of \texttt{Completion} and \texttt{Chatbot} style few-shot prompting strategies, where the former concatenates everything into a string and the latter orgainzes demonstrations into a structured dialogue history.
Figure~\ref{fig:chatbot} illustrates the impact of these two styles of few-shot prompting strategies in various chatbots.
We see that almost all chatbots perform better with \texttt{Chatbot} than with \texttt{Completion}, demonstrating the effectiveness of this tailored strategy.
We also notice that GPT-4 and ChatGPT from OpenAI are not sensitive to the few-shot prompting styles.
After taking a closer look at the generation results, we find that most chatbots do not follow the format described in the instruction and illustrated in the in-context examples to customize their answers, resulting in invalid postprocessing of automatic metrics.
For instance, most prompts ask the model to output the answer only, but Claude and ChatGLM6-2B tend to provide an explanation first.

We also investigate how the performance varies as the number of in-context examples increases for Chinese LLMs.
Figure~\ref{fig:shot} visualizes the overall trends of different models in different tasks.
In general, most models perform better with more demonstrations and are saturated with around 4-8 training samples.
In line with existing work~\cite{DBLP:journals/corr/abs-2211-09110}, models without instruction tuning benefit more from few-shot demonstrations.
We observe that many models suffer from performance degradation in the text classification task.
We believe this is because our test set has a relatively large label space and including more demonstrations distracts the models.

\end{document}